\documentclass[10pt]{article} 
\usepackage[accepted]{tmlr}

\usepackage{amsmath,amsfonts,bm}

\def\eqref#1{equation~\ref{#1}}

\def\1{\bm{1}}

\DeclareMathAlphabet{\mathsfit}{\encodingdefault}{\sfdefault}{m}{sl}
\SetMathAlphabet{\mathsfit}{bold}{\encodingdefault}{\sfdefault}{bx}{n}

\usepackage{url}
\usepackage{microtype}
\usepackage{graphicx}
\usepackage{subfig}
\usepackage{booktabs} 
\usepackage{minitoc}
\usepackage{hyperref}
\hypersetup{colorlinks=true, linkcolor=blue, citecolor=blue}

\usepackage{amsmath}
\usepackage{amssymb}
\usepackage{mathtools}
\usepackage{amsthm}

\usepackage[capitalize,noabbrev]{cleveref}

\theoremstyle{plain}
\newtheorem{theorem}{Theorem}[section]
\newtheorem{proposition}[theorem]{Proposition}

\theoremstyle{definition}
\newtheorem{definition}[theorem]{Definition}
\newtheorem{assumption}[theorem]{Assumption}
\newtheorem{condition}[theorem]{Condition}
\theoremstyle{remark}
\newtheorem{remark}[theorem]{Remark}

\usepackage{enumitem}
\usepackage{bm}
\usepackage{algorithm}
\usepackage{algpseudocode}
\usepackage{multirow}

\title{FedDAG: Federated DAG Structure Learning}

\author{\name Erdun Gao\thanks{Work was done during an internship at JD Explore Academy.} \email erdun.gao@student.unimelb.edu.au \\
      \addr School of Mathematics and Statistics, The University of Melbourne
      \AND
      \name Junjia Chen \email cjj19970505@stu.xjtu.edu.cn \\
      \addr Faculty of Electronic and Information Engineering, Xi'an Jiaotong University
      \AND
      \name Li Shen \email shenli100@jd.com\\
      \addr JD Explore Academy 
      \AND
      \name Tongliang Liu \email tongliang.liu@sydney.edu.au \\
      \addr TML Lab, Sydney AI Centre, The University of Sydney \\
      Department of Machine Learning, Mohamed bin Zayed University of Artificial Intelligence
      \AND
      \name Mingming Gong \email mingming.gong@unimelb.edu.au \\
      \addr School of Mathematics and Statistics, The University of Melbourne 
      \AND
      \name Howard Bondell \email howard.bondell@unimelb.edu.au \\
      \addr School of Mathematics and Statistics, The University of Melbourne
      }

\def\month{01}  
\def\year{2023} 
\def\openreview{\url{https://openreview.net/forum?id=MzWgBjZ6Le}} 

\begin{document}
\maketitle

\doparttoc 
\faketableofcontents

\begin{abstract}
To date, most directed acyclic graphs (DAGs) structure learning approaches require data to be stored in a central server. However, due to the consideration of privacy protection, data owners gradually refuse to share their personalized raw data to avoid private information leakage, making this task more troublesome by cutting off the first step. Thus, a puzzle arises: \textit{how do we discover the underlying DAG structure from decentralized data?} In this paper, focusing on the additive noise models (ANMs) assumption of data generation, we take the first step in developing a gradient-based learning framework named FedDAG, which can learn the DAG structure without directly touching the local data and also can naturally handle the data heterogeneity. Our method benefits from a two-level structure of each local model. The first level structure learns the edges and directions of the graph and communicates with the server to get the model information from other clients during the learning procedure, while the second level structure approximates the mechanisms among variables and personally updates on its own data to accommodate the data heterogeneity. Moreover, FedDAG formulates the overall learning task as a continuous optimization problem by taking advantage of an equality acyclicity constraint, which can be solved by gradient descent methods to boost the searching efficiency. Extensive experiments on both synthetic and real-world datasets verify the efficacy of the proposed method.
\end{abstract}

\section{Introduction}

Bayesian Networks (BNs) have become prevalent over the last few decades by leveraging the graph theory and probability theory to model the probabilistic relationships among a set of random variables, which can potentially provide a mechanism for evidential reasoning~\citep{pearl1985bayesian}. The success of BNs has contributed to a furry of downstream real-world application problems in econometrics~\citep{heckman2008econometric}, epidemiology~\citep{greenland1999causal}, biological sciences~\citep{imbens2015causal} and social sciences~\citep{marini1988causality}. However, learning the graph structure of a BN from purely observational data remains a significant challenge due to the NP-hard property, and therefore, cutting-edge research that has drawn considerable attention in both academic and industrial fields~\citep{koller2009probabilistic, jensen2007bayesian, glymour2019review, zheng2018dags, zheng2020thsis}.

\begin{figure*}[ht]
\centering
\includegraphics[width=0.9\textwidth]{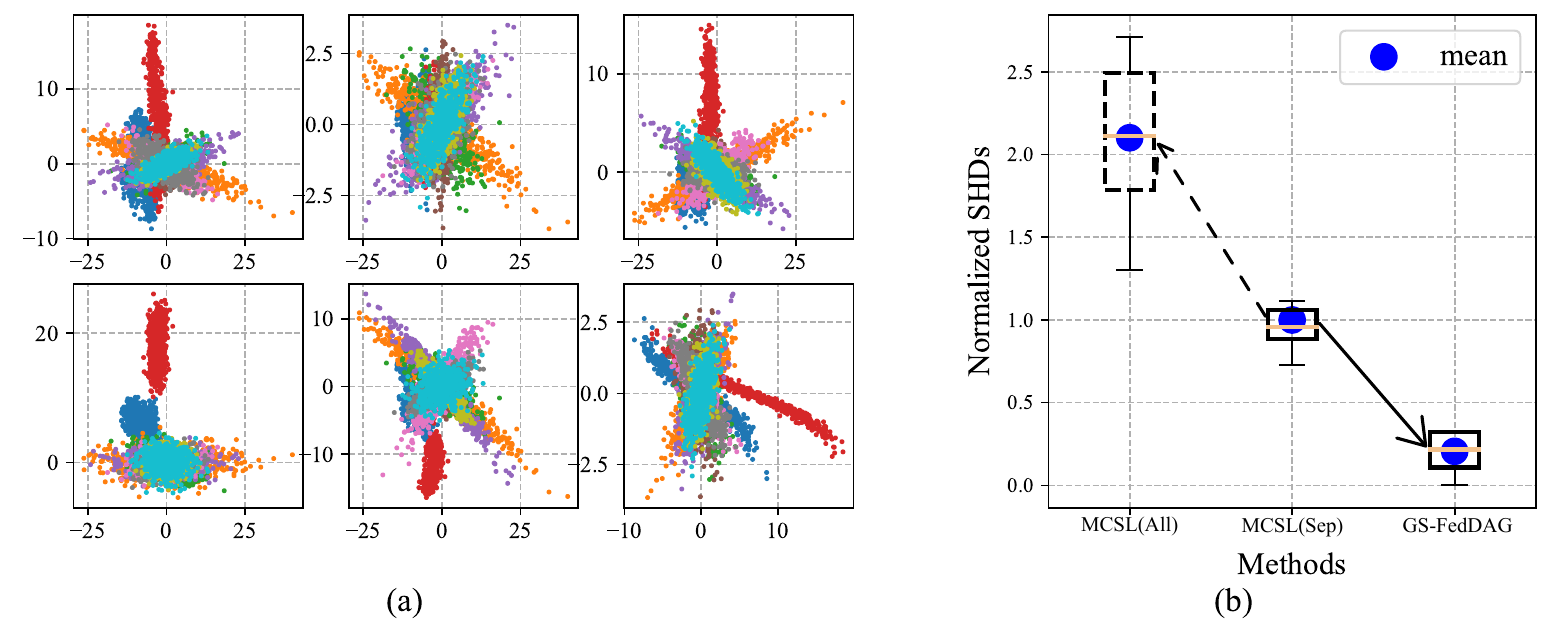}
\caption{(a) Visualization of heterogeneous data. Different colors represent data from different sources, while each sub-figure includes the distribution of one fixed dimension of data from all clients. (b) Normalized structural hamming distances (SHDs) ($\downarrow$) of three methods, where MCSL (Sep)~\citep{ng2022masked} separately trains one model on local data while MCSL (All) trains one model on all data, which is forbidden in FL.}
\label{fig:partial_visual_noniid}
\end{figure*}

Each BN is defined by a directed acyclic graph (DAG) and a set of parameters to depict the direction, strength, and shape of the mechanisms between variables~\citep{kitson2021survey}. Over the recent decades, a bunch of methods~\citep{spirtes2001causation, chickering2002optimal, zheng2020thsis} for discovering the DAG structure encoded among concerned events from the observational data have been proposed. In practice, however, finite sample problem bears the brunt of the performance decrease of DAG structure learning methods. Regularly (1) collecting data from various sources and then (2) designing the structure learning algorithm on all collected data can serve as a direct and standard pipeline to alleviate this issue in this field. However, owing to the issue of data privacy, data owners gradually prefer not to share their personalized data\footnote{Notice that in this paper, we restrict our scope to define the privacy leakage by sharing the raw data of users.} with others~\citep{kairouz2019advances}. Naturally, the new predicament, \textit{how do we discover the underlying DAG structure from decentralized data?} has arisen. In statistical learning problems such as regression and classification, federated learning (FL) has been proposed to learn from locally stored data~\citep{mcmahan2017communication}. Inspired by the developments in FL, we aim to develop a federated DAG structure learning framework that enables learning the graph structure from the decentralized data. Compared to the traditional FL methods in statistical learning, federated DAG learning, a \textbf{structural learning} task, has the following two main \textbf{differences}:
\begin{itemize}[leftmargin=*]
\item \textbf{Learning objective difference.} Most of the previous FL researches focus on learning \textit{an estimator} to estimate the conditional distribution $P(Y|X)$ in supervised learning tasks, e.g., image classification~\citep{mcmahan2017communication, li2020federated}, sequence tagging~\citep{bill2022fednlp}, and feature prediction~\citep{kairouz2019advances}. However, DAG structure learning, an unsupervised learning task~\citep{glymour2019review}, tries to \textit{find the underlying graph structure} among the concerned variables and the relationship mechanisms estimators to fit with the joint distribution of observations.

\item \textbf{Data heterogeneity difference.} The setup of data heterogeneity in FL is mainly assumed to be caused by some specific distribution shifts such as label shift (the shift of $P(Y)$)~\cite{lipton2018detecting} or covariate shift (the shift of $P(X)$)~\cite{reisizadeh2020robust}, while federated DAG learning handles a generative model, which can allow the data heterogeneity resulted from the joint distribution shift of all variables (Figure~\ref{fig:partial_visual_noniid}(a)). This would bring more challenges than the FL paradigm model design.
\end{itemize}

In this paper, we present FedDAG, a gradient-based framework for learning the underlying graph structure from decentralized data, including the case of heterogeneous data caused by mechanisms change. $(1)$ To alleviate the data leakage problem, FedDAG inherits the merits of FL, which proposes to deploy a local model to each client separately and collaboratively learn a joint model at the server end. Instead of sharing raw data, FedDAG exchanges model info among clients and the server to achieve collaboration. $(2)$ Taking into consideration of the first main difference between FedDAG and FL, a two-level structure consisting of a graph structure learning (GSL) part and a mechanisms approximating (MA) part, respectively, is adopted as the local model. $(3)$ Benefiting from this separated structure, the second difference between FL and FedDAG can naturally be handled by only sharing the GSL parts of clients during the learning procedure and locally updating the MA parts to get with data heterogeneity. Moreover, we provide the structure identifiability conditions for learning DAG structure from decentralized data in Appendix~\ref{app:identifiability}. Our contributions are summarized as follows:
\begin{itemize}[leftmargin=*]
\item[$\bullet$] We introduce the federated DAG structure learning task under the assumption that the underlying graph structure among different datasets remains invariant while mechanisms vary. We also show the structure identifiability conditions of DAG learning from the decentralized data.
\item[$\bullet$] We propose FedDAG, which separately learns the mechanisms on local data and jointly learns the DAG structure to handle the data heterogeneity elegantly. Meanwhile, since $0$ bits of raw data are shared, but only parameters of the GSL parts of models, the requirement of privacy protection is guaranteed, and the communication pressure is relatively low.
\item[$\bullet$] We evaluate our proposed method on data following structure equation models of additive noise structure with a variety of experimental settings, including simulations and real datasets, against recent state-of-the-art algorithms to show its superior performance and the ability to use one model in all settings.
\end{itemize}

\subsection{Potential applications of FedDAG}
Compared to traditional DAG learning methods, the homogeneous data (no distribution shift) setup of our method brings one more assumption that local data cannot be directly collected owing to the consideration of \textit{privacy leakage}. Then, we further extend our model to heterogeneous data, where mechanisms among variables may also vary among different local data. Therefore, our method could be directly applied to the applications of DAG structure learning, where privacy is also very important.

The first example can come from medical science. Exploring the underlying relations among concerned events from healthcare data can help to understand the disease mechanisms and causes~\citep{yang2013causal}. However, in real medical scenarios, the clinical data of patients are extremely sensitive and related to personal privacy, which faces stringent data protection regulations, such as HIPAA regulations~\citep{annas2003hipaa}. Each hospital may own finite clinical data for some rare diseases, which is not enough for DAG learning. How can hospitals cooperate to analyze the pathology while preventing sharing the private information (raw diagnostic data)? Naturally, this challenge can be addressed by our method. Depending on how each hospital collects data, e.g., medical devices and survey design, the data in each hospital may not share the same distribution. The second example can come from the recommendation system (RS)~\citep{wang2020causal, yang2020federated}. Leveraging BN model into RS is becoming prevail to perform robust recommendations by de-confounding some spurious relations. As users pay more attention to privacy and governments also exacerbate many strict regulations like the general data protection regulation (GDPR)~\citep{voigt2017eu}, it increases difficulties in collecting raw personal data to the server. Accordingly, we think that RS can also benefit from FedDAG learning.

\section{Preliminaries}
\label{sec:prelimi}
\textbf{Additive Noise Models (ANMs).} We consider a specific structural equation model (SEM), which is defined as a tuple $\mathcal{M}=\langle \mathcal{X}, \mathcal{F}\rangle$, where $\mathcal{X}=\{X_1, X_2,\cdots, X_d\}$ is a set of concerned variables. And $\mathcal{F}=\{f_1, f_2, \cdots, f_d\}$ is a set of functions. Each $f_i$, called the relationship mechanism of $X_i$, maps $\epsilon_i \cup {\bf PA}_i$ to $X_i$, i.e., $X_i=f_i({\bf PA}_i, \epsilon_i)$, where the ${\bf PA}_i$ corresponds to the set including all direct parents of $X_i$ and $\epsilon_i$ is the random noise. $\mathcal{M}$ can be leveraged to describe how nature assigns values to variables of interest~\citep{pearl2016causal}. In this paper, we focus on a commonly used model named ANMs. They assume that
\begin{equation}
\label{ANM}
    X_i = f_{i}({\bf PA}_i) + \epsilon_i, \quad i=1,2,\cdots, d,
\end{equation}
where $\epsilon_i$ is independent of variables in ${\bf PA}_i$ and mutually independent with any $\epsilon_j$ for $i\neq j$. 

\textbf{Bayesian Networks (BNs).} Let $X=(X_1,X_2,\cdots,X_d)$ be a vector that includes all variables in $\mathcal{X}$ with the index set $\mathbb{V}:=\{1,2,\cdots,d\}$ and $P(X)$ with the probability density function $p(X)$ be a marginal distribution induced from $\mathcal{M}$. A DAG $\mathcal{G} = (\mathbb{V}, \mathbb{E})$ consists of a nodes set $\mathbb{V}$ and an edge set $\mathbb{E} \subseteq \mathbb{V}^2$. Every 
SEM $\mathcal{M}$ can be associated with a DAG $\mathcal{G}_{\mathcal{M}}$, in which each node $i$ corresponds to the variable $X_i$ and directed edges point from ${\bf PA}_i$ to $X_i$\footnote{In the intact graph structure of ANMs, we just fix directed edges from $\epsilon_i$ to $X_i$ and assume the distribution of $\epsilon_i$. Therefore, in this paper, $\mathcal{G}$ is only defined over the endogenous variables.} for $i\in [d]$\footnote{For simplicity, we use $[d] = \{1,2,\cdots,d\}$ to represent the set of all integers from $1$ to $d$.}. A BN is defined as a pair $\langle P(X), \mathcal{G}_{\mathcal{M}} \rangle$. Then $\mathcal{G}_{\mathcal{M}}$ is called the graph structure associated with $\mathcal{M}$ and $P(X)$ is Markovian to $\mathcal{G}_{\mathcal{M}}$. Throughout the main text, we assume that there is no latent variable\footnote{This assumption can be relaxed to some restricted cases with latent variables. See Appendix \ref{app:as_a_framework} for details.}~\citep{spirtes2001causation} and then $p(X)$ can be factorized as
\begin{equation}
\label{eq:product_factorization}
p(X) = \prod_{i=1}^{d} p(X_i|X_{pa_i})
\end{equation} 
according to $\mathcal{G}_{\mathcal{M}}$~\citep{lauritzen1996graphical}. $X_{pa_i}$ is the parental vector that includes all variables in ${\bf PA}_i$.

\textbf{Characterizations of Acyclicity.} A DAG $\mathcal{G}$ with $d$ nodes can be represented by a binary adjacency matrix $\bm B = [ \bm B_{:,1} | \bm B_{:,2} | \cdots | \bm B_{:,d}] $ with $\bm B_{:,i} \in \{0,1\}^ d$ for $\forall i \in [d]$. NOTEARS \citep{zheng2018dags} formulates a sufficient and necessary condition for $\bm B$ representing a DAG by an equation. The formulation is as follows:
\begin{equation}
\label{eq:acyclicity_1}
    {\rm Tr}[e^{\bm B}] - d = 0,
\end{equation}
where ${\rm Tr}[\cdot]$ means the trace of a given matrix. $e^{(\cdot)}$, here, is the matrix exponential operation. However, NOTEARS is only designed to solve the linear Gaussian models, which assume that all relationship mechanisms are linear. Therefore, the DAG structure and relationship mechanisms can be modeled together by a weighted matrix. To extend NOTEARS to the non-linear cases, MCSL~\citep{ng2022masked} proposes to use a mask $\bm M$, parameterized by a continuous proxy matrix $\bm U$, to approximate the adjacency matrix $\bm B$. To enforce the entries of $\bm M$ to approximate the binary form, i.e., $0$ or $1$, a two-dimensional version of Gumbel-Softmax~\citep{jang2016categorical} approach named Gumbel-Sigmoid is designed to reparameterize $\bm U$ and to ensure the differentiability of the model. Then, $\bm M$ can be obtained element-wisely by
\begin{equation}
\label{Gumbel_sigmoid}
    \bm M_{ij} = \frac{1}{1+ {\rm exp}(-{\rm log}(\bm U_{ij} + {\rm Gumb}_{ij})/\tau)},
\end{equation}
where $\tau$ is the temperature, ${\rm Gumb}_{ij} = g^1_{ij}- g^0_{ij}$, $g^1_{ij}$ and $g^0_{ij}$ are two independent samples from ${\rm Gumbel}(0,1)$. For simplicity but equivalence, $g^1_{ij}$ and $g^0_{ij}$ also can be sampled from $ -{\rm log}({\rm log}(a))$ with $a \sim {\rm Uniform}(0,1)$. See Appendix D in~\citep{ng2022masked}. MCSL names Eq. (\ref{Gumbel_sigmoid}) as Gumbel-Sigmoid w.r.t. $\bm U$ and temperature $\tau$, which is written as $g_{\tau}(\bm U)$. Then, the acyclicity constraint can be reformulated as 
\begin{equation}
    {\rm Tr}[e^{(g_{\tau}(\bm U))}] - d = 0.
\end{equation}

\section{Problem definition}
\label{problemdef}
Here, we first describe the property of decentralized data and the data distribution shift among different clients if there exists data heterogeneity~\citep{huang2020causal, mooij2020joint, zhang2020domain}. Then, we define the problem, federated DAG structure learning, considered in this paper.

\textbf{Decentralized data and probability distribution set.} Let $\mathcal{C} = \{c_1, c_2, \cdots, c_m\}$ be the client set which includes $m$ different clients and $s$ be the only server. The data $\mathcal{D}^{c_k} \in \mathbb{R}^{n_{c_k} \times d}$, in which each observation $\mathcal{D}^{c_k}_i$ for $\forall i\in [n_{c_k}]$ independently sampled from its corresponding probability distribution $P^{c_k}(X)$, represent the personalized data owned by the client $c_k$. $n_{c_k}$ is the number of observations in $\mathcal{D}^{c_k}$. The dataset $\mathcal{D} = \{\mathcal{D}^{c_1}, \mathcal{D}^{c_2}, \cdots, \mathcal{D}^{c_m}\}$ is called a decentralized dataset and $P^{\mathcal{C}}(X) = \{P^{c_1}(X), P^{c_2}(X), \cdots, P^{c_m}(X)\}$ is defined as the decentralized probability distribution set. If $P^{c_{k_1}}(X)=P^{c_{k_2}}(X)$ for $\forall\ k_1,k_2 \in [m]$, then $\mathcal{D}$ is defined as a homogeneous decentralized dataset throughout this paper. The heterogeneous decentralized dataset is defined by assuming that there exists at least two clients, e.g., $c_{k_1}$ and $c_{k_2}$, on which the local data are sampled from different distributions, i.e., $P^{c_{k_1}}(X) \neq P^{c_{k_2}}(X)$.
\begin{assumption}
\label{ass:sameDAG}
{\rm \textbf{(Invariant DAG)}} For $\forall c_{k}$, $P^{c_k}(X) \in P^{\mathcal{C}}(X)$ admits the product factorization of Eq.~(\ref{eq:product_factorization}) relative to the same DAG $\mathcal{G}$. 
\end{assumption}

\begin{remark}
\label{remark:markov}
If $P^{\mathcal{C}}(X)$ satisfies Assumption~\ref{ass:sameDAG}, then, each $P^{c_k}(X) \in P^{\mathcal{C}}(X)$ is Markovian relative to $\mathcal{G}$.
\end{remark}

According to the general definition of \textit{mechanisms change} in~\citep{tian2001causal}, interventions can be seen as a special case of distribution shifts, where the external influence involves fixing certain variables to some predetermined values. Actually, in general, the external influence may be milder to merely change the conditional probability of certain variables given its causes. In this paper, we restrict our scope by assuming that the distribution shifts across $P^{c_k}(X)$ come from the changes of mechanisms in $\mathcal{F}$ or distribution shifts of the exogenous variables in $\mathcal{E}$ (see Appendix~\ref{App:diss_data_heterogeneity} for detailed discussion). More justifications on Assumption~\ref{ass:sameDAG} are in Appendix~\ref{app:invariant_dag}.

\begin{assumption}
\label{ass:data_shift}
For $\forall c_{k_1}, c_{k_2}$, if $P^{c_{k_1}}(X) \neq P^{c_{k_2}}(X)$, the distribution shifts are caused by (1) $\exists\ i \in [d]$, $P^{c_{k_1}}(X_i | X_{pa_i}) \neq P^{c_{k_2}}(X_i | X_{pa_i})$, i.e., $f^{c_{k_1}}_i \neq f^{c_{k_2}}_i$. (2) $\exists\ i \in [d]$, $P^{c_{k_1}}(\epsilon_i) \neq P^{c_{k_2}}(\epsilon_i)$.
\end{assumption}

\textbf{Federated DAG Structure Learning.} Given the decentralized dataset $\mathcal{D}$ consisting of data from $m$ clients while the corresponding $P^{\mathcal{C}}(X)$ satisfies Assumptions~\ref{ass:sameDAG} and~\ref{ass:data_shift}, federated DAG structure learning aims to identify the underlying DAG $\mathcal{G}$ from $\mathcal{D}$.
\section{Methodology}
\label{method}

\begin{figure*}
\begin{center}
    \includegraphics[width=0.8\textwidth]{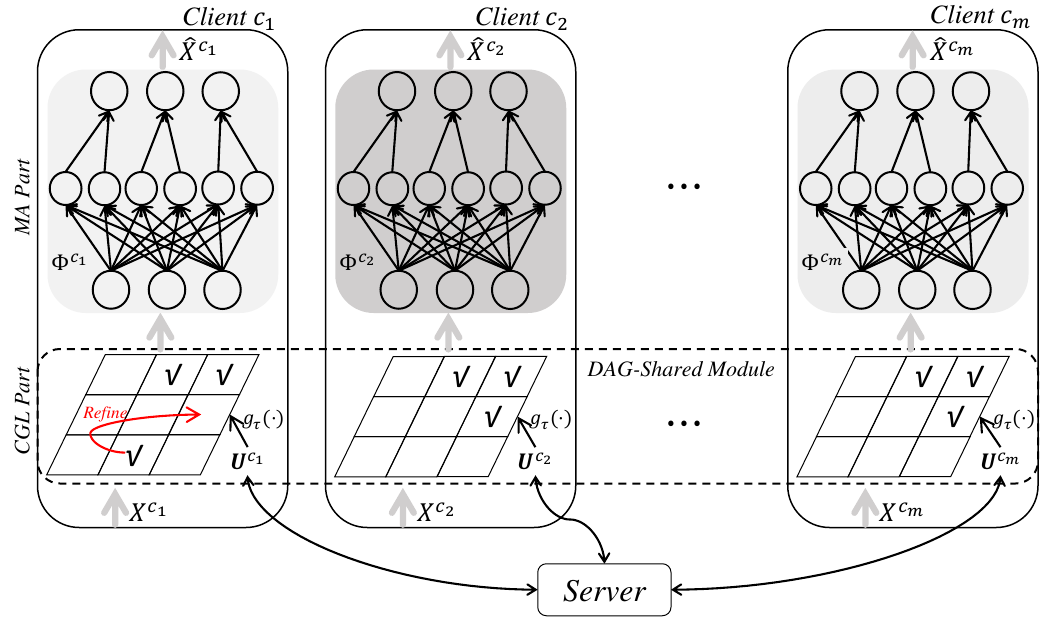}
\end{center}
\caption{An overview of FedDAG. Each solid-line box includes the local model for each client. For client $c_k$, the GSL part includes a continuous proxy $U^{c_k}$ and $g_{\tau}(\cdot)$, the Gumbel-Sigmoid function, which maps $U^{c_k}$ to approximate the binary adjacency matrix. To approximate the mechanisms, the MA part uses $\Phi^{c_k}$, including $d$ neural networks. $X^{c_k}$ represents observations on $c_k$ and $\hat{X}^{c_k}$ is the predicted data. $X^{c_k}$ firstly goes through the GSL part to select the parental variables and then the MA part to get $\hat{X}^{c_k}$. The server coordinates the FL procedures by leveraging $\bm U$ among clients.}
\label{fig:overview}
\end{figure*}

To solve the federated DAG structure learning problem, we formulate a continuous score-based method named FedDAG to learn the DAG structure from decentralized data. Firstly, we define an objective function that guides all models from different clients to federally learn the underlying DAG structure $\mathcal{G}$ (or adjacency matrix $\bm B$), and at the same time also to learn personalized mechanisms for each client. As shown in Figure~\ref{fig:overview}, for each client $c_{k}$, the local model consists of a graph learning part and a mechanisms approximation part. The GSL part is parameterized by a matrix $\bm U^{c_k} \in \mathbb{R}^{d\times d}$, which would be the same for all clients finally\footnote{Please notice that GSL parts of different clients may not be the same during the training procedure. So we index them.}. To make every entry of $\bm U^{c_k}$ to approximate the binary entry of the adjacency matrix, a Gumbel-Sigmoid method~\citep{jang2016categorical, ng2022masked}, represented as $g_{\tau}(\bm U^{c_k})$, is further leveraged to transform $\bm U^{c_k}$ to a differentiable approximation of the adjacency matrix. The mechanisms approximation parts $f^{c_k}_{1}, f^{c_k}_{2}, \cdots, f^{c_k}_{d}$ are parameterized by $d$ sub-networks, each of which has $d$ inputs and one output. In the learning procedure, the GSL parts (specifically $\bm U^{c_k}$) of participating clients are shared with the server $s$. Then, the processed information is broadcast to each client for self-updating its matrix. The details of our method are demonstrated in the following subsections.

\subsection{The overall learning objective}
Now we present the overall learning objective of FedDAG as the following optimization problem:
\begin{equation}
\label{eq:FCD}
\begin{aligned}
\mathop{\arg\max}_{\bm \Phi, \bm U}\quad & \sum_{k=1}^{m} \ \mathcal{S}^{c_k} (\mathcal{D}^{c_k}, \bm \Phi^{c_k}, \bm U) \\
{\rm subject\ to} \quad & g_{\tau}(\bm U) \in {\rm \textbf{DAGs}} \ \Leftrightarrow \ h(\bm U) = {\rm Tr}[e^{(g_{\tau}(\bm U))}] - d = 0,
\end{aligned}
\end{equation}

where $\bm \Phi^{c_k} := \{\bm \Phi^{c_k}_1, \bm \Phi^{c_k}_2,\cdots, \bm \Phi^{c_k}_d\}$ represents the MA part of the model on $c_k$. $\mathcal{S}^{c_k}(\cdot)$ is the scoring function for evaluating the fitness of the local model of client $c_{k}$ and observations $\mathcal{D}^{c_k}$. For score-based DAG structure learning, selecting a proper score function such as BIC score~\citep{schwarz1978estimating}, generalized score function~\citep{huang2018generalized} or equivalently taking the likelihood of $P(X)$ with a penalty function on model parameters~\citep{zheng2018dags, ng2022masked, zheng2020learning, lachapelle2020gradient} can guarantee to identify up the underlying ground-truth graph structure $\mathcal{G}$ because $\mathcal{G}$ is supposed to have the maximal score over Eq.~(\ref{eq:FCD}). Throughout all experiments in this paper, we assume the noise types are Gaussian with equal variance for each local distribution. And the overall score function utilized in this paper is as follows,
\begin{equation}
    \mathcal{S}^{c_k}(\mathcal{D}^{c_k}, \bm \Phi^{c_k}, \bm U^{c_k}) = -\frac{1}{2n_k}\sum_{i=1}^{n_k} \sum_{j}^{d} \Vert \mathcal{D}_{ij}^{c_k} - \bm \Phi^{c_k}_j(g_{\tau}(\bm U^{c_k}_{j,:}) \circ \mathcal{D}_{i}^{c_k}) \Vert_2^2 -\lambda \Vert g_{\tau}(\bm U) \Vert_1.
\end{equation}
In our score function, we take the negative Least Squares loss and a sparsity term, which corresponds to the model complexity penalty in the BIC score~\citep{schwarz1978estimating}.\footnote{The consistency of BIC score for learning graphs on ANMs is discussed in Appendix~\ref{app:bic_score}.} However, the global minimum is hard to reach by using the gradient descent method due to the non-convexity of $h(\bm U)$. More details on discussions of the optimization results can be found in Appendix.~\ref{app:discussion_on_method}.

In this paper, instead of directly taking the likelihood of $P(X)$, we leverage the well-known results on the density transformation to model the distribution of $P(\mathcal{E})$, i.e., maximizing the likelihood $P^{c_k}(\mathcal{E} | \mathcal{F}^{c_k}, \mathcal{G})$ for $\forall c_k \in \mathcal{C}$~\citep{mooij2011causal}. According to Eq.~(\ref{ANM}), we have $\epsilon_i = X_i - f_i({\bf PA}_i)$. That is to say, modeling $P(\mathcal{E})$ can be achieved by an auto-regressive model. To get $\epsilon_i$, the first step is to select the parental set ${\bf PA}_i$ for $X_i$. This can be realized by $\bm B[:, i] \circ X$, where $\circ$ means the element-wise product. In our paper, for client $c_k$, we predict the noise by $\epsilon_i= X_i - \bm \Phi_i(g_{\tau}(\bm U)[:,i] \circ X)$, where $g_{\tau}(\bm U)$ is to approximate $\bm B$ and $\bm \Phi_i(\cdot)$ is parameterized by a neural network to approximate $f_i$. The specific formulation of $\mathcal{S}^{c_k}$ would depend on the assumption of noise distributions.

\subsection{Federated DAG structure learning}
As suggested in NOTEARS~\citep{zheng2018dags}, the hard-constraint optimization problem in Eq.~(\ref{eq:FCD}) can be addressed by an Augmented Lagrangian Method (ALM) to get an approximate solution. Similar to the penalty methods, ALM transforms a constrained optimization problem by a series of unconstrained sub-problems and adds a penalty term to the objective function. ALM also introduces a Lagrangian multiplier term to avoid ill-conditioning by preventing the coefficient of the penalty term from going too large. To solve Eq.~(\ref{eq:FCD}), the sub-problem can be written as
\begin{equation}
\label{eq:sub-problem}
\mathop{\arg\max}_{\bm \Phi, \bm U} \sum_{k=1}^{m} \mathcal{S}^{c_k} \left(\mathcal{D}^{c_k}, \bm \Phi^{c_k}, g_{\tau}(\bm U) \right) - \alpha_t h(\bm U) - \frac{\rho_t}{2} {h(\bm U)}^2,
\end{equation}
where $\alpha_t$ and $\rho_t$ are the Lagrangian multiplier and penalty parameter of the $t$-th sub-problem, respectively. These parameters are updated after the sub-problem is solved. Since neural networks are adopted to fit the mechanisms in our work, there is no closed-form solution for Eq.~(\ref{eq:sub-problem}). Therefore, we solve it approximately via ADAM~\citep{kingma2015adam}. The method is described in Algorithms~\ref{alg:DAG-FCD} and~\ref{alg:SPS}. And in Algorithm~\ref{alg:DAG-FCD}, we share the same coefficients updating strategy as in~\citep{ng2022masked}.

\begin{algorithm}[tb]
\caption{FedDAG}
\label{alg:DAG-FCD}
\begin{algorithmic}[1]
\State{\bfseries Input:} $\mathcal{D}$, $\mathcal{C}$, Parameter-list = \{$\alpha_{init}$, $\rho_{init}$, $h_{tol}$, $it_{max}$, $\rho_{max}$, $\beta$, $\gamma$, $r$ \}
\State{\bfseries Output:} $\mathbb{E}g_{\tau}(\bm U_t), \bm \Phi_{t}$
\State{\textcolor{blue}{\#Parameter Initializing}}
\State{$t\leftarrow 1$, $\alpha_{t}$ $\leftarrow$ $\alpha_{init}$, $\rho_{t}$ $\leftarrow$ $\rho_{init}$}
\While{$t\leq it_{max}$ and $h(\bm U_t)\geq h_{tol}$ and $\rho \leq \rho_{max}$}
\State{\textcolor{blue}{\#Sub-problem Solving}}
\State{$U_{t+1}, \Phi_{t+1} \leftarrow $ SPS($\mathcal{D}, \mathcal{C}, \alpha_t, \rho_t, it_{in}, it_{fl}, r$)}
\State{\textcolor{blue}{\#Coefficients Updating}}
\State{$\alpha_{t+1} \leftarrow \alpha_t + \rho_{t} \mathbb{E}[h(\bm U_{t+1})]$,\quad $t\leftarrow t+1$}
\If{$\mathbb{E}[h(\bm U_{t+1})] > \gamma \mathbb{E}[h(\bm U_{t})]$}
\State{$\rho_{t+1} = \beta \rho_{t}$}
\Else
\State{$\rho_{t+1} = \rho_{t}$}
\EndIf
\EndWhile
\end{algorithmic}
\end{algorithm}

\begin{algorithm}[t]
\caption{Sub-Problem Solver (SPS) for FedDAG}
\label{alg:SPS}
\begin{algorithmic}[1]
\State{\bfseries Input:} $\mathcal{D}$, $\mathcal{C}$, Parameter-list = \{$\alpha_{t}$, $\rho_{t}$, $it_{in}$, $it_{fl}$, $r$\}
\State{\bfseries Output:} $\bm U_{new}$, $\bm \Phi^{it_{in}}$
\State{ Define ${\rm SP}^{c_k} = \mathcal{S}^{c_k} - \alpha_t h(\bm U^{c_k}) - \frac{\rho_t}{2} {h(\bm U^{c_k})}^2$ }
\For{$i$ in ($1,2,\cdots, it_{in}$)}
\For{{\bf each} client $c_k$}
\State{\textcolor{blue}{\#Self-updating}}
\State{$\bm U^{i,c_k}, \bm \Phi^{i,c_k} \leftarrow \mathop{\arg\max}_{\bm \Phi^{c_k}, \bm U^{c_k}} {\rm SP}^{c_k}$ }
\EndFor
\If{$i\ (\% \ it_{fl}) = 0$ or $i=it_{in}$}
\State{\textcolor{blue}{\#Aggregating: randomly select $r$ clients and collect their $\bm U$s into $\mathbb{U}$, then, send $\mathbb{U}$ to the server}}
\State{$\mathbb{U} \leftarrow {\rm Agg}(r, \mathcal{C})$}
\State{\textcolor{blue}{\#Server Updating: average $\bm U \in \mathbb{U}$}}
\State{$\bm U_{new} \leftarrow {\rm Avg}(\mathbb{U})$}
\State{\textcolor{blue}{\#Broadcasting: distribute $U_{new}$ to all clients}}
\State{$\mathcal{C}\leftarrow {\rm BD}(\bm U_{new})$}
\For{{\bf each} client $c_k$}
\State{\textcolor{blue}{\#Clients Updating}}
\State{$\bm U^{i, c_k} \leftarrow \bm U_{new}$}
\EndFor
\EndIf
\EndFor
\end{algorithmic}
\end{algorithm}

Each sub-problem as Eq.~(\ref{eq:sub-problem}) is solved mainly by distributing the computation across all local clients. Since data is prevented from sharing among clients and the server, each client owns its personalized model, which is only trained on its personalized data. The server communicates with clients by exchanging the parameters information of models and coordinates the joint learning task. To achieve so, our method alternately updates the server and clients in each communication round.

\textbf{Client Update.} For each model of client $c_k$, there are two main parts, named GSL part parameterized by $\bm U^{c_k}$ and MA part parameterized by $\bm \Phi^{c_k}$, respectively. Essentially, the joint objective in Eq.~(\ref{eq:sub-problem}) guides the learning process. In the self-updating procedure as described in Algorithm~\ref{alg:SPS}, the clients firstly receive the updated penalty coefficients $\alpha_t$ and $\rho_t$ and the averaged parameter $\bm U_{new}$. Then, the renewed learning personalized score of client $c_k$ is defined as ${\rm SP}^{c_k} = \mathcal{S}^{c_k} - \alpha_t h(\bm U^{c_k}) - \frac{\rho_t}{2} {h(\bm U^{c_k})}^2$. $it_{fl}$ times of local gradient-based parameter updates are operated to maximize its personalized score.

\textbf{Server Update.} After $it_{fl}$ local updates, the server randomly chooses $r$ clients to collect their $\bm U$s to the set $\mathbb{U}$. Then, $\bm U$s in $\mathbb{U}$ are averaged to get $\bm U_{new}$. The other operation on the server is updating the $\alpha_t, \rho_t$ to $\alpha_{t+1}, \rho_{t+1}$. The detailed calculating rules are described at lines $8-14$ in Algorithm~\ref{alg:DAG-FCD}. Then, the new penalty coefficients and parameters are broadcast to all clients. Notice that assuming that data distribution across clients is homogeneous (no distribution shift), $\bm \Phi^{c_k}$ of the chosen $r$ clients can also be collected and averaged to update the local models of clients in the same way, which is named as All-Shared FedDAG (AS-FedDAG) in this paper. For clarity, we name our general method as Graph-Shared (GS-FedDAG) to distinguish it from AS-FedDAG. It is worth noting that AS-FedDAG can further enhance the performance in the homogeneous case but introduce some additional communication costs.

\subsection{Thresholding}
For continuous optimization, as illustrated in the previous work~\citep{ng2022masked}, we leverage Gumbel-Sigmoid to approximate the binary mask. That is to say, the exact $0$ or $1$ is hard to get. The other issue is raised by ALM since the solution of ALM only satisfies the numerical precision of the constraint. This is because we set $h_{tol}$ and $it_{max}$ maximally but not infinite coefficients of penalty terms to formulate the last sub-problem. Therefore, some entries of the output $\bm M = \mathbb{E}g_{\tau}(\bm U)$ will be near but not exactly $0$ or $1$. To alleviate this issue, $\ell_1$ sparsity is added to the objective function. In our method, since all mask values are in $[0, 1]$, we take the median value $0.5$ as the threshold to prune the edges, which follows the same way in our baseline method MCSL~\citep{ng2022masked}. The iterative thresholding method is also taken to deal with the case that the learned graph is cyclic. This may happen if the number of variables is large ($40$ variables in our paper). Because, in numerical optimization, the constraint penalty exponentially decreases with the number of variables. To deal with the cyclic graph, we one-by-one cut the edge with the minimum value until the graph is acyclic. Until now, all continuous search methods for DAG learning suffer from these two problems. It is an interesting future direction to be investigated.

\subsection{Convergence analysis}
Let us quickly review our method. For each client $c_k$, the model parameters include $\bm \Phi^{c_k}$ and $\bm U^{c_k}$. Each client optimizes its parameters on its own data $\mathcal{D}^{c_k}$. Like NOTEARS and its following works, our method can reach a stationary point instead of the global maximum (the ground-truth DAG). Then, we separate our discussion into homogeneous and heterogeneous data.

\subsubsection{Homogeneous data}
For the no distribution shift case, we have $\bm \Phi^{c_1} = \bm \Phi^{c_2}= \cdots = \bm \Phi^{c_m}$ and $\bm U^{c_1} = \bm U^{c_2} = \cdots = \bm U^{c_m}$. Our method named AS-FedDAG (All-Shared FedDAG) sets a central server, which regularly (1) receives all parameters (or $\bm U^{c_k}$ for GS-FedDAG), (2) averages these parameters to get $\bm \Phi^{new}$ and $\bm U^{new}$ and (3) broadcasts $\bm \Phi^{new}$ and $\bm U^{new}$ to all clients during the learning procedures. AS-FedDAG benefits from an advanced technique named FedAvg~\citep{mcmahan2017communication} for solving the FL problem in the homogeneous data case. FedAvg solves a similar problem by averaging all parameters learned from each client in the learning process. 

\subsubsection{Heterogeneous data}
To solve the overall constraint-based problem, we take ALM to convert the hard constraint to a soft constraint with a series of increasing penalty co-efficiencies. The convergence of ALM for the non-convex problem has been well studied~\citep{Nemirovski99optimization} and presented in NOTEARS~\citep{zheng2018dags}. Thus, we only consider the convergence analysis of our method directly from the inner optimization, i.e., the $t$-th sub-problem, as follows.
\begin{equation}
\mathop{\arg\max}_{\bm \Phi, \bm U} \sum_{k=1}^{m} \mathcal{S}^{c_k} \left(\mathcal{D}^{c_k}, \bm \Phi^{c_k}, g_{\tau}(\bm U) \right) - \alpha_t h(\bm U) - \frac{\rho_t}{2} {h(\bm U)}^2.
\end{equation}
Here, for simplification, we just define that $\hat{\mathcal{S}}^{c_k}(\bm \Phi^{c_k}, \bm U) = - \mathcal{S}^{c_k} \left(\mathcal{D}^{c_k}, \bm \Phi^{c_k}, g_{\tau}(\bm U) \right) + \alpha_t h(\bm U) + \frac{\rho_t}{2} {h(\bm U)}^2$. Then, the overall optimization problem can be reformulated as follows.
\begin{equation}
    \mathop{\arg\min}_{\bm \Phi, \bm U} \hat{\mathcal{S}}(\bm U, \bm \Phi) := \sum_{k=1}^{m} \hat{\mathcal{S}}^{c_k}(\bm U, \bm \Phi^{c_k}).
\end{equation}
Through the following part, we use $\nabla_{\boldsymbol{U}}$ and $\nabla_{\boldsymbol{\Phi}}$ to represent the gradients of $\hat{\mathcal{S}}(\bm U, \bm \Phi)$ w.r.t $\boldsymbol{U}$ and $\boldsymbol{\Phi}^{c_k}$, respectively. And, we use $\tilde{\nabla}_{\boldsymbol{U}}$ and $\tilde{\nabla}_{\boldsymbol{\Phi}}$ to represent the stochastic gradients calculated by a mini-batch of observations w.r.t $\boldsymbol{U}$ and $\boldsymbol{\Phi}^{c_k}$, respectively.

\begin{definition}
\label{ass:pgd}
(Partial Gradient Diversity). The gradient diversity among all local learning objectives as:
\begin{equation}
\sum_{i=1}^{m} \left\| \nabla_{\boldsymbol{U}} \hat{\mathcal{S}}^{c_k}(\bm U, \bm \Phi^{c_k}) - \nabla_{\boldsymbol{U}} \hat{\mathcal{S}}(\bm U, \bm \Phi) \right\|^2 \leq \delta^2.
\end{equation}
\end{definition}
Note that the notation of gradient diversity is introduced~\citep{yin2018gradient, haddadpour2019convergence} as a measurement to compute the similarity among gradients updated on different clients. 

\begin{assumption}
\label{ass:smooth}
(Smoothness and Lower Bound). The local objective function $\hat{\mathcal{S}}^{c_k}(\cdot)$ of the $k$-th client is differentiable for $\forall k \in [m]$. Also, $\nabla_{\boldsymbol{U}} \hat{\mathcal{S}}^{c_k}(\bm U, \bm \Phi^{c_k})$ is $L_{\boldsymbol{U}}$-Lipschitz w.r.t $\boldsymbol{U}$ and $L_{\boldsymbol{U}\boldsymbol{\Phi}}$ w.r.t $\boldsymbol{\Phi}^{c_k}$, and $\nabla_{\boldsymbol{\Phi}} \hat{\mathcal{S}}^{c_k}(\bm U, \bm \Phi^{c_k})$ is $L_{\boldsymbol{\Phi}}$-Lipschitz w.r.t $\boldsymbol{\Phi}^{c_k}$ and $L_{\boldsymbol{\Phi}\boldsymbol{U}}$ w.r.t $\boldsymbol{U}$. We also assume the overall objective function can be bounded by a constant $\hat{\mathcal{S}}^*$ and denote $\Delta \hat{\mathcal{S}}_0 = \hat{\mathcal{S}}(\boldsymbol{U}_0, \boldsymbol{\Phi}_0) - \hat{\mathcal{S}}^*$.

\end{assumption}
The relative cross-sensitivity of $\nabla_{\boldsymbol{U}} \hat{\mathcal{S}}^{c_k}$ w.r.t $\boldsymbol{\Phi}^{c_k}$ and $\nabla_{\boldsymbol{\Phi}} \hat{\mathcal{S}}^{c_k}$ w.r.t $\boldsymbol{U}$ with the scalar
\begin{equation}
\chi:=\max \left\{L_{\boldsymbol{U} \boldsymbol{\Phi}}, L_{\boldsymbol{\Phi} \boldsymbol{U}}\right\} / \sqrt{L_{\boldsymbol{U}} L_{\boldsymbol{\Phi}}}.
\end{equation}

\begin{assumption}
\label{ass:bound}
(Bounded Local Variance) For each local data $\mathcal{D}^{c_k}, k\in [m]$, we can independently sample a batch of data denoted as $\xi \subset \mathcal{D}^{c_k}$. Then, there exist constant $\delta_{\boldsymbol{U}}$ and $\delta_{\boldsymbol{\Phi}}$ such that
\begin{equation*}
    \begin{aligned}
    & \mathbf{E} \left[ \left\| \tilde{\nabla}_{\boldsymbol{U}} \hat{\mathcal{S}}^{c_k}(\bm \Phi^{c_k}, \bm U) - \nabla_{\boldsymbol{U}} \hat{\mathcal{S}}^{c_k}(\bm U, \bm \Phi^{c_k}) \right\|^2 \right] \leq \sigma_{\boldsymbol{U}}^2, \\
    & \mathbf{E}\left[ \left\| \tilde{\nabla}_{\boldsymbol{\Phi}} \hat{\mathcal{S}}^{c_k}(\bm \Phi^{c_k}, \bm U) - \nabla_{\boldsymbol{\Phi}} \hat{\mathcal{S}}^{c_k}(\bm U, \bm \Phi^{c_k}) \right\|^2 \right] \leq \sigma_{\boldsymbol{\Phi}}^2,
    \end{aligned}
\end{equation*}
\end{assumption}
The bounded variance assumption is a standard assumption on the stochastic gradients~\citep{haddadpour2019convergence, pillutla2022federated}.

\begin{theorem}
\label{thm:convergence}
(Convergence of GS-FedDAG). For GS-FedDAG with all clients involved in the aggregation, for $\forall 0 \leq it \leq T-1$, under Assumptions~\ref{ass:smooth}, \ref{ass:bound} and \ref{ass:pgd}, and the learning rate for the $\boldsymbol{U}$ part is set as $\eta/(L_{\boldsymbol{U}}it_{in})$ and the learning rate for the $\boldsymbol{\phi}$ part is set as $\eta/(L_{\boldsymbol{\phi}}it_{in})$. Then, for $\eta$, depending on the problem parameters, we have

\begin{equation}
\begin{aligned}
    & \frac{1}{T} \sum_{it=0}^{T-1} \left(\frac{1}{L_{\boldsymbol{U}}}\mathbb{E}\left[ \left\| \nabla_{\boldsymbol{U}} \hat{\mathcal{S}}^{c_k}(\bm \Phi^{c_k}_{it}, \bm U_{it}) \right\|^2 \right] + \frac{1}{L_{\boldsymbol{\Phi}}}\mathbb{E}\left[ \frac{1}{m}\sum_{i=1}^{m}\left\| \nabla_{\boldsymbol{U}} \hat{\mathcal{S}}^{c_k}(\bm \Phi^{c_k}_{it}, \bm U_{it}) \right\|^2 \right] \right) \leq \\
    & \frac{(\Delta \hat{\mathcal{S}}_0\sigma^{2}_{\operatorname{FedDAG},1})^{1/2}}{\sqrt{T}} + \frac{(\Delta \hat{\mathcal{S}}_0^2\sigma^{2}_{\operatorname{FedDAG},2})^{1/3}}{T^{2/3}} + \mathcal{O}(\frac{1}{T}).
\end{aligned}
\end{equation}
where we define the effective variance terms
\begin{equation}
\begin{aligned}
& \sigma_{\operatorname{FedDAG}, 1}^{2}=\left(1+\chi^{2}\right)\left(\frac{\sigma_{\boldsymbol{U}}^{2}}{L_{\boldsymbol{U}}}+\frac{\sigma_{\boldsymbol{\Phi}}^{2}}{L_{\boldsymbol{\Phi}}}\right), \\
& \sigma_{\operatorname{FedDAG}, 2}^{2}=\left(1+\chi^{2}\right)\left(\frac{\delta^{2}}{L_{\boldsymbol{U}}}+\frac{\sigma_{\boldsymbol{U}}^{2}}{L_{\boldsymbol{U}}}+\frac{\sigma_{\boldsymbol{\Phi}}^{2}}{L_{\boldsymbol{\Phi}}}\right)\left(1-\frac{1}{it_{in}}\right),
\end{aligned}
\end{equation}
where $it_{in}$ is the total step of one inner loop used in lines $4-21$ in Algorithm~\ref{alg:SPS}.
\end{theorem}
From Theorem~\ref{thm:convergence}, we can see that the gradients $\nabla_{\boldsymbol{U}} \hat{\mathcal{S}}^{c_k}(\bm \Phi^{c_k}_{it}, \bm U_{it})$ w.r.t $\bm U$ and $\nabla_{\boldsymbol{U}} \hat{\mathcal{S}}^{c_k}(\bm \Phi^{c_k}_{it}, \bm U_{it})$ w.r.t $\boldsymbol{\Phi}$ at the $t$-th step can be bounded if we choose a proper $\eta$, which affects the learning rates of the model. 

The proof of Theorem~\ref{thm:convergence} can be borrowed from the proof of Theorem $2$ in \citep{pillutla2022federated}. Notice that, in our theorem, we have assumed that all clients participate in the aggregation for simplification and the conclusion can be easily extended to the general partial participation case.

\subsection{Privacy and costs discussion}
\textbf{Privacy issues of FedDAG.} The strongest motivation of FL is to avoid \textit{personalized raw data leakage}. To achieve this, FedDAG proposes to exchange the parameters for modeling the graph. Here, we argue that the information leakage of local data is rather limited. The server, receiving parameters with client index, may infer some data property. However, according to the data generation model~(\ref{ANM}), the distribution of local data is decided by (1) DAG structure, (2) noise types/strengths, and (3) mechanisms. The gradient information of the shared matrix is decided by (1) the learning objective and (2) model architecture, which are agnostic to the server. Especially for the network part, clients may choose different networks to make the inference more complex. Moreover, suppose the graph structure is also private information for clients. In that case, this problem can be easily solved by selecting a client to serve as the proxy server\footnote{Notice that the DAG structure encoded in the data is not a secret for the data owners (clients).}. The proxy server needs to play two roles, including training its own model and taking on the server's duties. Then, other clients communicate with the proxy server instead of a real server in the communication round. Moreover, the aim of our work, and federated learning in general, is not to provide a full solution to privacy protection. Instead, it is the first step towards this goal, i.e., no data sharing between clients. To further protect privacy, more constraints need to be added to the federated learning framework, such as the prevention of information leakage from gradient sharing, which are studied under the privacy umbrella. To further enhance privacy protection, our method can also include more advanced privacy protection techniques~\citep{wei2020federated}, which would be an interesting work to be investigated.

\textbf{Communication cost.} Since FedDAG requires exchanging parameters between the server and clients. Additional communication costs are raised. In our method, however, we argue that GS-FedDAG only brings rather small additional communication pressures. For the case of $d$ variables, a single communication only exchanges a $d\times d$ matrix twice (sending and receiving). For homogeneous data, which assumes that local data are sampled from the same distribution, one can also transmit the neural network together to further improve the performance since mechanisms are also shared among clients. The trade-off between performances and communication costs can also be controlled by $r$ in Algorithm~\ref{alg:SPS}, i.e., enlarging or reducing $r$. Surprisingly, we find that reducing $r$ does not harm the performance severely (see Table~\ref{tab:results_client_selection} in Appendix~\ref{app:sup_exp} for detailed results). Moreover, the partial communication method, which only chooses some clients to exchange training information, is also leveraged to address the issue that not all clients are always online at the same time.

\section{Experimental Results}
\label{experiment}
In this section, we study the empirical performances of FedDAG on both synthetic and real-world data. More detailed ablation experiments can also be found in Appendix~\ref{app:sup_exp}.

\textbf{Baselines} We compare our method with various baselines including some continuous search methods, named NOTEARS \citep{zheng2018dags}, NOTEARS-MLP (N-S-MLP, for short) \citep{zheng2020learning}, DAG-GNN \citep{yu2019dag} and MCSL \citep{ng2022masked}, and also two traditional combinatorial search methods named PC \citep{spirtes2001causation} and GES \citep{chickering2002optimal}. The comparison results with another method named causal additive models (CAM)~\citep{buhlmann2014cam} are put in Appendix~\ref{app:cam}. Furthermore, we also include a concurrent work named NOTEARS-ADMM~\citep{ng2022towards}, which also considers learning the Bayesian network in the federated setup. Since NOTEARS-ADMM focuses more on the homogeneous case and linear settings and pays less attention to the nonlinear cases, we only include the results on linear cases of NOTEARS-ADMM in the main paper for fair comparisons. More detailed comparisons are shown in Appendix~\ref{app:comp_admm}. Moreover, we also compare our FedDAG with a voting method~\citep{na2010distributed} in Appendix~\ref{app:voting_method}, which also tries to learn DAG from decentralized data. We provide two training ways for these compared methods. The first way named "All data" is using all data to train only one model, which, however, is not permitted in FedDAG since the ban of data sharing in our setting. For the homogeneous data case, the results on this setting can be an \textit{approximate upper bound} of our method but unobtainable. The second one named "Separated data" is separately training each siloed model over its personalized data, of which the performances reported are the average results of all clients.

\textbf{Metrics.}  We report two metrics named Structural Hamming Distance (SHD) and True Positive Rate (TPR) averaged over $10$ random repetitions to evaluate the discrepancies between estimated DAG and the ground-truth graph $\mathcal{G}$. See more details about SHD, and TPR in Appendix~\ref{app:detailed_metric}. Notice that PC and GES can only reach the completed partially DAG (CPDAG, or MEC) at most, which shares the same Skeleton with the ground-truth DAG $\mathcal{G}$. When we evaluate SHD, we just ignore the direction of undirected edges learned by PC and GES. That is to say, these two methods can get SHD $0$ if they can identify the CPDAG. The implementation details of all methods are given in Appendix~\ref{app:implement}.

\subsection{Synthetic data}
The synthetic data we consider here is generated from Gaussian ANMs (Model~(\ref{ANM})). Two random graph models named Erd{\H{o}}s-R{\'e}nyi (ER) and Scale-Free (SF) (detailed definitions are shown in Appendix~\ref{app:graph_type}.) are adopted to generate the graph structure $\mathcal{G}$. And then, for each node $V_i$ corresponding to $X_i$ in $\mathcal{G}$, we sample a function from the given function sets to simulate $f_i$. Finally, data are generated according to a specific sampling method. In the following experiments, we take $10$ clients and each with $600$ observations (unless otherwise specified in some ablation studies.) throughout this paper. According to Assumption~\ref{ass:sameDAG}, data across all clients share the same DAG structure for both homogeneous and heterogeneous data settings. Due to the space limit, more ablation experiments, such as \textit{unevenly distributed observations}, \textit{varying clients}, \textit{dense graph}, \textit{different non-linear functions}, and \textit{different number of observations}, etc., are put in Appendix~\ref{app:sup_exp}. All detailed discussions on the experimental results are in Appendix~\ref{app:result_discussions}.

\subsubsection{Homogeneous data setting}
\textbf{Results on linear models.} \ For a fair comparison, here, we also provide the linear version of our method. Since linear data are parameterized with an adjacency matrix, we can directly take the adjacency matrix as our model instead of a GSL part and a MA part. During training, the matrix is communicated and averaged by the server to coordinate the joint learning procedures.
\begin{table}[h]
\centering
\caption{Results on the linear model (Homogeneous data).}
\label{tab:iidlinear}
\resizebox{0.95\textwidth}{!}{
\begin{tabular}{cllllllllllll}
\toprule
  ~& \multicolumn{2}{c}{ER2 with 10 nodes} & \multicolumn{2}{c}{SF2 with 10 nodes} & \multicolumn{2}{c}{ER2 with 20 nodes} & \multicolumn{2}{c}{SF2 with 20 nodes} \\
    \cmidrule(lr){2-3} \cmidrule(lr){4-5} \cmidrule(lr){6-7} \cmidrule(lr){8-9}
 ~& \multicolumn{1}{c}{SHD $\downarrow$} & \multicolumn{1}{c}{TPR $\uparrow$} & \multicolumn{1}{c}{SHD $\downarrow$} & \multicolumn{1}{c}{TPR $\uparrow$}
    & \multicolumn{1}{c}{SHD $\downarrow$} & \multicolumn{1}{c}{TPR $\uparrow$} & \multicolumn{1}{c}{SHD $\downarrow$} & \multicolumn{1}{c}{TPR $\uparrow$}
\\ \midrule
PC-All   &  9.0\,$\pm$\,3.9  & 0.58\,$\pm$\,0.14   & 4.4\,$\pm$\,1.3   &  0.76\,$\pm$\,0.07
    &  18.2\,$\pm$\,5.9 & 0.59\,$\pm$\,0.12 & 22.3\,$\pm$\,4.8 & 0.48\,$\pm$\,0.08 \\
GES-All   &  7.5\,$\pm$\,10.1  & 0.82\,$\pm$\,0.25   & 4.1\,$\pm$\,5.6   &  0.89\,$\pm$\,0.14
    &  25.2\,$\pm$\,22.1 & 0.81\,$\pm$\,0.16 & 22.1\,$\pm$\,11.8 & 0.74\,$\pm$\,0.15 \\
NOTEARS-All   &  1.6\,$\pm$\,1.6  & 0.93\,$\pm$\,0.06   & \textit{1.4\,$\pm$\,1.1}   &  \textit{0.92\,$\pm$\,0.05}
    &  \textit{3.0\,$\pm$\,2.7} & \textit{0.94\,$\pm$\,0.06} & \textit{6.9\,$\pm$\,7.0} & \textit{0.86\,$\pm$\,0.12} \\
NOTEARS-Sep & 3.0\,$\pm$\,2.2  &  0.85\,$\pm$\,0.08  & 3.6\,$\pm$\,2.1   & 0.83\,$\pm$\,0.10
    &  4.1\,$\pm$\,2.4  & 0.91\,$\pm$\,0.05 &  10.2\,$\pm$\,5.9 & 0.82\,$\pm$\,0.10 \\
 NOTEARS-ADMM &  4.7\,$\pm$\,3.9  &  0.89\,$\pm$\,0.12  & 4.4\,$\pm$\,3.0   & 0.86\,$\pm$\,0.09
    & 7.9\,$\pm$\,5.9  &  0.89\,$\pm$\,0.07  & 10.7\,$\pm$\,5.3   & 0.82\,$\pm$\,0.08 \\
\midrule
AS-FedDAG &  \textbf{1.3\,$\pm$\,1.5}  &  \textbf{0.94\,$\pm$\,0.07}  & \textbf{1.6\,$\pm$\,1.0}   & \textbf{0.91\,$\pm$\,0.06}
    & \textbf{3.9\,$\pm$\,3.1} & 0.91\,$\pm$\,0.06  & \textbf{9.4\,$\pm$\,6.7}   & 0.82\,$\pm$\,0.12 \\
\bottomrule
\end{tabular}}
\end{table}

NOTEARS-ADMM is also a DAG structure learning method from decentralized data. Different from our averaging strategy to exchange training information among clients, the optimization problem is solved by the alternating direction method of multipliers (ADMM). From Table~\ref{tab:iidlinear}, we find that our method can consistently show its advantage in the linear case. In the \textit{ER2 with 10 nodes} setting, our AS-FedDAG is even better than NOTEARS with all training data. While it is possible and the detailed explanation can be found in Appendix~\ref{app:result_discussions}.

\textbf{Results on the nonlinear model.} \ For the nonlinear setting, all data are generated by an ANM and divided into $10$ pieces. Each $f_i$ is sampled from a Gaussian Process (GP) with RBF kernel of bandwidth one (See Table~\ref{tab:results_sem_type_10nodes} and Table~\ref{tab:results_sem_type_20nodes} in Appendix.~\ref{app:sup_exp} for results of other functions.) and noises are sampled from one zero-mean Gaussian distribution with fixed variance. We consider graphs of $d$ nodes and $2d$ expected edges. 
\begin{table*}[t]
\centering
\caption{Results on the nonlinear ANM with GP (Homogeneous data).}
\label{tab:results_gp}
\resizebox{0.95\textwidth}{!}{
\begin{tabular}{clllllllllllll}
\toprule
~&  ~& \multicolumn{2}{c}{ER2 with 10 nodes} & \multicolumn{2}{c}{SF2 with 10 nodes} 
    & \multicolumn{2}{c}{ER2 with 40 nodes} & \multicolumn{2}{c}{SF2 with 40 nodes} \\
    \cmidrule(lr){3-4} \cmidrule(lr){5-6} \cmidrule(lr){7-8} \cmidrule(lr){9-10}
~& ~& \multicolumn{1}{c}{SHD $\downarrow$} & \multicolumn{1}{c}{TPR $\uparrow$} & \multicolumn{1}{c}{SHD $\downarrow$} & \multicolumn{1}{c}{TPR $\uparrow$}
    & \multicolumn{1}{c}{SHD $\downarrow$} & \multicolumn{1}{c}{TPR $\uparrow$} & \multicolumn{1}{c}{SHD $\downarrow$} & \multicolumn{1}{c}{TPR $\uparrow$}
\\ \midrule
\multirow{6}{*}{\rotatebox{90} {All data}} 
& PC &    15.3\,$\pm$\,2.6  & 0.37\,$\pm$\,0.10 &   14.1\,$\pm$\,4.3   &   0.44\,$\pm$\,0.20
    &    84.9\,$\pm$\,13.4 & 0.40\,$\pm$\,0.08 &  95.0\,$\pm$\,10.4 & 0.36\,$\pm$\,0.07 \\
& GES   &    13.0\,$\pm$\,3.9  & 0.50\,$\pm$\,0.18 &   9.6\,$\pm$\,4.4   &   0.71\,$\pm$\,0.17
    &    59.0\,$\pm$\,9.8 & 0.53\,$\pm$\,0.08 &  73.8\,$\pm$\,11.9 & 0.47\,$\pm$\,0.10 \\
& NOTEARS    &    16.5\,$\pm$\,2.0  & 0.05\,$\pm$\,0.04   & 14.5\,$\pm$\,1.1   &  0.09\,$\pm$\,0.07
    &  71.2\,$\pm$\,7.2 & 0.08\,$\pm$\,0.03 & 70.8\,$\pm$\,2.3& 0.07\,$\pm$\,0.03 \\
& N-S-MLP  & 8.1\,$\pm$\,3.8  & 0.56\,$\pm$\,0.17   & 8.3\,$\pm$\,2.8   &  0.51\,$\pm$\,0.16
    &  45.3\,$\pm$\,6.8 & 0.43\,$\pm$\,0.08 & 49.2\,$\pm$\,7.7& 0.39\,$\pm$\,0.09 \\
& DAG-GNN   &    16.2\,$\pm$\,2.1  & 0.07\,$\pm$\,0.06 &   15.2\,$\pm$\,0.8   &   0.05\,$\pm$\,0.05
    &    73.0\,$\pm$\,7.7 & 0.06\,$\pm$\,0.03 &  72.4\,$\pm$\,1.6 & 0.05\,$\pm$\,0.02 \\
& MCSL & 1.9\,$\pm$\,1.5  & \textit{0.90\,$\pm$\,0.08}   & \textit{1.6\,$\pm$\,1.2}  &  \textit{0.91\,$\pm$\,0.07} &  \textit{25.4\,$\pm$\,13.1} & \textit{0.68\,$\pm$\,0.14} & 31.6\,$\pm$\,10.0 &  \textit{0.59\,$\pm$\,0.13} \\ \midrule
\multirow{6}{*}{\rotatebox{90} {Sep  data}} & PC&    14.1\,$\pm$\,2.4  & 0.31\,$\pm$\,0.06 &   13.6\,$\pm$\,2.7   &   0.30\,$\pm$\,0.10
    &    83.8\,$\pm$\,7.4 & 0.24\,$\pm$\,0.03 &  86.1\,$\pm$\,4.6 & 0.23\,$\pm$\,0.04 \\
& GES   &   12.7\,$\pm$\,2.7  & 0.37\,$\pm$\,0.09 &   12.7\,$\pm$\,2.4   &   0.33\,$\pm$\,0.11
    &    71.0\,$\pm$\,6.7 & 0.29\,$\pm$\,0.03 &  73.2\,$\pm$\,4.4 & 0.29\,$\pm$\,0.05 \\
& NOTEARS &  16.5\,$\pm$\,2.0  & 0.06\,$\pm$\,0.04 &   14.6\,$\pm$\,1.0   &   0.09\,$\pm$\,0.06
    &    71.1\,$\pm$\,7.3 & 0.08\,$\pm$\,0.03 &  70.7\,$\pm$\,2.0 & 0.07\,$\pm$\,0.03 \\
& N-S-MLP  & 8.5\,$\pm$\,2.9  & 0.56\,$\pm$\,0.13 &   8.7\,$\pm$\,2.9   &   0.53\,$\pm$\,0.16
    &    51.0\,$\pm$\,6.9 & 0.41\,$\pm$\,0.06 &  53.6\,$\pm$\,5.5 & 0.39\,$\pm$\,0.08 \\
& DAG-GNN  &   15.7\,$\pm$\,2.3  & 0.11\,$\pm$\,0.05 &   14.5\,$\pm$\,1.0   &   0.10\,$\pm$\,0.06
    &    71.5\,$\pm$\,7.5 & 0.08\,$\pm$\,0.02 &  70.8\,$\pm$\,1.8 & 0.07\,$\pm$\,0.02 \\
& MCSL &  7.1\,$\pm$\,3.2  & 0.83\,$\pm$\,0.08 &   6.9\,$\pm$\,2.8   &   0.84\,$\pm$\,0.08
    &    77.3\,$\pm$\,19.8 & 0.64\,$\pm$\,0.11 &  72.9\,$\pm$\,16.4 & \textbf{0.58\,$\pm$\,0.13} \\ \midrule
& GS-FedDAG &  \textbf{2.4\,$\pm$\,2.0}  & \textbf{0.86\,$\pm$\,0.13} &   \textbf{2.7\,$\pm$\,2.2}   &   \textbf{0.86\,$\pm$\,0.13}
    &   \textbf{36.5\,$\pm$\,12.1} & \textbf{0.65\,$\pm$\,0.15} & \textbf{46.4\,$\pm$\,10.4} & 0.57\,$\pm$\,0.13 \\
& AS-FedDAG &  \textbf{1.8\,$\pm$\,2.0}  & \textbf{0.89\,$\pm$\,0.12} &   \textbf{2.5\,$\pm$\,2.7}   &   \textbf{0.85\,$\pm$\,0.15}
    &   \textbf{30.0\,$\pm$\,12.3} & \textbf{0.74\,$\pm$\,0.15} &  \textbf{31.5\,$\pm$\,10.0} & \textbf{0.59\,$\pm$\,0.13} \\
\bottomrule
\end{tabular}}
\end{table*}

Experimental results are reported in Table \ref{tab:results_gp} with nodes $10$ and $40$. Since all local data are homogeneous, here, we also provide another effective training method named AS-FedDAG, in which the MA parts are also shared among clients. In all settings, AS-FedDAG shows a better performance than GS-FedDAG because more model information is shared during training. While GS-FedDAG can also show a consistent advantage over other methods. When separately training local models, all models suffer from data scarcity. Therefore, we can observe that both GS-FedDAG and AS-FedDAG perform better than the other methods in the fashion of separate training. NOTEARS and DAG-GNN, as continuous search methods, obtain unsatisfactory results due to the weak model capacity and improper model assumption. In contrast, the BIC score of GES gets a linear-Gaussian likelihood, which is incapable to deal with non-linear data\footnote{Please find the ablation experiment with linear data and more discussions of the experimental results in Appendix~\ref{app:sup_exp}.}.  With the number of nodes increasing, GS-FedDAG still shows better results than the closely-related baseline method MCSL. However, NOTEAES-MLP can show a comparable result with GS-FedDAG owing to the advantage over MCSL. 

Here, we give a more detailed explanation of why our FedDAG method performs better than the baseline methods. For PC and GES, they can only reach the CPDAG (or MEC) at most, which shares the same skeleton with the ground-truth DAG. When we evaluate the SHD, we just ignore the direction of undirected edges learned by PC and GES. That is to say, these two methods can get SHD $0$ if they can identify the true CPDAG. Therefore, the final results are not caused by unfair comparison. For PC, the independence test is leveraged to decode the (conditional) independence from the data distribution. Therefore, the accuracy would be affected by (1) the number of observations and (2) the effectiveness of \textit{the non-parametric kernel independence test} method. GES leverages greedy search with BIC score. However, the likelihood part of BIC in GES is Linear Gaussian, which is unsuitable for data generated by the Non-linear model. NOTEARS is a linear model but the mechanisms are non-linear. The reason will be the unfitness between data and model. Therefore, the comparisons with GES and NOTEARS on linear homogeneous data are implemented in the Table~\ref{tab:iidlinear}. DAG-GNN is also a non-linear model. However, the non-linear assumption of DAG-GNN is not the same as the data generation model ANMs assumed in our paper. The second reason comes from its \textit{mechanisms approximation} modules are compulsory to share some parameters. Both NOTEARS-MLP and MCSL have their advantages. Please refer to Tables \ref{tab:results_sem_type_10nodes} and \ref{tab:results_sem_type_20nodes}, you will find that NOTEARS-MLP performs better when the non-linear functions are MIM and MLP while MCSL works better on GP and GP-add models.

\begin{figure*}[h]
\centering
\subfloat{
  \includegraphics[height=0.15\textwidth]{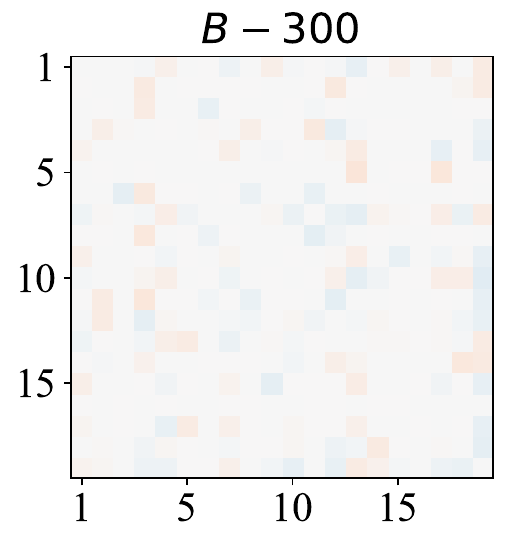}
}
\subfloat{
  \includegraphics[height=0.15\textwidth]{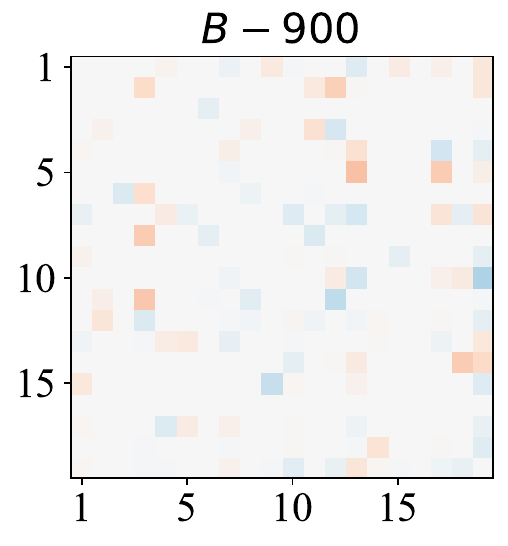}
}
\subfloat{
  \includegraphics[height=0.15\textwidth]{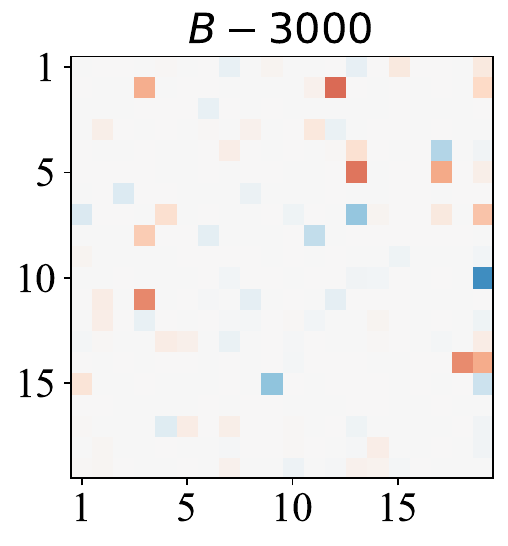}
}
\subfloat{
  \includegraphics[height=0.15\textwidth]{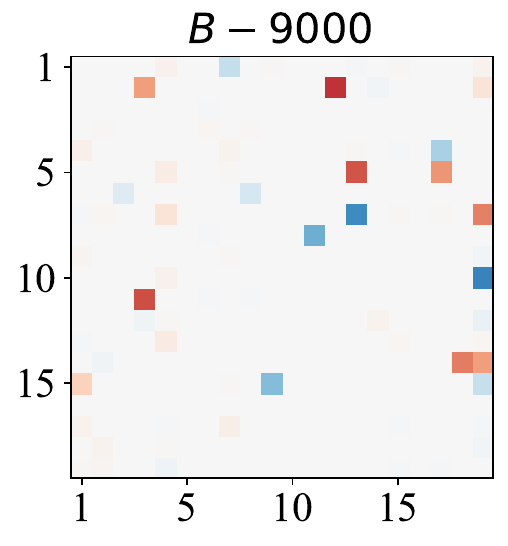}
}
\subfloat{
  \includegraphics[height=0.15\textwidth]{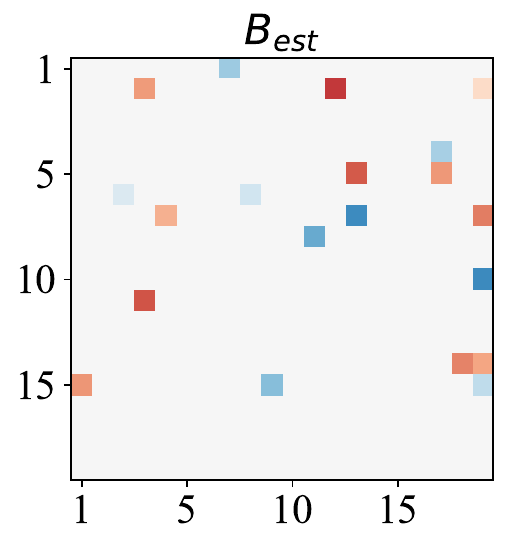}
}
\subfloat{
  \includegraphics[height=0.15\textwidth]{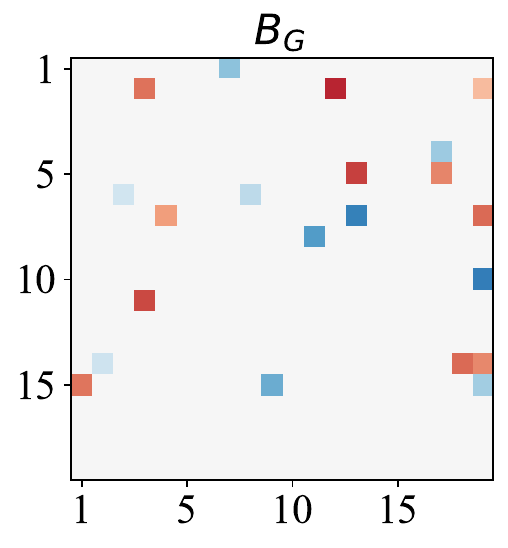}
}
\caption{Visualization of the learned graph during the optimization process. $B-{n}$ means the learned graph in the $n$ steps. $B_{est}$ is the final estimated DAG. $B_G$ is the ground-truth DAG.}
\label{fig:optz_process}
\vskip -0.12in
\end{figure*}

\begin{figure*}[h]
\centering
\includegraphics[width=0.98\textwidth]{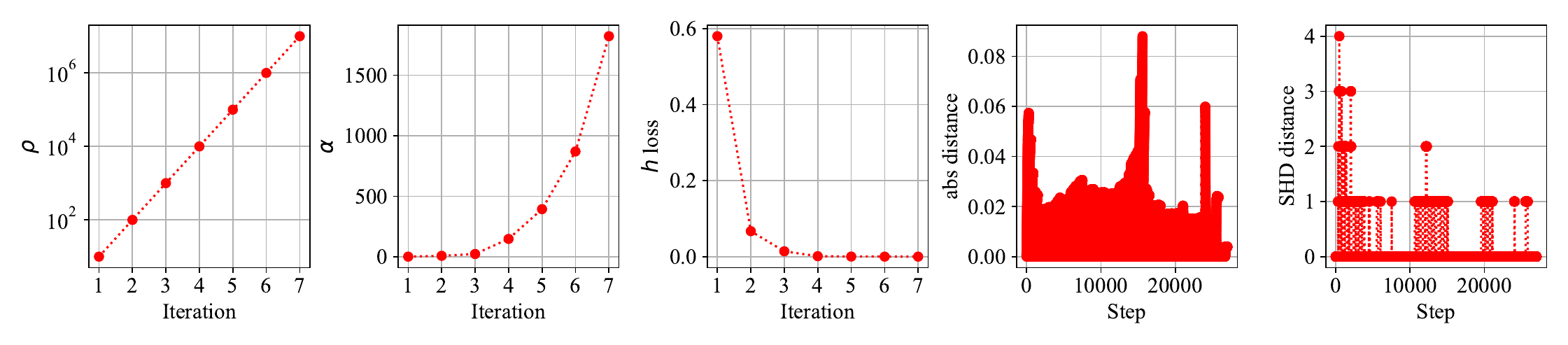}
\caption{Parameters changing during the optimization process. The first three sub-figures include the changes of penalty coefficients $\rho$, $\alpha$, and the DAG constraint loss $h_{loss}$. The fourth sub-figure records the $\ell_1$ distance between two learned graphs on the different clients. The fifth sub-figure records the SHD distance between two learned graphs on the different clients.}
\label{fig:para_change}
\vskip -0.12in
\end{figure*}

\textbf{Visualization of the learned DAG of FedDAG.} \ We take an example of the AS-FedDAG optimization process on linear Gaussian model with NOTEARS as the baseline method and plot the change of estimated parameters in Fig.~\ref{fig:optz_process} and Fig.~\ref{fig:para_change}. In this example, the number of nodes is set as $10$ and the edges are $10$. The data is simulated by ER graph and evenly assign $200$ observations on two different clients. In Fig.~\ref{fig:optz_process}, we can see that the learned graph is asymptotically approximating the ground-truth DAG $B_G$, including the existence of edges and their weights. From Fig.~\ref{fig:para_change}, we can find that with the increase of the penalty coefficients, $h_{loss}$ decreases quickly. For learned graphs on the different clients, we can see that the SHD distance is smaller during the optimization procedures.

\subsubsection{Heterogeneous data setting}
As defined in Section~\ref{problemdef}, the heterogeneous data property of data across clients come from the changes in mechanisms or the shift of noise distributions. To simulate the heterogeneous data, we first generate a graph structure shared by all clients and then decide the types of mechanisms $f_i^{c_k}$ and noises $\epsilon_i$ for $i\in [d]$ for each client $c_k$. In our experiments, we allow that $f^{c_k}$ can be linear or non-linear for each client. If being linear, $f^{c_k}$ here is a weighted adjacency matrix with coefficients sampled from Uniform $([-2.0,-0.5]\cup[0.5,2.0])$, with equal probability. If being non-linear, $f^{c_k}_i$ is independently sampled from GP, GP-add, MLP, or MIM functions~\citep{yuan2011identifiability}, randomly. Then, a fixed zero-mean Gaussian noise is set to each client with a randomly sampled variance from \{0.8, 1\}. 

We can see that the conclusion of experimental results on the heterogeneous data setting is rather similar to that of the homogeneous data. As can be read from Table~\ref{tab:results_noniid}, GS-FedDAG always shows the best performances across all settings. If taking all data together to train one model using other methods, we can see that data heterogeneity would put great trouble to all compared methods while GS-FedDAG plays pretty well. Here, we also provide the experimental results of AS-FedDAG on this setting. We can find that the model misspecification problem would lead to unsatisfactory results, which motivate us to design the GS-FedDAG. Moreover, GS-FedDAG shows consistently good results with different numbers of observations on each client (see Table~\ref{tab:results_noniid_datanum}). NOTEARS takes second place at the setting of $40$ nodes because there are some linear data among clients, which is also the reason that GS-FedDAG shows lower SHDs on heterogeneous data in Table~\ref{tab:results_noniid} than Table~\ref{tab:results_gp}. Compared with non-linear models, NOTEARS easily fits well with even fewer linear data.

\begin{table*}[tb]
\centering
\caption{Results on ANMs with heterogeneous data.}
\label{tab:results_noniid}
\resizebox{0.95\textwidth}{!}{
\begin{tabular}{clllllllllllll}
\toprule
~&  ~& \multicolumn{2}{c}{ER2 with 10 nodes} & \multicolumn{2}{c}{SF2 with 10 nodes} 
    & \multicolumn{2}{c}{ER2 with 40 nodes} & \multicolumn{2}{c}{SF2 with 40 nodes} \\
    \cmidrule(lr){3-4} \cmidrule(lr){5-6} \cmidrule(lr){7-8} \cmidrule(lr){9-10}
~& ~& \multicolumn{1}{c}{SHD $\downarrow$} & \multicolumn{1}{c}{TPR $\uparrow$} & \multicolumn{1}{c}{SHD $\downarrow$} & \multicolumn{1}{l}{TPR $\uparrow$}
    & \multicolumn{1}{c}{SHD $\downarrow$} & \multicolumn{1}{c}{TPR $\uparrow$} & \multicolumn{1}{c}{SHD $\downarrow$} & \multicolumn{1}{c}{TPR $\uparrow$}
\\ \midrule
\multirow{7}{*}{\rotatebox{90} {All   data}} & PC &  22.3\,$\pm$\,4.2  & 0.41\,$\pm$\,0.11 &   21.0\,$\pm$\,3.6   &   0.41\,$\pm$\,0.12
    &    151.9\,$\pm$\,14.2  & 0.27\,$\pm$\,0.08 &  152.5\,$\pm$\,5.4 & 0.26\,$\pm$\,0.04 \\
& GES   &  26.4\,$\pm$\,6.2  & 0.53\,$\pm$\,0.14 &   25.4\,$\pm$\,4.6   &   0.54\,$\pm$\,0.13
    &  NaN & NaN & NaN & NaN \\
& NOTEARS    & 20.4\,$\pm$\,4.1  & 0.49\,$\pm$\,0.14 &   18.7\,$\pm$\,3.3   &   0.45\,$\pm$\,0.11
    &    164.8\,$\pm$\,47.4 & 0.39\,$\pm$\,0.07 &  178.1\,$\pm$\,33.0 & 0.40\,$\pm$\,0.10 \\
& N-S-MLP  &  22.8\,$\pm$\,5.0  & 0.87\,$\pm$\,0.07 &   24.7\,$\pm$\,3.3   &   0.88\,$\pm$\,0.07
    &    344.4\,$\pm$\,71.9 & \textit{0.92\,$\pm$\,0.08} &  325.0\,$\pm$\,50.2 & \textit{0.85\,$\pm$\,0.08} \\
& DAG-GNN    &  21.2\,$\pm$\,6.0  & 0.39\,$\pm$\,0.11 &   16.6\,$\pm$\,3.0   &   0.48\,$\pm$\,0.18
    &    146.6\,$\pm$\,41.6 & 0.29\,$\pm$\,0.08 &  168.2\,$\pm$\,34.2 & 0.31\,$\pm$\,0.09 \\
& MCSL &  19.4\,$\pm$\,4.4  & 0.75\,$\pm$\,0.19 &   19.0\,$\pm$\,4.0   &   0.81\,$\pm$\,0.14
    &    118.6\,$\pm$\,18.1 & 0.68\,$\pm$\,0.11 &  126.9\,$\pm$\,16.5 & 0.59\,$\pm$\,0.12 \\ \midrule
\multirow{7}{*}{\rotatebox{90} {Sep  data}} & PC&   12.5\,$\pm$\,2.7  & 0.45\,$\pm$\,0.07 &   11.0\,$\pm$\,2.1   &   0.49\,$\pm$\,0.07
    &    65.7\,$\pm$\,11.0 & 0.43\,$\pm$\,0.06 &  73.7\,$\pm$\,5.5 & 0.36\,$\pm$\,0.05 \\
& GES   &   12.9\,$\pm$\,2.6  & 0.58\,$\pm$\,0.07 &   10.3\,$\pm$\,2.8   &   0.60\,$\pm$\,0.09
    &    68.2\,$\pm$\,20.8 & 0.65\,$\pm$\,0.09 &  77.2\,$\pm$\,13.8 & 0.60\,$\pm$\,0.07 \\
& NOTEARS  &  7.6\,$\pm$\,2.6  & 0.60\,$\pm$\,0.11 &   7.6\,$\pm$\,1.8   &   0.58\,$\pm$\,0.09
    &    34.9\,$\pm$\,12.7 & 0.63\,$\pm$\,0.11 &  43.4\,$\pm$\,8.4 & 0.53\,$\pm$\,0.10 \\
& N-S-MLP  &  5.2\,$\pm$\,1.4  & 0.80\,$\pm$\,0.05 &   6.1\,$\pm$\,1.6   &   0.76\,$\pm$\,0.05
    &    46.0\,$\pm$\,10.2 & 0.73\,$\pm$\,0.08 &  56.0\,$\pm$\,9.5 & 0.66\,$\pm$\,0.09 \\
& DAG-GNN  & 8.2\,$\pm$\,2.9  & 0.67\,$\pm$\,0.12 &   8.4\,$\pm$\,2.1   &   0.67\,$\pm$\,0.09
    &    45.7\,$\pm$\,13.5 & 0.64\,$\pm$\,0.11 &  52.7\,$\pm$\,8.4 & 0.60\,$\pm$\,0.11 \\
& MCSL &  9.2\,$\pm$\,1.8  & 0.72\,$\pm$\,0.06 &   8.9\,$\pm$\,2.0   &   0.71\,$\pm$\,0.08
    &    76.1\,$\pm$\,13.7 & 0.53\,$\pm$\,0.09 &  78.1\,$\pm$\,6.3 & 0.47\,$\pm$\,0.07 \\ \midrule
& AS-FedDAG &   3.4\,$\pm$\,1.7  & 0.97\,$\pm$\,0.04 &  2.7\,$\pm$\,1.6   &   0.90\,$\pm$\,0.07
    &  35.9\,$\pm$\,17.0 & 0.84\,$\pm$\,0.09 &  41.8\,$\pm$\,12.6 & \textbf{0.73\,$\pm$\,0.07} \\
& GS-FedDAG &   \textbf{1.9\,$\pm$\,1.6}  & \textbf{0.99\,$\pm$\,0.02} &   \textbf{2.6\,$\pm$\,1.3}  &   \textbf{0.93\,$\pm$\,0.07}
    &  \textbf{24.3\,$\pm$\,10.2} & \textbf{0.86\,$\pm$\,0.09} &  \textbf{33.9\,$\pm$\,10.9} & 0.73\,$\pm$\,0.09 \\
\bottomrule
\end{tabular}}
\end{table*}

\subsection{Real data}
We consider a real public dataset named \textbf{fMRI Hippocampus} \citep{poldrack2015long} to discover the underlying relationships among six brain regions. This dataset records signals from six separate brain regions in the resting state of one person in $84$ successive days and the anatomical structure provide $7$ edges as the ground truth graph (see Figure~\ref{fig:real_data_gt_graph} in (Appendix~\ref{app:sup_exp}). Herein, we separately select $500$ records in each of the $10$ days, which can be regarded as different local data. It is worth noting that though this data does not have a real data privacy problem, we can use this dataset to evaluate the learning accuracy of our method. Here, in Table~\ref{tab:real_data} we show part of the experimental results while others lie in Table~\ref{tab:real_data_supp}). AS-FedDAG shows the best performance on all criteria while GS-FedDAG also performs better than most of the other methods. 

\begin{table*}[!ht]
\centering
\caption{Empirical results on \textbf{fMRI Hippocampus} dataset (Part 1).}
\label{tab:real_data}
\resizebox{0.95\textwidth}{!}{
\begin{tabular}{lllllllll}
\toprule
  & \multicolumn{3}{c}{All data} & \multicolumn{3}{c}{Separate data} & \multirow{2}{*}{GS-FedDAG} & \multirow{2}{*}{AS-FedDAG} \\ \cmidrule(lr){2-4} \cmidrule(lr){5-7} 
 &  \multicolumn{1}{c}{PC} & \multicolumn{1}{c}{NOTEARS} & \multicolumn{1}{c}{MCSL} &  \multicolumn{1}{c}{PC} & \multicolumn{1}{c}{NOTEARS} & \multicolumn{1}{c}{MCSL} \\ \midrule
SHD $\downarrow$ & 9.0\,$\pm$\,0.0 & \textit{5.0\,$\pm$\,0.0}  & 9.0\,$\pm$\,0.6 & 8.7\,$\pm$\,1.3 & 8.0\,$\pm$\,1.9 & 8.3\,$\pm$\,1.7 & \textbf{6.4\,$\pm$\,0.9} &  \textbf{5.0\,$\pm$\,0.0} \\
NNZ & 11.0\,$\pm$\,0.0 & 4.0\,$\pm$\,0.0  & 12.0\,$\pm$\,0.6 & 7.6\,$\pm$\,1.3 & 5.4\,$\pm$\,1.5 & 9.0\,$\pm$\,1.7 & 6.8\,$\pm$\,0.6 & 5.0\,$\pm$\,0.0 \\
TPR $\uparrow$ & 0.43\,$\pm$\,0.00 & 0.29\,$\pm$\,0.00  & \textit{0.44\,$\pm$\,0.04} & 0.26\,$\pm$\,0.11 & 0.19\,$\pm$\,0.18 & \textbf{0.35\,$\pm$\,0.15} & 0.27\,$\pm$\,0.12 & \textbf{0.29\,$\pm$\,0.00} \\
FDR $\downarrow$ & 0.73\,$\pm$\,0.00 & \textit{0.50\,$\pm$\,0.00}  & 0.74\,$\pm$\,0.03 & 0.76\,$\pm$\,0.10 & 0.78\,$\pm$\,0.19 & 0.73\,$\pm$\,0.11 & \textbf{0.72\,$\pm$\,0.11} & \textbf{0.60\,$\pm$\,0.00} \\
\bottomrule
\end{tabular}}
\end{table*}

\section{Related work}

Two mainstreams, named constraint-based and score-based methods, push the development of DAG structure learning. Constraint-based methods, including SGS and PC \citep{spirtes2001causation}, take conditional independence constraints induced from the observed distribution to decide the graph skeleton and part of the directions. Another branch of methods \citep{chickering2002optimal} define a score function, which evaluates the fitness between the distribution and graph, and identifies the graph $\mathcal{G}$ with the highest score after searching the DAG space. To avoid solving the combinatorial optimization problem, NOTEARS \citep{zheng2018dags} introduces an equivalent acyclicity constraint and formulates a fully continuous optimization for searching the graph. Following this work, many works leverages this constraint to non-linear case~\citep{ng2019graph, zheng2020learning, lachapelle2020gradient, zhu2020causal, wang2021ordering, gao2021dag, ng2022masked}, low-rank graph~\citep{fang2020low}, interventional data~\citep{brouillard2020differentiable, ke2019learning, scherrer2021learning}, time-series data~\citep{pamfil2020dynotears}, incomplete data~\citep{gao2022missdag, geffner2022deep} and unmeasured confounding~\citep{bhattacharya2021differentiable}. GOLEM~\citep{ng2020role} leverages the full likelihood and soft constraint to solve the optimization problem. \citet{ng2022convergence}, DAG-NoCurl~\citep{yu2021nocurl} and NOFEARS~\citep{wei2020nofears} focus on the optimization aspect.

The second line of related work is on the Overlapping Datasets (OD)~\citep{danks2009intergrating, tillman2011learning, triantafillou2015constraint, huang2020causal2} problem in DAG structure learning. However, OD assumes that each dataset owns observations of partial variables and targets learning the integrated DAG from multiple datasets. In these works, data from different sites need to be collected on a central server.

The last line is on federated learning~\citep{yang2019federated, kairouz2019advances}, which provides the joint training paradigm to learn from decentralized data while avoiding sharing raw data during the learning process. FedAvg~\citep{mcmahan2017communication} first formulates and names federated learning. FedProx~\citep{li2020federated} studies the heterogeneous case and provides the convergence analysis results. SCAFFOLD leverages variance reduction by correcting client-shift to enhance training efficiency. Besides these fundamental problems in FL itself, this novel learning way has been widely co-operated with or applied to many real-world tasks such as healthcare~\citep{sheller2020federated}, recommendation system~\citep{yang2020federated}, and smart transport~\citep{samarakoon2019distributed}.

\subsection{Concurrent work (NOTEARS-ADMM)}
In NOTEARS-ADMM~\citep{ng2022towards}, the authors also consider the same setting for discovering the underlying relations from distributed data owing to privacy and security concerns. The main advantage of our FedDAG over NOTEARS-ADMM is to handle heterogeneous data, which is very common in real applications. Then, NOTEARS-ADMM mainly considers the linear case, which shares the same learning object with our method. Instead of taking an average to share training information, ADMM is taken to make the adjacency matrix close. More detailed experimental comparisons can be found in Appendix~\ref{app:comp_admm}, from which we can see that our FedDAG performs better in most settings.
\section{Conclusion and Discussions}
\label{conclusion}
Learning the underlying DAG structure from decentralized data brings considerable challenges to traditional DAG learning methods. In this context, we have introduced one of the first federated DAG structure learning methods called FedDAG, which uses a two-level structure for each local model. During the learning procedure, each client tries to learn an adjacency matrix to approximate the graph structure and neural networks to approximate the mechanisms. The matrix parts of some participating clients are aggregated and processed by the server and then broadcast to each client for updating its personalized matrix. The overall problem is formulated as a continuous optimization problem and solved by gradient descent. Structural identifiability conditions are provided, and extensive experiments on various data sets are to show the effectiveness of our FedDAG.

The first limitation of our framework is with the \textit{no latent variable} assumption, which is seldom right in real scenarios. While, as a general framework, the advanced methods~\citep{bhattacharya2021differentiable}, which can handle the no observed confounder case, can be well incorporated with our method to deal with the federated setup (More details can be seen in Appendix \ref{app:as_a_framework}). Another limitation relies on privacy protection. As we said, we focus on the statistical and optimization perspectives of federated DAG structure learning and leave the problem of combining the advanced privacy protection methods~\citep{wei2020federated} into our framework as a future work. The last limitation is to loose the invariant DAG assumption and allow the causal graph change among different clients, which is more common in the real world.
\subsection*{Acknowledgement}

LS is supported by the Major Science and Technology Innovation 2030 “Brain Science and Brain-like Research” key project (No. 2021ZD0201405). EG is supported by an Australian Government Research Training Program (RTP) Scholarship. This research was supported by The University of Melbourne’s Research Computing Services and the Petascale Campus Initiative. This research was undertaken using the LIEF HPC-GPGPU Facility hosted at the University of Melbourne. This Facility was established with the assistance of LIEF Grant LE170100200. TL was partially supported by Australian Research Council Projects DP180103424, DE-190101473, IC-190100031, DP-220102121, and FT-220100318. MG was supported by ARC DE210101624. HB was supported by ARC FT190100374.

\clearpage
\bibliography{ref}
\bibliographystyle{tmlr}

\clearpage
\appendix
\onecolumn
\addcontentsline{toc}{section}{Appendix}
\part{Appendix} 
\parttoc
\newpage
\section{Structure identifiability}
\label{app:identifiability}

Besides exploring effective DAG structure learning methods, identifiability conditions of graph structure~\citep{spirtes2001causation} are also important. In general, unique identification of the ground truth DAG is impossible from purely observational data without some specific assumptions. However, accompanying some specific data generation assumptions, the graph can be identified \citep{peters2011identifiability,peters2014identifiability,zhang2009identifiability,shimizu2006linear,hoyer2008nonlinear}. We first give the definition of identifiability in the  decentralized setting. 
\begin{definition}
Consider a decentralized distribution set $P^\mathcal{C}(X)$ satisfying Assumption \ref{ass:sameDAG}. Then, $\mathcal{G}$ is said to be identifiable if $P^\mathcal{C}(X)$ cannot be induced from any other DAG.
\end{definition}

\begin{condition}
{\rm (Minimality condition)} Given the joint distribution $P(X)$, $P(X)$ is Markovian to a DAG $\mathcal{G}$ but not Markovian to any sub-graph of $\mathcal{G}$. 
\end{condition}

\begin{condition}
\label{con:ranm}
{\rm (Cond. 19 in \citep{peters2014identifiability})} The triple $(f_j, P(X_i), P(\epsilon_j))$ does not solve the following differential equation $\forall x_i$, $x_j$ with $v''(x_j-f(x_i))f'(x_i) \neq 0$: 
\begin{equation*}
    \xi^{\prime \prime \prime}=\xi^{\prime \prime}\left(-\frac{\nu^{\prime \prime \prime} f^{\prime}}{\nu^{\prime \prime}}+\frac{f^{\prime \prime}}{f^{\prime}}\right)-2 \nu^{\prime \prime} f^{\prime \prime} f^{\prime}+\nu^{\prime} f^{\prime \prime \prime}+\frac{\nu^{\prime} \nu^{\prime \prime \prime} f^{\prime \prime} f^{\prime}}{\nu^{\prime \prime}}-\frac{\nu^{\prime}\left(f^{\prime \prime}\right)^{2}}{f^{\prime}}.
\end{equation*}
Here, $f:= f_j$ and $\xi:= \log P(X_i)$, and $v:= \log P(\epsilon_j)$ are the logarithms of the strictly positive densities. 
\end{condition}

\begin{definition}
{\rm (Restricted ANM. Def. 27 in \citep{peters2014identifiability})} Consider an ANM with $d$ variables. This SEM is called restricted ANM if $\forall j\in \mathbb{V}, i\in {\bf PA}_j$ and all sets $\mathbb{S} \subseteq \mathbb{V}$ with $\mathbf{PA}_{j} \backslash\{i\} \subseteq \mathbb{S} \subseteq \mathbf{PA}_{j} \backslash\{i, j\}$, there is an $x_{\mathbb{S}}$ with $P(x_{\mathbb{S}})>0$, s.t. the tripe 
\begin{equation*}
    \left(f_{j}(x_{\mathbf{P A}_{j} \backslash\{i\}}, \underbrace{\cdot}_{X_{i}}), P\left(X_{i} \mid X_{\mathbb{S}}=x_{\mathbb{S}}\right), P\left(\epsilon_{j}\right)\right)
\end{equation*}
satisfies Condition\ref{con:ranm}. Here, the under-brace indicates the input component of $f_j$ for variable $X_i$. In particular, we require the noise variables to have non-vanishing densities and the functions $f_j$ to be continuous and three times continuously differentiable.
\end{definition}

\begin{assumption}
\label{ass:fcdid_condition}
{\rm (Faithfulness)} Let $P^\mathcal{C}(X)$ satisfy Assumption~\ref{ass:sameDAG}. At least one distribution $P^{c_k}(X) \in P^\mathcal{C}(X)$ meets Assumption \ref{ass:id_condition} and the other distributions are faithful to $\mathcal{G}$.
\end{assumption}

\begin{assumption}
\label{ass:id_condition}
Let a distribution $P(X)$ with $X=( X_1, X_2, \cdots, X_d)$ be induced from a restricted ANM \ref{con:ranm} with graph $\mathcal{G}$, and $P(X)$ satisfies \textit{Minimality condition} w.r.t $\mathcal{G}$.
\end{assumption}

\begin{proposition}
\label{prop:identification_conclusion}
Given $P^\mathcal{C}(X)$ satisfying Assumption \ref{ass:fcdid_condition}, and then, $\mathcal{G}$ can be identified up from $P^\mathcal{C}(X)$.
\end{proposition}

Proof. From Remark~\ref{remark:markov}, we have $P^{c_k}(X) \in P^\mathcal{C}(X)$ for $\forall c_{k}$, is Markov with $\mathcal{G}$. For each $c_{k}\in \mathcal{C}$ with $P^{c_k}(X)$ does not satisfy Assumption~\ref{ass:id_condition}, the Completed Partially DAG (CPDAG) $\hat{\mathcal{G}}$~\citep{pearl2009causality}, which represents the CPDAG induced by $\mathcal{G}$, can be identified~\citep{spirtes2001causation}. (1) That also says that these distributions can be induced from any DAG induced from $\mathcal{M}(\mathcal{G})$, including $\mathcal{G}$ definitely. Notice that skeleton($\hat{\mathcal{G}}$) = Skeleton($\mathcal{G}$) and any  $X_i\leftarrow X_j$ in $\hat{\mathcal{G}}$ is also existed in $\mathcal{G}$. Then, for those $c_k$ with with $P^{c_k}(X)$ satisfying Assumption~\ref{ass:id_condition}, $\mathcal{G}$ can be identified. (2) That is to say, distributions satisfying Assumption~\ref{ass:id_condition}  can only be induced from $\mathcal{G}$. Then, two kinds of graph, $\hat{\mathcal{G}}$ and $\mathcal{G}$, are obtained. Therefore, $\mathcal{G}$ can be easily identified. With (1) and (2), $P^{c_k}(X) \in P^\mathcal{C}(X)$ for $\forall c_{k}$ can only be induced by $\mathcal{G}$. Then, $\mathcal{G}$ is said to be identifiable $\hfill\blacksquare$

Future work is to relax our invariant DAG assumption to the invariant CPDAG assumption, which means that a group of DAGs across different clients share the same conditional independence. For this case, the generalized score functions~\citep{huang2018generalized} can be adopted to search for the Markov Equivalence Class. However, it is not straightforward to incorporate this method into our FedDAG framework since the score of this method is motivated by a kernel-based (conditional) independence test rather than penalized likelihood. Moreover, this method does not support a continuous search strategy. It would be interesting to explore the penalized likelihood-based method for this case and incorporate it into our framework.
\section{Implementations}
\label{app:implement}
The comparing DAG structure learning methods used in this paper all have available implementations, listed below:
\begin{itemize}[leftmargin=*]
\item MCSL: Codes are available at \texttt{gCastle} \url{https://github.com/huawei-noah/trustworthyAI/tree/master/gcastle}. The first author of MCSL added the implementation in this package.
\item NOTEARS and NOTEARS-MLP: Codes are available at the first author's GitHub repository \url{https://github.com/xunzheng/notears}
\item NOTEARS-ADMM: Codes are available at the first author's GitHub repository \url{https://github.com/ignavierng/notears-admm}
\item DAG-GNN: Codes are available at the author's GitHub repository \url{https://github.com/fishmoon1234/DAG-GNN}
\item PC and GES: the implementations of PC and GES are available at \texttt{causal-learn} package repository \url{https://causal-learn.readthedocs.io/en/latest/index.html}
\item CAM: the codes are available at \texttt{CRAN R} package repository \url{https://cran.r-project.org/src/contrib/Archive/CAM/}
\end{itemize}
Our implementation is highly based on the existing Tool-chain named \texttt{gCastle}~\citep{zhang2021gcastle}, which includes many gradient-based DAG structure learning methods.

\subsection{Graph generation}
\label{app:graph_type}
To simulate DAG for generating observations, we introduce two kinds of graph generation methods named Erd{\H{o}}s-R{\'e}nyi (ER) and Scale-Free (SF) graphs. To simulate the ER graph generation, we firstly randomly sample a topological order
and by adding directed edges where it is allowed independently with probability $p = \frac{2s}{d^2-d}$ where $s$ is
the number of edges in the resulting DAG. To generate Scale-free (SF) graphs, we first take the
Barabasi-Albert model and then add all nodes one by one. From the above descriptions, we can find that the degree distribution of ER graphs follows a Poisson distribution, and the degree of SF graphs follows a power law: few nodes, often called hubs, have a high degree~\citep{lachapelle2020gradient}.

\subsection{SEM simulation}

In our experimental setup, for each client, we randomly choose a nonlinear type from the given four functions with equal probability in the heterogeneous data setting. The nonlinear function choice is totally the same as used in NOTEARS-MLP~\citep{zheng2020learning}. The details are as follows.

We simulate the SEM $X_j = f_j(X_{pa_j}) + Z_j$ for all $j \in [d]$ in the topological order induce by $G$. 

GP: $f_j$ is drawn from the Gaussian process with RBF kernel with length-scale one.

GP-add: $f_j(X_{pa_j}) = \sum_{k\in pa_j} f_{jk}(X_k)$, where each $f_{jk}$ is from GP.

MLP: $f_j$ is randomly initialized MLP with one hidden layer of size $100$ and Sigmoid activation.

MiM: also named as index model. $f_j(X_{pa_j}) = \sum_{m=1}^3 h_m(\sum_{k\in pa_j} \theta_{jmk}X_k)$, where $h_1$ = tanh, $h_2$ = cos, $h_3$ = sin, and each $\theta_{jmk}$ is uniformly drawn from range $[-2, -0.5] \cup [0.5, 2]$.

\subsection{Detailed metrics}
\label{app:detailed_metric}
SHD is a kind of measurement which is defined to calculate the Hamming distance between two partially directed acyclic graphs (PDAG) by counting the number of edges for which
the edge type differs in both PDAGs. In PDAG, there exist four kinds of edges between two nodes: $i \rightarrow j$, $i \leftarrow j$, $i-j$, and $i\ j$. SHD just counts the different edges between the two graphs.
SHD is defined over the space of PDAGs, so we can, of course, use it to calculate distances in DAG and CPDAG spaces.

True Positive Rate (TPR) and False Discovery Rate (FDR) are two common metrics in the machine learning community. True positive rate, also referred to as sensitivity or recall, is used to measure the percentage of actual positives which are correctly identified. The FDR is defined as the expected proportion of errors committed by falsely rejecting the null hypothesis. Let $TP$ be true positives (samples correctly classified as positive), $FN$ be false negatives (samples incorrectly classified as negative), $FP$ be false positives (samples incorrectly classified as positive), and $TN$ be true negatives (samples correctly classified as negative). Then, $TPR = \frac{TP}{TP+FN}$ and $FDR=\frac{FP}{FP+TP}$.

\subsection{Hyper-parameters setting}
In all experiments, there is no extra hyper-parameter to adjust for PC (with Fisher-z test and $p$-value 0.01) and GES (BIC score). For NOTEARS, NOTEARS-MLP, and DAG-GNN, we use the default hyper-parameters provided in their papers/codes. For MCSL, the hyper-parameters that need to be modified are $\rho_{init}$ and $\beta$. Specifically, if experimental settings ($10$ variables and $20$ variables) are the same as those in their paper, we just take all the recommended hyper-parameters. For settings not implemented in their paper ($40$ variables exactly), we have two kinds of implementations. The first one is taking a linear interpolation for choosing the hyper-parameters. The second one is taking the same parameters as ours. We find that the second choice always works better. In our experiment, we report the experimental results in a second way. Notice that CAM pruning is also introduced to improve the performance of MCSL, which however can not guarantee a better result in our settings. For simplicity and fairness, we just take the direct outputs of MCSL.

Similar to MCSL~\citep{ng2022masked} and GraN-DAG~\citep{lachapelle2020gradient}, we implement several experiments on simulated data with known graph structure to search for the hyper-parameters and then use these hyper-parameters for all the simulated experiments. Specifically, we use seeds from $1$ to $10$ to generate the simulated data to search for the best combination of hyper-parameters while all our experimental results reported in this paper are all conducted using seeds from $2021$ to $2030$.

\subsection{Hyper-parameters in real-data setting}
\label{sec:hp_selection}
Most DAG learning methods have hyper-parameters, more or less, which need to be decided prior to learning. Moreover, NN-based methods are especially sensitive to the selection of hyper-parameters. For instance, Gran-DAG~\citep{lachapelle2020gradient} defines a really large hyper-parameters space for searching the optimal combination, which even uses different learning rates for the first sub-problem and the other sub-problems. MCSL and GS-FedDAG are sensitive to the selection of $\rho_{init}$ and $\beta$ when constructing and solving the sub-problem. As pointed out in~\citep{kairouz2019advances}, NOTEARS focuses more on optimizing the scoring term in the early stage and pays more attention to approximate DAG in the late stage. If NOTEARS cannot find a graph near $\mathcal{G}$ in the early stage, then, it would lead to a worse result.

To alleviate this problem, one may choose to (1) enlarge the learning rate or take more steps when solving the first few sub-problems as Gran-DAG; (2) reduce the value of coefficient $\rho_{init}$ to let the optimizer pay more attention to the scoring term in the early stages as MCSL. The other trick we find when dealing with real data is increasing $\ell_1$. This mostly results from that real data may not fit well with the data generation assumptions in most papers. Therefore, we choose to conduct a grid search to find the best combination of $\rho_{init}, \beta, \ell_1$ for DAG structure learning on real data. 

In the practice of DAG structure learning, it is impossible to have $\mathcal{G}$ to select the hyper-parameters. One common approach is trying multiple hyper-parameter combinations and keeping the one yielding the best score evaluated on a validation set~\citep{koller2009probabilistic,ng2022masked,lachapelle2020gradient}. However, the direct use of this method may not work for some algorithms, such as MCSL, NOTEARS-MLP, and GS-FedDAG. This mainly lies in the similar explanations of the property of the traditional solution of FL. In the late stage of optimization, the optimizer focuses heavily on finding \textit{a DAG} by enlarging the penalty coefficient $\rho$. Then, the learning of relationship mechanisms would be nearly ignored. To address this problem, we first report the DAG directly learned by a combination of hyper-parameters. And then, we replace the parameters part for describing the graph with the learned DAG. Afterward, we just take the score without DAG constraint to optimize the relationship mechanisms approximating part (which may not be the same name in the other algorithms). Finally, the validation set is taken to evaluate the learned model. The final hyper-parameters used on  the real dataset in our paper are as follows:
\begin{table}[ht]
\centering
\caption{The hyper-parameters used on real data.}
\label{tab:Hyper-p_realdata}
\begin{tabular}{l|lll}
\toprule
Parameters & $\rho_{init}$ & $\beta$ & $\lambda_{\ell_1}$ \\ \midrule
Values & $0.008$ & $2$ & $0.3$  \\ \bottomrule
\end{tabular}
\end{table}

\subsection{Model parameters}
The GSL part in each local model is parameterized by a $d\times d$ matrix named $\bm U$ and the Gumbel-Sigmoid approach is leveraged for approximating the binary form. Each entry in $\bm U$ is initialized as $0$. The temperature $\tau$ is set to $0.2$ for all settings. Then, for the relationship mechanism approximating part, we use $4$ dense layers with $16$ variables in each hidden layer. All weights in the Network are initialized using the Xavier uniform initialization. The number of parameters used in each method is shown in Table~\ref{tab:model_para}.
\begin{table}[h]
    \centering
    \caption{Model parameters for each client with different nodes.}
    \label{tab:model_para}
    \begin{tabular}{l|lll}
    \toprule
          & $10$ nodes & $20$ nodes & $40$ nodes  \\ \midrule
         NOTEARS& $100$ & $400$ & $1600$  \\ \midrule
         Mask of MCSL & $100$ & $400$ & $1600$  \\ \midrule
         NN of MCSL & $9930$ & $19860$ & $39720$ \\ \bottomrule
    \end{tabular}
\end{table}

\subsection{Training parameters}
Our GS-FedDAG and AS-FedDAG reach this point and are implemented with the following hyper-parameters. We take ADAM \citep{kingma2015adam} with learning rate $3\times10^{-2}$ and all the observational data $\mathcal{D}^{c_k}$ on each client are used for computing the gradient. And the detailed parameters used in Algorithms \ref{alg:DAG-FCD} and \ref{alg:SPS} are listed in Table \ref{tab:Hyper-p}.
\begin{table}[ht]
\centering
\caption{The hyper-parameters used on simulated data in this paper.}
\label{tab:Hyper-p}
\begin{tabular}{l|llllllll}
\toprule
Parameters & $\alpha_{init}$ & $h_{tol}$ & $it_{max}$ & $it_{inner}$ & $it_{fl}$ & $\gamma$ &$\rho_{max}$ &  $\lambda_{\ell_1}$ \\ \midrule
Values & $0$ & $1\times 10^{-10}$ & $25$ & $1000$ & $200$ & $0.25$ & $1\times 10^{14}$ & $0.01$ \\ \bottomrule
\end{tabular}
\end{table}

Notice that as illustrated in MCSL \citep{ng2022masked}, the performance of the algorithm is affected by the initial value of $\rho_{init}$ and the choice of $\beta$. Since a small initial of $\rho_{init}$ and $\beta$ would result in a rather long training time. As said in~\citep{kaiser2021unsuitability}, MLE plays an important role in the early stage of training and highly affects the final results. Therefore, carefully picking a proper combination of $\rho_{init}$ and $\beta$ will lead to a better result.  In our method, we tune these two parameters via the same scale of experiment with seeds $1\sim 10$. For each variable scale and training type, the parameters are adjusted once and are applied to all other experiments with the same variable scale. We find the combinations of the following parameters in Table \ref{tab:phobeta} work well in our method. Our method also adopts a $\ell_1$ sparsity term on $g_{\tau}(\bm U)$, where the sparsity coefficient $\lambda_{\ell_1}$ is chosen as $0.01$ for all settings.
\begin{table}[ht]
\caption{The combinations of $\rho_{init}$ and $\beta$ on simulated data in our method.}
\label{tab:phobeta}
\centering
\begin{tabular}{llllllll}
\toprule
  & \multicolumn{2}{c}{10 nodes} & \multicolumn{2}{c}{20 nodes} & \multicolumn{2}{c}{40 nodes} \\ \cmidrule(lr){2-3} \cmidrule(lr){4-5} \cmidrule(lr){6-7}
 &  \multicolumn{1}{c}{$\rho_{init}$} & \multicolumn{1}{c}{$\beta$}& \multicolumn{1}{c}{$\rho_{init}$} & \multicolumn{1}{c}{$\beta$} & \multicolumn{1}{c}{$\rho_{init}$} & \multicolumn{1}{c}{$\beta$}\\ \midrule
AS-FedDAG & $6\times 10^{-3}$ & $10$  & $1\times 10^{-5}$ & $20$ & $1\times 10^{-11}$ & $120$ \\
GS-FedDAG &$6\times 10^{-3}$ & $10$  & $6\times 10^{-5}$ & $20$ & $1\times 10^{-11}$ & $120$ \\ \bottomrule
\end{tabular}
\end{table}

\subsection{Sensitivity analysis of hyper-parameters}
\begin{figure}
\begin{center}
    \includegraphics[width=0.9\textwidth]{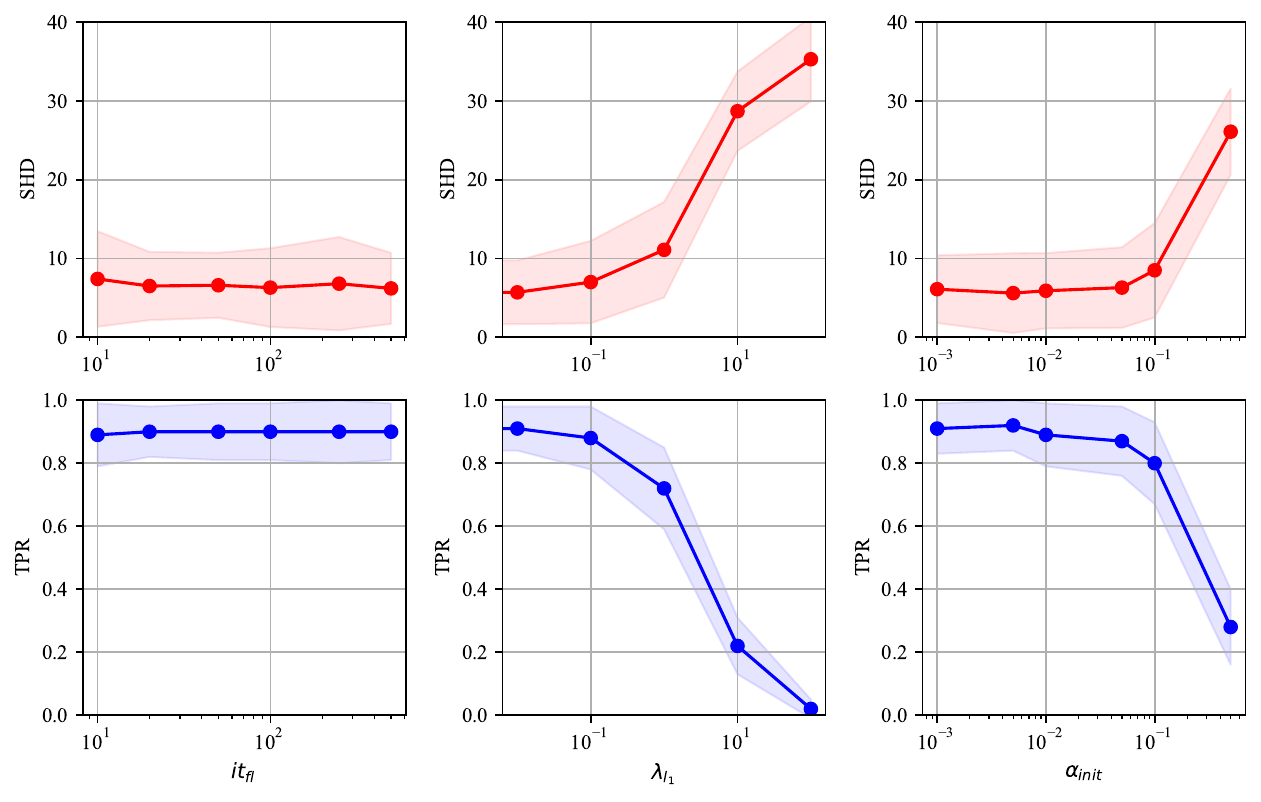}
\end{center}
\caption{The sensitivity analysis of hyper-parameters}
\label{fig:sensitivity}
\end{figure}
Here, we show the sensitivity analysis of $it_{fl}$, $\alpha_{init}$, and $\lambda_{l_1}$. From the experimental results in Figure \ref{fig:sensitivity}, we find that our method is relatively robust to $it_{fl}$. That is to say, the $it_{fl}$ can be reduced to alleviate the pressure of communication costs while the performance can be well kept. $\lambda_{l_1}$ is the coefficient of $l_1$ sparsity, which will affect the final results. Because we have no sparsity information of the underlying graph, we set $\lambda_{l_1}=0.01$ in all settings. When dealing with real data, we recommend the audiences adjust this parameter by using our parameter-tuning method provided in Section \ref{sec:hp_selection}. The results of $\alpha_{init}$ are exactly as expected. As discussed before, our method tries to maximize the likelihood term of the total loss in the early stages, which is important to find the final ground-truth DAG. If setting a relatively large $\alpha_{init}$, the early learning stages would be affected. Therefore, we recommend directly taking $\alpha_{init}$ as $0$ in all settings.
\section{Discussions on our method}
\label{app:discussion_on_method}

\subsection{Novelty and contributions}
Firstly, we acknowledge the contribution of our baseline method MCSL~\citep{ng2022masked}, which performs well in many settings and helps to guarantee the performance of our proposed method. We also appreciate the excellent baseline method FedAvg~\citep{mcmahan2017communication}, which provides an efficient federated learning way. Our FedDAG is highly inspired and benefits from these two works. 
The main contributions, which can be taken by our proposed method, are (1) \textbf{one of the first works} that investigate the practical problem of DAG structure learning in a federated setup and (2) further providing the FedDAG approach that can guarantee the \textbf{privacy protection} by avoiding the raw data leakage and allow the \textbf{data heterogeneity} across the clients. Another concurrent work NOTEARS-ADMM~\citep{ng2022towards} also considers the same problem while our GS-FedDAG can (1) gain better performances in most of the settings, (2) well handle the nonlinear cases, (3) allow heterogeneous data, and (4) provide a quite flexible federated DAG structure learning framework.

\textbf{Discussions on the simple averaging} \ Even though averaging is the simplest way to aggregating and exchanging information, we find it is \textbf{quite an effective} way to solve the federated DAG structure learning problem, which is an advantage of our method. Our simple averaging for homogeneous cases can nearly approach the same performance as using all data. For the heterogeneous cases, GS-FedDAG can still obtain satisfactory results. While, as future work, more advanced information aggregation methods~\citep{Wang2020Federated} can be well incorporated into our framework to boost the performances further.

\subsection{Difference with graph neural network (GNN) learning}

Four main reasons make DAG structure learning and GNN two different research lines. (1) Nodes in DAG represent variables, and directed edges describe the single-direction relation between different variables. In GNN, graph talks more about graph-type data, such as social networks, protein networks, and traffic networks. (2) Networks in DAG structure learning are leveraged to learn the relationship mechanisms, while networks in GNN are taken to achieve node embedding and feature extraction. (3) Learned DAG can be taken for interventional and counterfactual reasoning~\citep{kitson2021survey}. (4) DAG structure learning cares more about identifiability. It is essential to identify the true underlying relationship of the observations precisely. In the federated setup, most existing federated GNN~\citep{xie2021federated, liu2022federated} methods assume that the underlying graphs are known and localized. What is being learned in the federation is the weight aggregation of the GNN but not its graph. This also leads to the main difference between our federated DAG learning and federated GNN learning. 

\subsection{FedDAG as a framework}
\label{app:as_a_framework}
In this paper, we restrict our attention to the case that all concerned variables can be well observed. We also only take MCSL~\citep{ng2022masked} as the baseline method. However, all gradient-based methods can be incorporated into our AS-FedDAG framework to deal with homogeneous data. To deal with the heterogeneous data, we prefer that the baseline methods can separately learn the DAG structure and relationship mechanisms. The other baseline methods that can be easily combined into our framework are NOTEARS-MLP~\citep{zheng2020learning} and DAG-GNN~\citep{yu2019dag}. Unfortunately, many works are not in this fashion, such as GraN-DAG~\citep{lachapelle2020gradient}, CD-RL~\citep{zhu2020causal}, and their following works.

\textbf{Latent variables.} In this paper, we carry with no unobserved common confounder assumption. Handling latent confounders is a fundamentally important but challenging problem in the traditional DAG structure learning, not to mention the federated setup. Until now, the theoretical results on the structure identifiability of DAG learning with latent confounder are always too weak to be used in practice since too strict assumptions are taken. In the recent progress of the latent variables research, \citet{bhattacharya2021differentiable} takes the acyclic directed mixed graphs (ADMGs) to describe the graphs with latent confounders. With different types of restrictions, three classes of proprieties, named Ancestral graph, Acid graph, and Bow-free graph, are given. According to different proprieties, different graph constraints are given. For example, $\operatorname{trace}\left(e^{D}\right)-d+\operatorname{sum}(D \circ B)=0$ is set for the Bow-free graph\footnote{See more details at Section 4 in~\citep{bhattacharya2021differentiable}}, where $D$ is the adjacency matrix recording the directed edges and $B$ records the double-directed edges. We can directly replace the constraint to incorporate this method in our framework. However, this method can only deal with the linear Gaussian case, which is somewhat limited.

\subsection{Broader impact statement}
In federated learning, the server and some clients participate in this process. While as we talked about above, the DAG is shared among all clients. FedDAG is motivated by the "data on each client is not enough for identifying up the ground-truth DAG." The graph information is not private for clients. For the server, it depends. In our previous motivations, we only cared about the "raw data leakage" problem but did not consider the privacy of the graph. Some relations can be public in real-world scenarios, such as disease research. For these cases, our method can still work. However, graph structure may sometimes also be private information. This problem can be easily solved by picking one client as the proxy server. 

\subsection{The consistency results by BIC score}
\label{app:bic_score}
Actually, for linear additive noise models with Gaussian noises, the consistency results for maximizing the BIC score to identify the DAG (Markov Equivalence Class or DAG) have been well established~\citep{tian2001causal, huang2020causal2}. For this case, with the DAG space constraint, the unique maximum of score function $\mathcal{S}^{c_k} (\mathcal{D}^{c_k}, \bm \Phi^{c_k}, \bm U^{c_k})$ with BIC score corresponds to the ground-truth DAG. Even for the high-dimensional consistency for linear Gaussian SEM when the model is identifiable~\citep{bryon2019globally}. Since the ground-truth $\mathcal{G}$ corresponds to each $\mathcal{S}^{c_k}$, the global maximum $\mathop{\arg\max}_{\bm \Phi, \bm U}\sum_{k=1}^{m} \ \mathcal{S}^{c_k} (\mathcal{D}^{c_k}, \bm \Phi^{c_k}, \bm U)$ with DAG constraint can lead to the ground-truth DAG graph. For nonlinear ANMs, however, even many practical methods, e.g., MCSL~\citep{ng2022masked}, NOTEARS-MLP~\citep{zheng2020learning}, and CD-RL~\citep{zhu2020causal}, have been proposed to solve this problem by maximizing the BIC score, the theoretical results of consistency are still lacking and would be an interesting future work to be investigated. Therefore, our framework based on these methods inherits the theoretical limit for the nonlinear case. From our paper, however, empirical results can still show the method's effectiveness. 


\subsection{Does the global maximum of Eq.~(\ref{eq:FCD}) correspond to the ground-truth DAG?}

Firstly, for observations of identifiable ANMs on each client, the unique maximum of score function $\mathcal{S}^{c_k} (\mathcal{D}^{c_k}, \bm \Phi^{c_k}, \bm U^{c_k})$ with BIC score corresponds to the ground-truth DAG~\citep{zheng2018dags, ng2022masked}. Even for the high-dimensional consistency of linear Gaussian SEM in the case when the model is identifiable.
Since the ground-truth $\mathcal{G}$ corresponds to each $\mathcal{S}^{c_k}$, the global maximum $\mathop{\arg\max}_{\bm \Phi, \bm U}\sum_{k=1}^{m} \ \mathcal{S}^{c_k} (\mathcal{D}^{c_k}, \bm \Phi^{c_k}, \bm U^{c_k})$ with DAG constraint can lead to the ground-truth DAG.

\subsection{Can Algorithms~\ref{alg:DAG-FCD} and \ref{alg:SPS} solve Eq.~(\ref{eq:FCD})?}
 
Unfortunately, the global maximum of Eq.~(\ref{eq:FCD}) can not be well reached by the gradient-based optimization methods, which is mainly caused by the non-convex property of the acyclicity constraint. Firstly, discovering the ground-truth DAG is an NP-hard problem. Traditional methods like PC and GES search the discrete DAG space to solve this problem, which is relatively time-consuming. Then, NOTEARS introduces an equality constraint (\ref{eq:acyclicity_1}) to formulate the DAG search problem as a continuous optimization problem, which can be easily solved by the gradient descent methods. However, the trade-off is that this equality constraint is non-convex, which pushes us away from finding the ground-truth DAG (the global minima of (\ref{eq:FCD})). That is to say, using gradient descent to solve (\ref{eq:FCD}) only can reach the local minima of (\ref{eq:FCD}). This similar conclusion stands for recent continuous optimization-based CD methods such as GraNDAG~\citep{lachapelle2020gradient}, DAG-GNN~\citep{yu2019dag}, and NOTEARS-MLP~\citep{zheng2020learning}.

\subsection{Can $U$ finally satisfy the acyclicity constraint in Eq.~(\ref{eq:acyclicity_1})?}

For simplicity, please do not mind if we explain our method by setting some parameters with specific values. Firstly, following NOTEARS~\citep{zheng2018dags} and MCSL~\citep{ng2022masked}, we take the Augmented Lagrangian Method (ALM) to covert the constrained optimization problem into a series of sub-problems without the hard constraint but with two penalty terms. For the $t$-th sub-problem, the specific formulation of Eq.~(\ref{eq:sub-problem}) is related to $\alpha_t$ and $\rho_t$. $\alpha_t$ and $\rho_t$  will be updated to $\alpha_{t+1}$ and $\rho_{t+1}$ after solving the $t$-th sub-problem for $1000$ steps (gradient descent step). When dealing with each sub-problem, each client locally updates its personalized model \textit{with acyclicity penalty terms}, which is indeed for the acyclicity constraint. During the $1000$ steps, $\bm U$s are averaged every $200$ steps (Yes, the simple average is nothing with acyclicity). When finishing $1000$ steps (also \textit{the $5$-th $200$ steps} is just finished), a new $\bm U^{new}$ is obtained. Then, $\alpha_{t}$ and $\rho_{t}$ are updated to $\alpha_{t+1}$ and $\rho_{t+1}$ to formulate the next sub-problem of ALM, which are described in steps $5\sim 9$ in Algorithm~\ref{alg:DAG-FCD}. Then, a new circulation begins. Therefore, we argue that (1) \textit{the acyclicity constraint is guaranteed by taking the acyclicity penalty when solving each sub-problem}. (2) \textit{the convergence of $\bm U$ is supported by the convergence analysis of Personalized FedAvg of heterogeneous data}.

\section{Supplementary experimental details}
\label{app:sup_exp}

\subsection{Uneven distributions}
\begin{figure*}[ht]
\centering
\includegraphics[width=0.95\textwidth]{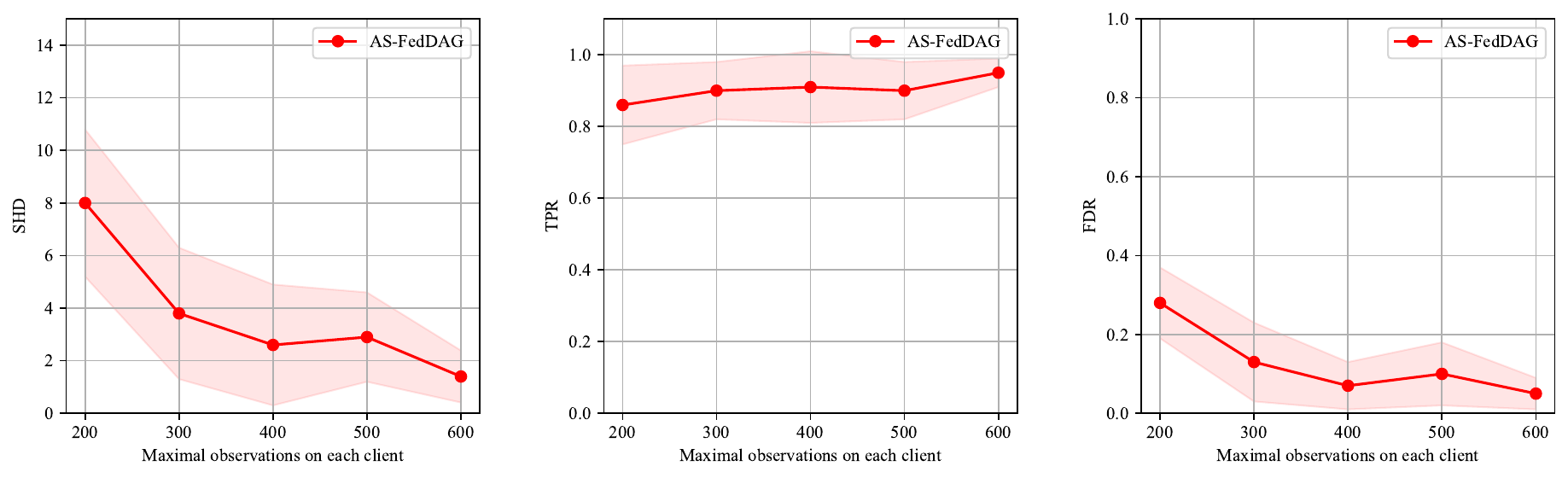}
\caption{Results of uneven distributions on different clients.}
\label{fig:uneven_distribution}
\end{figure*}

For federated learning problems in the real world, different clients may own different amounts of observations. To verify the stability of our method, we simulate the setting of uneven distributions in different clients. For each client, the number of observations is randomly chosen from a list $[20\%, 40\%, 60\%, 80\%] \times n$, where $n$ is the maximal observation. The experimental results are shown in Fig.~\ref{fig:uneven_distribution}, from which we can find that our method show relatively stable performance in this setting.

\subsection{Varying clients}
\begin{figure*}[ht]
\centering
\includegraphics[width=0.95\textwidth]{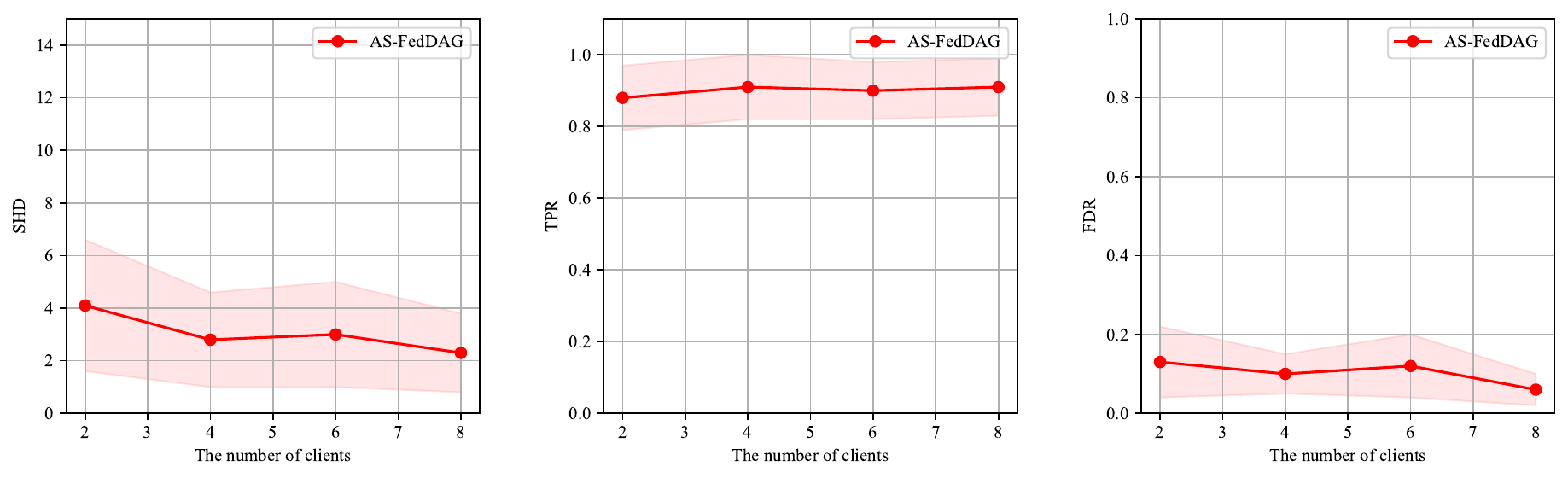}
\caption{Results of performances with varying clients.}
\label{fig:varying_clients}
\end{figure*}

In this setting, we now consider a fixed number of samples distributed across different clients. We conduct experiments for $(2,4,6,8)$ clients and show the results in Fig.~\ref{fig:varying_clients}. With the increase in clients number, our method can show better performance.

\subsection{Dense graphs} 
Our method is also implemented on some denser graphs. Experimental results in Table \ref{tab:dense4-IID} and Table \ref{tab:dense4-Non-IID}. From these experimental results, we can see that our method shows consistently better performance over other methods on the denser graph setting. For the homogeneous case, both AS-FedDAG and GS-FedDAG obtain the nearly low SHD as MCSL trained on all data and far better than all methods trained on separated data. For the heterogeneous case, our GS-FedDAG still shows the best performance. Compared to NOTEARS in 20 variables case, GS-FedDAG shows similar SHD results but a much better TPR result. Therefore, how to reduce the false discovery rate of GS-FedDAG would be an exciting thing.
\begin{table}[ht]
\centering
\caption{Results on nonlinear ANM with dense graphs (Homogeneous data).}
\label{tab:dense4-IID}
\resizebox{0.95\textwidth}{!}{
\begin{tabular}{clllllllllllll}
\toprule
~&  ~& \multicolumn{2}{c}{ER4 with 10 nodes} & \multicolumn{2}{c}{SF4 with 10 nodes} & \multicolumn{2}{c}{ER4 with 20 nodes} & \multicolumn{2}{c}{SF4 with 20 nodes} \\
    \cmidrule(lr){3-4} \cmidrule(lr){5-6} \cmidrule(lr){7-8} \cmidrule(lr){9-10}
~& ~& \multicolumn{1}{c}{SHD $\downarrow$} & \multicolumn{1}{c}{TPR $\uparrow$} & \multicolumn{1}{c}{SHD $\downarrow$} & \multicolumn{1}{c}{TPR $\uparrow$}
    & \multicolumn{1}{c}{SHD $\downarrow$} & \multicolumn{1}{c}{TPR $\uparrow$} & \multicolumn{1}{c}{SHD $\downarrow$} & \multicolumn{1}{c}{TPR $\uparrow$}
\\ \midrule
\multirow{3}{*}{\rotatebox{90} {All   data}} & PC &   27.3\,$\pm$\,3.2  & 0.29\,$\pm$\,0.07 &   18.9\,$\pm$\,4.9   &   0.37\,$\pm$\,0.16
    &    68.2\,$\pm$\,9.5 & 0.23\,$\pm$\,0.06 &  60.2\,$\pm$\,9.3 & 0.30\,$\pm$\,0.08 \\
& NOTEARS   &   34.3\,$\pm$\,1.7  & 0.03\,$\pm$\,0.02   & 22.7\,$\pm$\,1.3   &  0.05\,$\pm$\,0.05
    &  71.8\,$\pm$\,7.2 & 0.03\,$\pm$\,0.01 & 62.8\,$\pm$\,0.9 & 0.02\,$\pm$\,0.01 \\
& MCSL & \textit{15.5\,$\pm$\,5.9}  &  0.57\,$\pm$\,0.15  & \textit{4.5\,$\pm$\,3.1}   & \textit{0.83\,$\pm$\,0.11}
    &  \textit{33.8\,$\pm$\,10.4}  & 0.55\,$\pm$\,0.11 &  \textit{19.8\,$\pm$\,7.5}  & 0.69\,$\pm$\,0.11 \\ 
    \midrule
\multirow{3}{*}{\rotatebox{90} {Sep  data}} 
& PC&  31.5\,$\pm$\,2.1  &  0.14\,$\pm$\,0.03  & 20.4\,$\pm$\,0.58  & 0.21\,$\pm$\,0.03
    &  68.7\,$\pm$\,8.1  & 0.13\,$\pm$\,0.03 &  60.9\,$\pm$\,2.8 &  0.15\,$\pm$\,0.02 \\
& NOTEARS & 34.3\,$\pm$\,1.8  &  0.03\,$\pm$\,0.01  & 22.7\,$\pm$\,1.0   & 0.06\,$\pm$\,0.04
    &  70.1\,$\pm$\,6.9  & 0.03\,$\pm$\,0.01 &  62.3\,$\pm$\,0.56 & 0.03\,$\pm$\,0.01 \\
& MCSL & \textbf{15.8\,$\pm$\,3.3}  &  \textbf{0.61\,$\pm$\,0.09}  & 8.3\,$\pm$\,4.3   & \textbf{0.78\,$\pm$\,0.11}
    &  49.3\,$\pm$\,11.8  & \textbf{0.63\,$\pm$\,0.10} &  39.7\,$\pm$\,5.6  & \textbf{0.73\,$\pm$\,0.07} \\  \midrule
& GS-FedDAG & \textbf{16.9\,$\pm$\,4.9}  &  \textbf{0.53\,$\pm$\,0.12}  & \textbf{5.4\,$\pm$\,3.0} & 0.78\,$\pm$\,0.12
    &  \textbf{35.4\,$\pm$\,10.9}  & 0.53\,$\pm$\,0.11 &  \textbf{20.7\,$\pm$\,5.1} & 0.69\,$\pm$\,0.08 \\
& AS-FedDAG &  17.4\,$\pm$\,4.8  &  \textbf{0.53\,$\pm$\,0.12}  & \textbf{5.5\,$\pm$\,2.8}  & \textbf{0.79\,$\pm$\,0.11}
    &  \textbf{40.7\,$\pm$\,4.8}  & \textbf{0.57\,$\pm$\,0.10} &  \textbf{24.1\,$\pm$\,5.8} & \textbf{0.71\,$\pm$\,0.09} \\
\bottomrule
\end{tabular}}
\end{table}

\begin{table}[ht]
\centering
\caption{Results on nonlinear ANM with dense graphs (Heterogeneous data).}
\label{tab:dense4-Non-IID}
\resizebox{0.95\textwidth}{!}{
\begin{tabular}{clllllllllllll}
\toprule
~&  ~& \multicolumn{2}{c}{ER4 with 10 nodes} & \multicolumn{2}{c}{SF4 with 10 nodes} & \multicolumn{2}{c}{ER4 with 20 nodes} & \multicolumn{2}{c}{SF4 with 20 nodes} \\
    \cmidrule(lr){3-4} \cmidrule(lr){5-6} \cmidrule(lr){7-8} \cmidrule(lr){9-10}
~& ~& \multicolumn{1}{c}{SHD $\downarrow$} & \multicolumn{1}{c}{TPR $\uparrow$} & \multicolumn{1}{c}{SHD $\downarrow$} & \multicolumn{1}{c}{TPR $\uparrow$}
    & \multicolumn{1}{c}{SHD $\downarrow$} & \multicolumn{1}{c}{TPR $\uparrow$} & \multicolumn{1}{c}{SHD $\downarrow$} & \multicolumn{1}{c}{TPR $\uparrow$}
\\ \midrule
\multirow{3}{*}{\rotatebox{90} {Sep  data}} 
& PC&  29.3\,$\pm$\,1.3  &  0.23\,$\pm$\,0.03  & 20.3\,$\pm$\,2.1   & 0.31\,$\pm$\,0.06
    &  71.9\,$\pm$\,8.1  & 0.19\,$\pm$\,0.03 &  62.7\,$\pm$\,2.8 & 0.22\,$\pm$\,0.03 \\
& NOTEARS & 20.5\,$\pm$\,2.6  &  0.45\,$\pm$\,0.08  & 12.2\,$\pm$\,2.9   & 0.54\,$\pm$\,0.11
    &  43.2\,$\pm$\,7.0  & 0.49\,$\pm$\,0.08 &  \textbf{39.4\,$\pm$\,6.8} & 0.47\,$\pm$\,0.10 \\
& MCSL & 20.0\,$\pm$\,3.2  &  0.52\,$\pm$\,0.07  & 13.7\,$\pm$\,2.2   & 0.65\,$\pm$\,0.07
    &  65.1\,$\pm$\,7.7  & 0.33\,$\pm$\,0.05 &  59.4\,$\pm$\,5.3 & 0.31\,$\pm$\,0.05 \\ \midrule
& GS-FedDAG & \textbf{8.5\,$\pm$\,3.7}  &  \textbf{0.84\,$\pm$\,0.09}  & {4.5\,$\pm$\,2.0}   & \textbf{0.93\,$\pm$\,0.07}
    &  \textbf{40.7\,$\pm$\,14.5}  & \textbf{0.74\,$\pm$\,0.07} &  39.9\,$\pm$\,10.8 & \textbf{0.68\,$\pm$\,0.07} \\
\bottomrule
\end{tabular}}
\end{table}

\subsection{Comparisons with voting method}
\label{app:voting_method}
There is another interesting research line~\citep{na2010distributed}, which also tries to learn DAG from decentralized data. We add a DAG combination method proposed in~\citep{na2010distributed}, which proposes to vote for each entry of the adjacency matrix to get the final DAG. From the experimental results in Table~\ref{tab:vote}, we can find that For PC and NOTEARS, the combining method seems to contribute little improvement. This is because the reported DAGs local clients are too bad to get a good result. For MCSL, this combing method works well for improving performance. The reason is easy to be inferred from the results. For MCSL, DAGs reported by local clients are of bad SHDs but good TPR, which means that the False Discovery Rates (FDRs) are high. In contrast, the combing method can further reduce the FDRs and keep the TPRs still good. Then, SHD can be further reduced. Luckily, our GS-FedDAG still shows the best performances in all settings. 
\begin{table}[ht]
\centering
\caption{Comparison with the voting method.}
\label{tab:vote}
\resizebox{0.95\textwidth}{!}{
\begin{tabular}{clllllllllllll}
\toprule
~&  ~& \multicolumn{4}{c}{Homogeneous data (GP)} & \multicolumn{4}{c}{Heterogeneous data} \\
    \cmidrule(lr){3-6} \cmidrule(lr){7-10}
~&  ~& \multicolumn{2}{c}{ER2 with 10 nodes} & \multicolumn{2}{c}{ER2 with 20 nodes} & \multicolumn{2}{c}{ER2 with 10 nodes} & \multicolumn{2}{c}{ER2 with 20 nodes} \\
    \cmidrule(lr){3-4} \cmidrule(lr){5-6} \cmidrule(lr){7-8} \cmidrule(lr){9-10}
~& ~& \multicolumn{1}{c}{SHD $\downarrow$} & \multicolumn{1}{c}{TPR $\uparrow$} & \multicolumn{1}{c}{SHD $\downarrow$} & \multicolumn{1}{c}{TPR $\uparrow$}
    & \multicolumn{1}{c}{SHD $\downarrow$} & \multicolumn{1}{c}{TPR $\uparrow$} & \multicolumn{1}{c}{SHD $\downarrow$} & \multicolumn{1}{c}{TPR $\uparrow$}
\\ \midrule
\multirow{3}{*}{\rotatebox{90} {Sep data}} 
& PC &   14.1\,$\pm$\,2.4  & 0.31\,$\pm$\,0.06 &   32.7\,$\pm$\,6.5   &  0.28\,$\pm$\,0.07
    &    12.5\,$\pm$\,2.7 & 0.45\,$\pm$\,0.07 &  28.5\,$\pm$\,6.3 & 0.44\,$\pm$\,0.07 \\
& NOTEARS   &  16.5\,$\pm$\,2.0  & 0.06\,$\pm$\,0.04   & 31.7\,$\pm$\,6.0   &  0.11\,$\pm$\,0.04
    &  7.6\,$\pm$\,2.6 & 0.60\,$\pm$\,0.11 & 15.0\,$\pm$\,3.1 & 0.62\,$\pm$\,0.09 \\
& MCSL & 7.1\,$\pm$\,3.2  &  0.83\,$\pm$\,0.08  & 24.8\,$\pm$\,5.5   & \textit{0.88\,$\pm$\,0.07}
    &  9.2\,$\pm$\,1.8  & 0.72\,$\pm$\,0.06 &  23.3\,$\pm$\,5.8  & 0.56\,$\pm$\,0.08 \\ 
    \midrule
\multirow{3}{*}{\rotatebox{90} {Voting}} 
& PC&  13.3\,$\pm$\,3.0  &  0.27\,$\pm$\,0.11  & 29.7\,$\pm$\,5.9  & 0.22\,$\pm$\,0.05
    &  11.4\,$\pm$\,3.4  & 0.36\,$\pm$\,0.13 &  25.5\,$\pm$\,6.8 &  0.29\,$\pm$\,0.13 \\
& NOTEARS & 15.6\,$\pm$\,2.2  &  0.11\,$\pm$\,0.06  & 32.6\,$\pm$\,6.2   & 0.09\,$\pm$\,0.05
    &  7.8\,$\pm$\,4.0  & 0.56\,$\pm$\,0.20 &  18.4\,$\pm$\,11.6 & 0.49\,$\pm$\,0.30 \\
& MCSL & 8.0\,$\pm$\,3.1  &  0.85\,$\pm$\,0.16  & 18.1\,$\pm$\,7.8   & \textbf{0.88\,$\pm$\,0.06}
    &  6.9\,$\pm$\,2.2  & 0.71\,$\pm$\,0.13 &  10.1\,$\pm$\,4.6  & 0.79\,$\pm$\,0.09 \\  \midrule
& GS-FedDAG & \textbf{2.4\,$\pm$\,2.0}  &  \textbf{0.86\,$\pm$\,0.12}  & \textbf{6.2\,$\pm$\,4.0}   & 0.85\,$\pm$\,0.10
    &  \textbf{1.9\,$\pm$\,1.6}  & \textbf{0.99\,$\pm$\,0.02} &  \textbf{6.2\,$\pm$\,4.7} & \textbf{0.89\,$\pm$\,0.09} \\
& AS-FedDAG &  \textbf{1.8\,$\pm$\,2.0}  &  \textbf{0.89\,$\pm$\,0.12}  & \textbf{5.0\,$\pm$\,4.2}   & \textbf{0.88\,$\pm$\,0.11}
    &  NaN  & NaN &  NaN & NaN \\
\bottomrule
\end{tabular}}
\end{table}

\subsection{Comparisons with CAM}
\label{app:cam}
Here, we add one more identifiable baseline named causal additive model (CAM) \citep{buhlmann2014cam}, which also serves as a baseline in MCSL~\citep{ng2022masked}, GraNDAG~\citep{lachapelle2020gradient}, and DAG-GAN~\citep{yu2019dag}. From results in Table \ref{tab:cam-iid} and \ref{tab:cam-noniid}, we can see that our methods always show an advantage over CAM. CAM also assumes a non-linear ANM for data generation. However, CAM limits the non-linear function to be additive. In normal ANM, $X_i = f_i(X_{pa_i}) + \epsilon_i$ while CAM assumes $X_i = \sum_{j\in X(pa_i)} f_{i \leftarrow j}(X_j) + \epsilon_i$, which limits the capacity of its model. From the above experimental results, we can see that our methods show consistent advantages over CAM. 

\begin{table}[ht]
\centering
\caption{Comparisons with CAM on nonlinear ANM (Homogeneous data-GP).}
\label{tab:cam-iid}
\resizebox{0.95\textwidth}{!}{
\begin{tabular}{clllllllllllll}
\toprule
~&  ~& \multicolumn{2}{c}{ER2 with 10 nodes} & \multicolumn{2}{c}{SF2 with 10 nodes} & \multicolumn{2}{c}{ER2 with 20 nodes} & \multicolumn{2}{c}{SF2 with 20 nodes} \\
    \cmidrule(lr){3-4} \cmidrule(lr){5-6} \cmidrule(lr){7-8} \cmidrule(lr){9-10}
~& ~& \multicolumn{1}{c}{SHD $\downarrow$} & \multicolumn{1}{c}{TPR $\uparrow$} & \multicolumn{1}{c}{SHD $\downarrow$} & \multicolumn{1}{c}{TPR $\uparrow$}
    & \multicolumn{1}{c}{SHD $\downarrow$} & \multicolumn{1}{c}{TPR $\uparrow$} & \multicolumn{1}{c}{SHD $\downarrow$} & \multicolumn{1}{c}{TPR $\uparrow$}
\\ \midrule
All data & CAM&  9.5\,$\pm$\,2.9  &  0.87\,$\pm$\,0.09  & 9.1\,$\pm$\,3.1   & 0.84\,$\pm$\,0.10
    &  21.4\,$\pm$\,4.7  & 0.77,$\pm$\,0.08 &  26.6\,$\pm$\,6.1 & 0.75\,$\pm$\,0.07 \\
 \midrule
Sep data & CAM&  11.8\,$\pm$\,2.6  &  0.40\,$\pm$\,0.10  & 11.1\,$\pm$\,1.5   & 0.38\,$\pm$\,0.11
    &  24.3\,$\pm$\,5.8  & 0.40\,$\pm$\,0.07 &  26.8\,$\pm$\,2.0 & 0.36\,$\pm$\,0.06 \\
\midrule
& GS-FedDAG & 2.4\,$\pm$\,2.0  &  0.86\,$\pm$\,0.12  & 2.7\,$\pm$\,2.2   & \textbf{0.86\,$\pm$\,0.13}
    &  6.2\,$\pm$\,4.0  & 0.85\,$\pm$\,0.10 &  14.7\,$\pm$\,7.0 & \textbf{0.80\,$\pm$\,0.11} \\
\midrule
& AS-FedDAG & \textbf{1.8\,$\pm$\,2.0}  &  \textbf{0.89\,$\pm$\,0.12}  & \textbf{2.5\,$\pm$\,2.7}   & 0.85\,$\pm$\,0.15
    &  \textbf{5.0\,$\pm$\,4.2}  & \textbf{0.88\,$\pm$\,0.11} &  \textbf{7.8\,$\pm$\,5.5} & 0.80\,$\pm$\,0.14 \\
\bottomrule
\end{tabular}}
\end{table}

\begin{table}[ht]
\centering
\caption{Comparisons with CAM on nonlinear ANM (Heterogeneous data).}
\label{tab:cam-noniid}
\resizebox{0.95\textwidth}{!}{
\begin{tabular}{clllllllllllll}
\toprule
~&  ~& \multicolumn{2}{c}{ER2 with 10 nodes} & \multicolumn{2}{c}{SF2 with 10 nodes} & \multicolumn{2}{c}{ER2 with 20 nodes} & \multicolumn{2}{c}{SF2 with 20 nodes} \\
    \cmidrule(lr){3-4} \cmidrule(lr){5-6} \cmidrule(lr){7-8} \cmidrule(lr){9-10}
~& ~& \multicolumn{1}{c}{SHD $\downarrow$} & \multicolumn{1}{c}{TPR $\uparrow$} & \multicolumn{1}{c}{SHD $\downarrow$} & \multicolumn{1}{c}{TPR $\uparrow$}
    & \multicolumn{1}{c}{SHD $\downarrow$} & \multicolumn{1}{c}{TPR $\uparrow$} & \multicolumn{1}{c}{SHD $\downarrow$} & \multicolumn{1}{c}{TPR $\uparrow$}
\\ \midrule
All data & CAM&  31.9\,$\pm$\,4.8  &  0.39\,$\pm$\,0.15  & 31.8\,$\pm$\,4.4   & 0.31\,$\pm$\,0.17
    &  104.6\,$\pm$\,15.4  & 0.46 $\pm$\,0.15 &  116.9\,$\pm$\,13.8  & 0.35\,$\pm$\,0.07 \\
 \midrule
Sep data & CAM&  18.0\,$\pm$\,1.7  &  0.52\,$\pm$\,0.04  & 17.8\,$\pm$\,2.1   & 0.51\,$\pm$\,0.3
    &  47.5\,$\pm$\,9.2  & 0.52\,$\pm$\,0.04 &  53.0\,$\pm$\,6.1 & 0.50\,$\pm$\,0.03 \\
\midrule
& GS-FedDAG & \textbf{1.9\,$\pm$\,1.6}  &  \textbf{0.99\,$\pm$\,0.02}  & \textbf{2.6\,$\pm$\,1.3}   & \textbf{0.93\,$\pm$\,0.07}
    &  \textbf{6.2\,$\pm$\,4.7}  & \textbf{0.89\,$\pm$\,0.09} &  \textbf{11.5\,$\pm$\,6.7} & \textbf{0.81\,$\pm$\,0.14} \\
\bottomrule
\end{tabular}}
\end{table}

\subsection{Comparisons with NOTEARS-ADMM}
\label{app:comp_admm}
\begin{figure}[ht]
\centering
\subfloat[ER1 with 10 nodes.]{
  \includegraphics[width=0.45\textwidth]{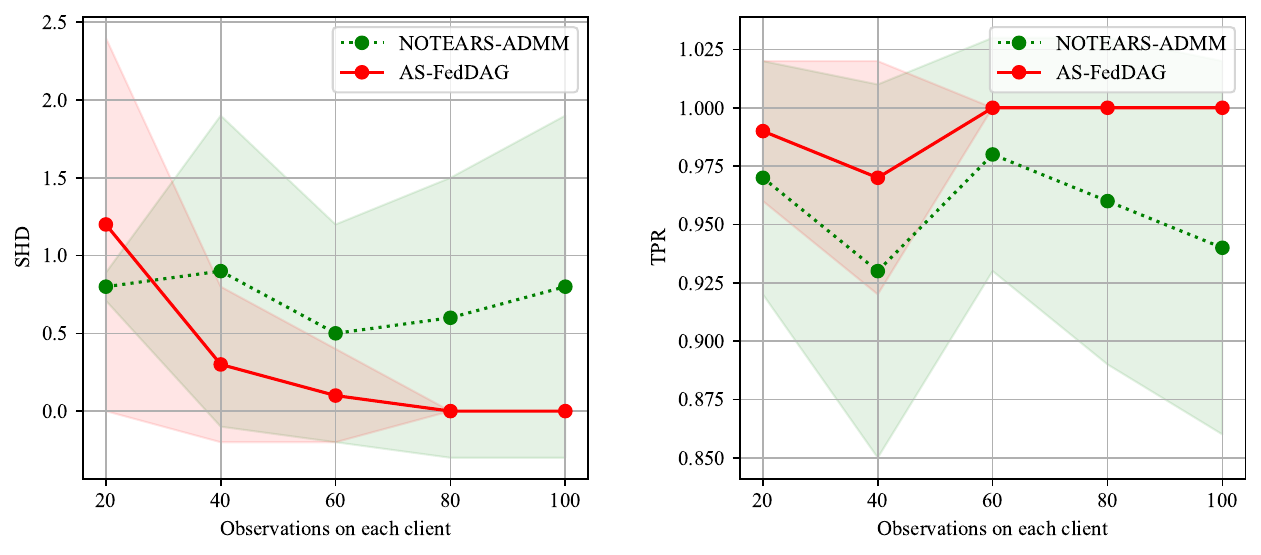}
}
\subfloat[ER2 with 10 nodes.]{
  \includegraphics[width=0.45\textwidth]{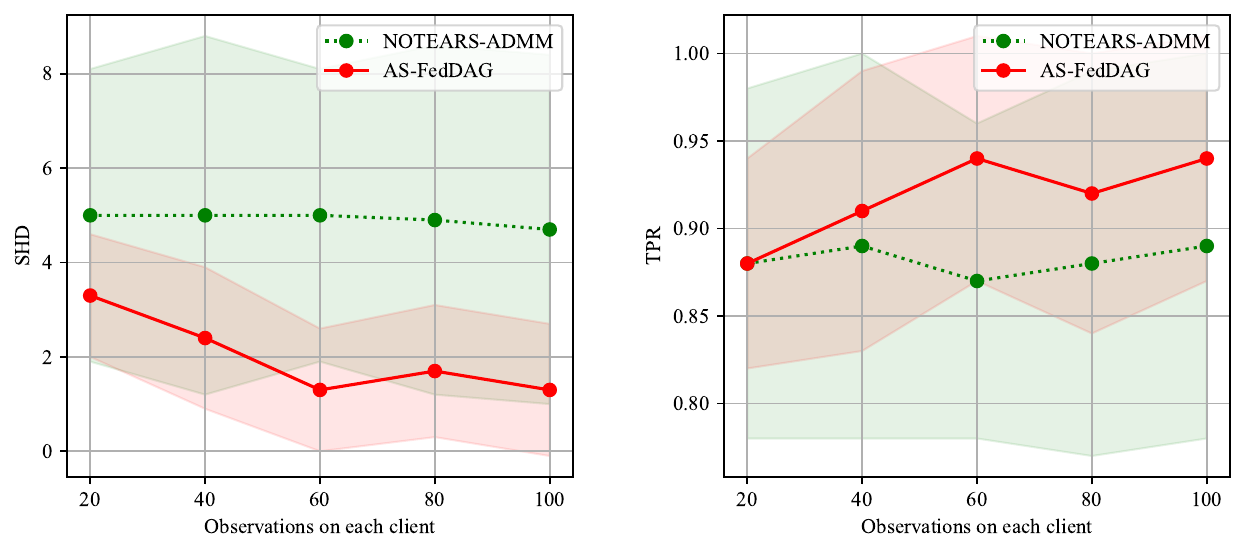}
}\\
\subfloat[ER1 with 20 nodes.]{
  \includegraphics[width=0.45\textwidth]{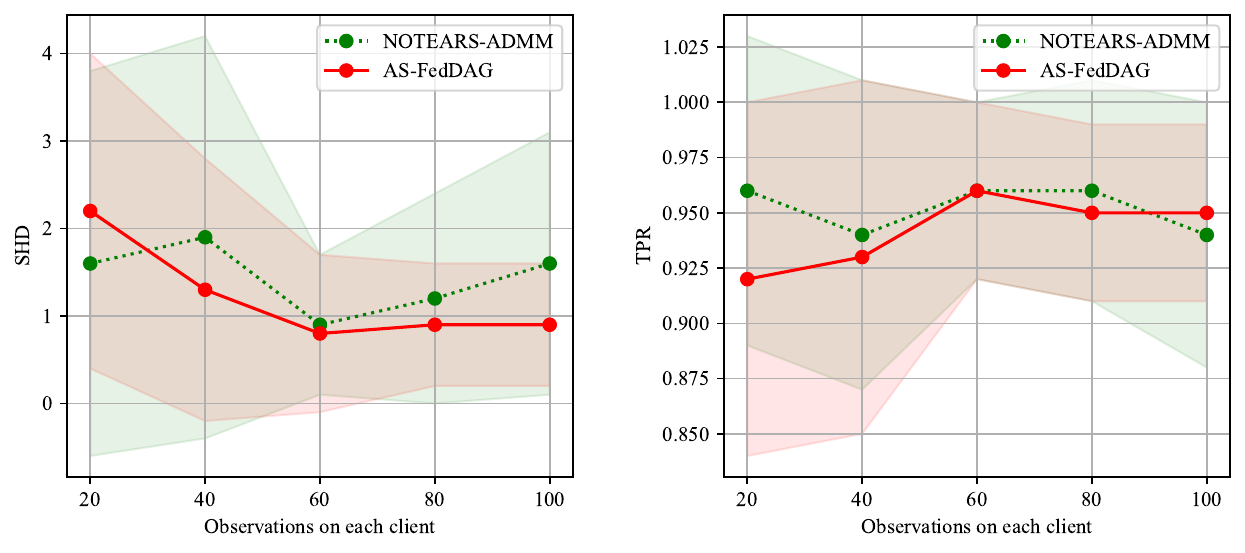}
}
\subfloat[ER1 with 20 nodes.]{
  \includegraphics[width=0.45\textwidth]{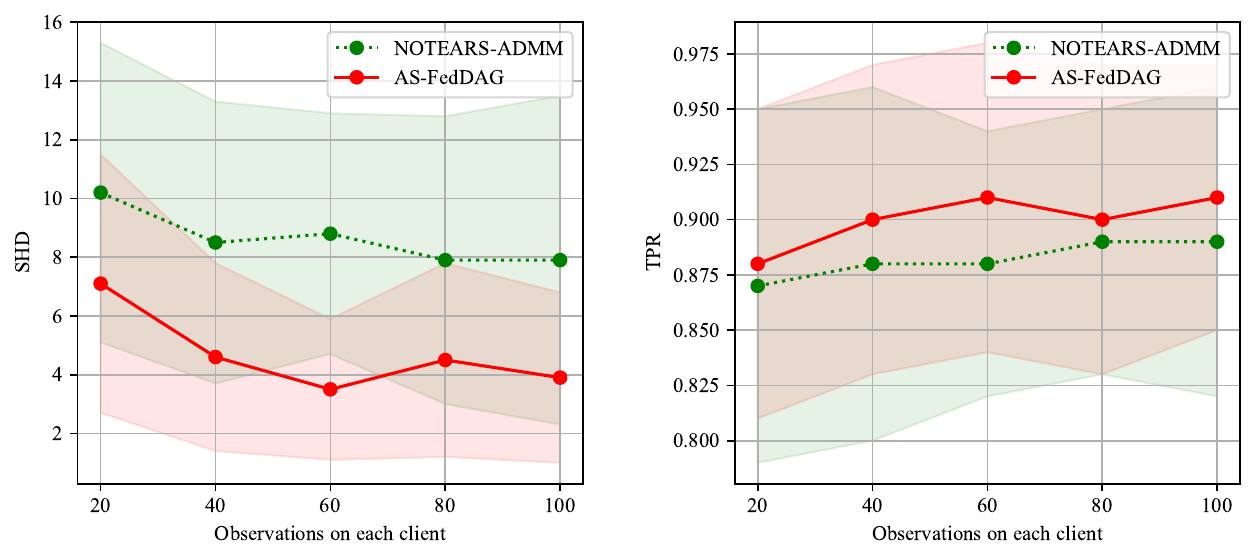}
}
\caption{Comparisons with NOTEARS-ADMM on the linear model (IID).}
\label{fig:comp_admm_linear}
\end{figure}

\begin{figure}[ht]
\centering
\subfloat[ER1 with 10 nodes (GP).]{
  \includegraphics[width=0.45\textwidth]{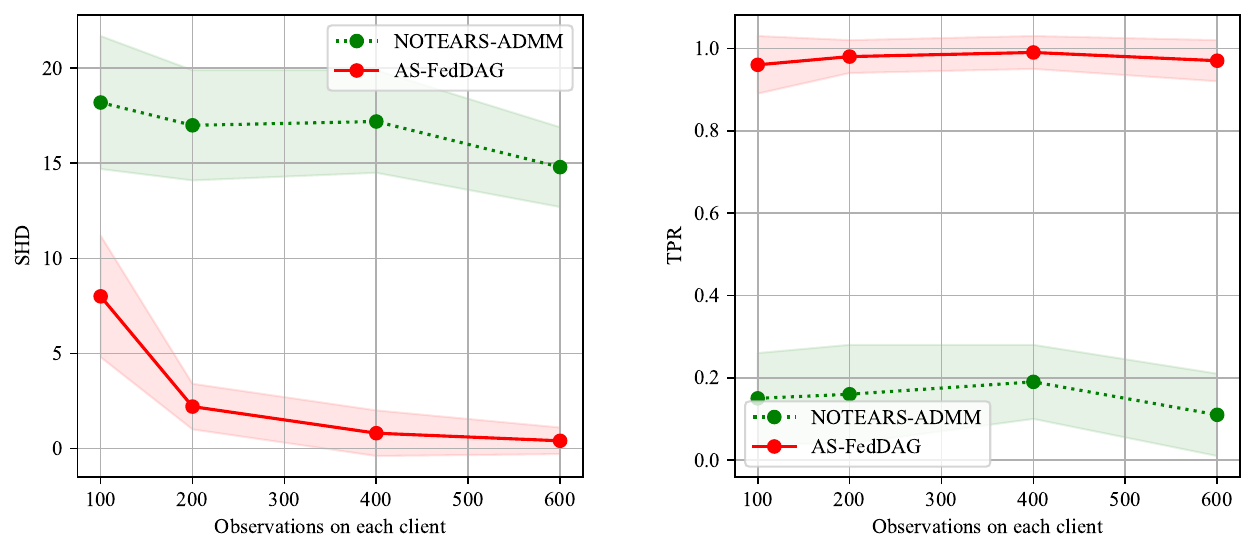}
}
\subfloat[ER2 with 10 nodes (GP).]{
  \includegraphics[width=0.45\textwidth]{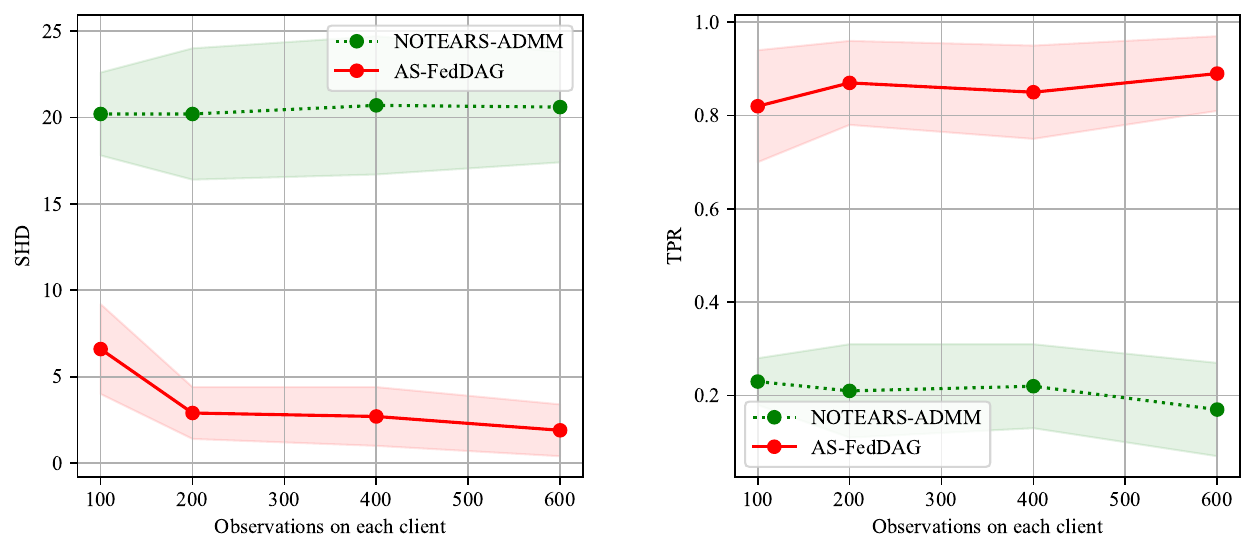}
}\\
\subfloat[ER1 with 10 nodes (MLP).]{
  \includegraphics[width=0.45\textwidth]{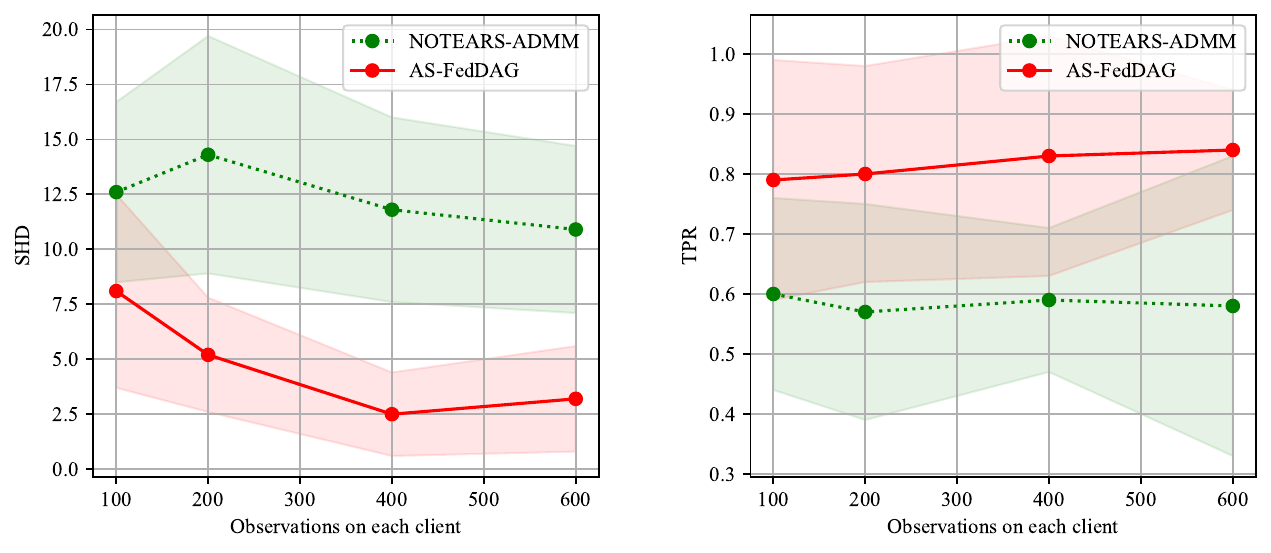}
}
\subfloat[ER2 with 10 nodes (MLP).]{
  \includegraphics[width=0.45\textwidth]{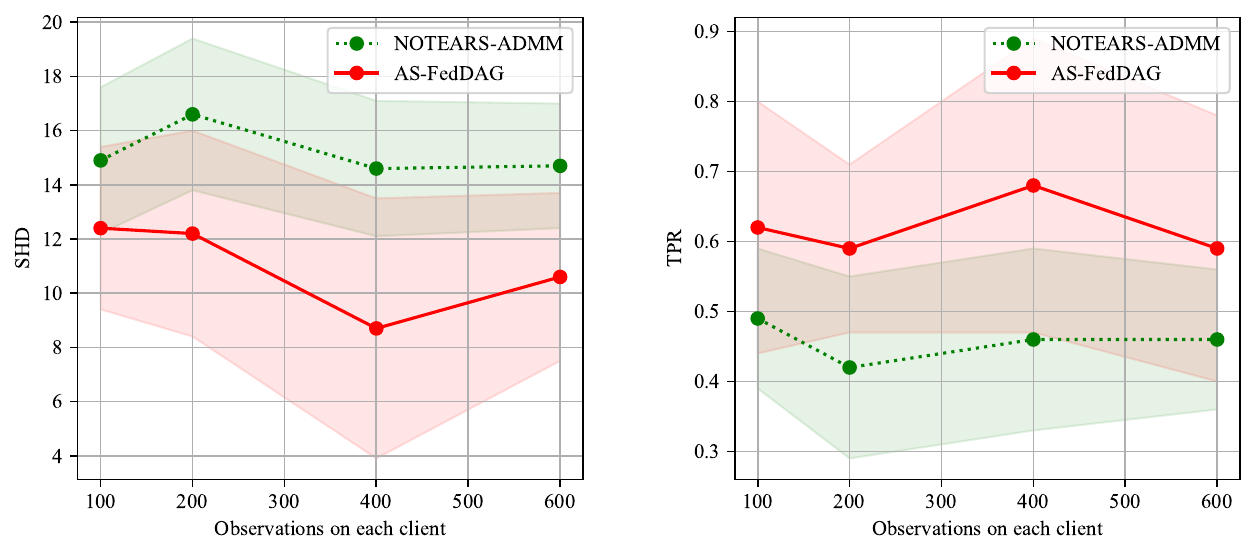}
}
\caption{Comparisons with NOTEARS-ADMM on nonlinear models (Homogeneous data).}
\label{fig:comp_admm_nonlinear}
\end{figure}

In this subsection, we give the experimental comparisons with NOTEARS-ADMM in detail to verify the advantage of our averaging strategy is simple but effective. Firstly, we conduct the results on linear models, which are the main part in~\citet{ng2022towards}. As shown in Fig.~\ref{fig:comp_admm_linear}, even on linear models, our AS-FedDAG can consistently show its advantage over NOTEARS-ADMM. Then, for the nonlinear models, we consider two different functions named MLP and Gaussian process (GP). The results are presented in Fig.~\ref{fig:comp_admm_nonlinear}, from which we can see that FedDAG always show better performance over all settings. Since NOTEARS-ADMM can not handle heterogeneous data, we do not give the results on heterogeneous data for fair comparison.

\begin{table}[ht]
\centering
\caption{Results on nonlinear ANM with different functions (Homogeneous data, 10 nodes, ER2).}
\label{tab:results_sem_type_10nodes}
\resizebox{0.95\textwidth}{!}{
\begin{tabular}{clllllllllllll}
\toprule
~&  ~& \multicolumn{2}{c}{GP} & \multicolumn{2}{c}{MIM} & \multicolumn{2}{c}{MLP} & \multicolumn{2}{c}{GP-add} \\
    \cmidrule(lr){3-4} \cmidrule(lr){5-6} \cmidrule(lr){7-8} \cmidrule(lr){9-10}
~& ~& \multicolumn{1}{c}{SHD $\downarrow$} & \multicolumn{1}{c}{TPR $\uparrow$} & \multicolumn{1}{c}{SHD $\downarrow$} & \multicolumn{1}{c}{TPR $\uparrow$}
    & \multicolumn{1}{c}{SHD $\downarrow$} & \multicolumn{1}{c}{TPR $\uparrow$} & \multicolumn{1}{c}{SHD $\downarrow$} & \multicolumn{1}{c}{TPR $\uparrow$}
\\ \midrule
\multirow{6}{*}{\rotatebox{90} {All   data}} & PC &    15.3\,$\pm$\,2.6  & 0.37\,$\pm$\,0.10 &   11.0\,$\pm$\,4.9   &   0.60\,$\pm$\,0.16
    &    11.8\,$\pm$\,4.3 & 0.61\,$\pm$\,0.14 &  14.0\,$\pm$\,4.7 & 0.49\,$\pm$\,0.16 \\
& GES   &   13.0\,$\pm$\,3.9  & 0.50\,$\pm$\,0.18 &   9.6\,$\pm$\,4.4   &   0.71\,$\pm$\,0.17
    &    15.8\,$\pm$\,6.0 & 0.63\,$\pm$\,0.14 &  14.4\,$\pm$\,4.9 & 0.57\,$\pm$\,0.17 \\
& DAG-GNN   &    16.2\,$\pm$\,2.1  & 0.07\,$\pm$\,0.06 &   13.7\,$\pm$\,2.4   &   0.26\,$\pm$\,0.10
    &    18.2\,$\pm$\,3.3  & 0.36\,$\pm$\,0.12 &   13.3\,$\pm$\,2.3   &   0.24\,$\pm$\,0.10 \\
& NOTEARS    &    16.5\,$\pm$\,2.0  & 0.05\,$\pm$\,0.04   & 12.1\,$\pm$\,3.2   &  0.34\,$\pm$\,0.13
    &  13.3\,$\pm$\,3.4 & 0.35\,$\pm$\,0.15 & 13.4\,$\pm$\,2.2 & 0.23\,$\pm$\,0.09 \\
& N-S-MLP  & 8.1\,$\pm$\,3.8  & 0.56\,$\pm$\,0.17   & 1.6\,$\pm$\,1.3   &   0.95\,$\pm$\,0.06
    &  \textit{5.6\,$\pm$\,1.3} & \textit{0.81\,$\pm$\,0.11 } & 6.8\,$\pm$\,4.0& 0.65\,$\pm$\,0.16 \\
& MCSL &  \textit{1.9\,$\pm$\,1.5}  & \textit{0.90\,$\pm$\,0.08}   & \textit{0.7\,$\pm$\,1.2}  &  \textit{0.97\,$\pm$\,0.06}
    &  12.7\,$\pm$\,3.6 & 0.58\,$\pm$\,0.24 & \textit{1.9\,$\pm$\,1.7} & \textit{0.91\,$\pm$\,0.07}\\ \midrule
\multirow{6}{*}{\rotatebox{90} {Sep  data}} & PC&    14.1\,$\pm$\,2.4  & 0.31\,$\pm$\,0.06 &   11.1\,$\pm$\,3.6   &   0.48\,$\pm$\,0.14
    &    13.2\,$\pm$\,3.6 & 0.42\,$\pm$\,0.09 &  13.5\,$\pm$\,3.2 & 0.37\,$\pm$\,0.12 \\
& GES   &   12.7\,$\pm$\,2.7  & 0.37\,$\pm$\,0.09 &   10.6\,$\pm$\,3.3   &   0.54\,$\pm$\,0.12
    &    14.6\,$\pm$\,4.6 & 0.50\,$\pm$\,0.13 &  12.0\,$\pm$\,2.6 & 0.48\,$\pm$\,0.08 \\
& DAG-GNN  &   15.7\,$\pm$\,2.3  & 0.11\,$\pm$\,0.05 &   11.7\,$\pm$\,3.3   &   0.37\,$\pm$\,0.12
    &    17.7\,$\pm$\,3.6 & 0.39\,$\pm$\,0.11 &  13.0\,$\pm$\,2.0 & 0.26\,$\pm$\,0.10 \\
& NOTEARS &  16.5\,$\pm$\,2.0  & 0.06\,$\pm$\,0.04 &   12.3\,$\pm$\,3.0   &   0.33\,$\pm$\,0.12
    &    13.4\,$\pm$\,3.4 & 0.35\,$\pm$\,0.14 &  13.3\,$\pm$\,2.3 & 0.24\,$\pm$\,0.09 \\
& N-S-MLP  & 8.5\,$\pm$\,2.9  & 0.56\,$\pm$\,0.13 &   2.8\,$\pm$\,1.5   &    \textbf{0.93\,$\pm$\,0.06}
    &   {\bf 6.4\,$\pm$\,1.3}  & {\bf 0.81\,$\pm$\,0.11} &  7.4\,$\pm$\,2.9 & 0.67\,$\pm$\,0.13 \\
& MCSL & 7.1\,$\pm$\,3.2  & 0.83\,$\pm$\,0.08 &   4.4\,$\pm$\,2.1   &   \textbf{0.91\,$\pm$\,0.06}
    &    13.4\,$\pm$\,3.9 & 0.57\,$\pm$\,0.21 &  6.5\,$\pm$\,3.5 & 0.84\,$\pm$\,0.07 \\ \midrule
& GS-FedDAG &   \textbf{2.4\,$\pm$\,2.0}  & \textbf{0.86\,$\pm$\,0.12} &   \textbf{2.1\,$\pm$\,1.4}   &   0.91\,$\pm$\,0.07
    &    11.1\,$\pm$\,3.1 & 0.57\,$\pm$\,0.20 &  {\bf 2.6\,$\pm$\,1.6} & {\bf 0.87\,$\pm$\,0.09} \\ 
& AS-FedDAG &  {\bf 1.8\,$\pm$\,2.0}  & {\bf 0.89\,$\pm$\,0.12 } &  {\bf 1.7\,$\pm$\,1.6}   &   0.91\,$\pm$\,0.08
    &    \textbf{10.5\,$\pm$\,3.5} & \textbf{0.59\,$\pm$\,0.22} &  \textbf{2.4\,$\pm$\,1.6} &  \textbf{0.87\,$\pm$\,0.08} \\
\bottomrule
\end{tabular}}
\end{table}

\begin{table}[ht]
\centering
\caption{Results on nonlinear ANM with different functions (Homogeneous data, 20 nodes, ER2).}
\label{tab:results_sem_type_20nodes}
\resizebox{0.95\textwidth}{!}{
\begin{tabular}{clllllllllllll}
\toprule
~&  ~& \multicolumn{2}{c}{ GP} & \multicolumn{2}{c}{MIM} & \multicolumn{2}{c}{MLP} & \multicolumn{2}{c}{GP-add} \\
    \cmidrule(lr){3-4} \cmidrule(lr){5-6} \cmidrule(lr){7-8} \cmidrule(lr){9-10}
~& ~& \multicolumn{1}{c}{SHD $\downarrow$} & \multicolumn{1}{c}{TPR $\uparrow$} & \multicolumn{1}{c}{SHD $\downarrow$} & \multicolumn{1}{c}{TPR $\uparrow$}
    & \multicolumn{1}{c}{SHD $\downarrow$} & \multicolumn{1}{c}{TPR $\uparrow$} & \multicolumn{1}{c}{SHD $\downarrow$} & \multicolumn{1}{c}{TPR $\uparrow$}
\\ \midrule
\multirow{6}{*}{\rotatebox{90} {All   data}} & PC &    32.7\,$\pm$\,9.4  & 0.48\,$\pm$\,0.13 &   22.8\,$\pm$\,5.8   &   0.60\,$\pm$\,0.15
    &    33.7\,$\pm$\,12.3 & 0.50\,$\pm$\,0.13 &  35.2\,$\pm$\,8.0 & 0.50\,$\pm$\,0.09 \\
& GES   &   27.1\,$\pm$\,8.5  & 0.56\,$\pm$\,0.11 &   21.5\,$\pm$\,6.1   &   0.78\,$\pm$\,0.09
    &    44.9\,$\pm$\,12.5 & 0.65\,$\pm$\,0.11 &  41.7\,$\pm$\,11.6 & 0.66\,$\pm$\,0.08 \\
& DAG-GNN   &    32.5\,$\pm$\,6.8  & 0.10\,$\pm$\,0.08 &   26.7\,$\pm$\,7.4   &   0.26\,$\pm$\,0.13
    &    32.1\,$\pm$\,10.4  & 0.38\,$\pm$\,0.08 &   27.2\,$\pm$\,2.4   &   0.24\,$\pm$\,0.08 \\
& NOTEARS    &   31.8\,$\pm$\,6.0  & 0.11\,$\pm$\,0.04  & 25.6\,$\pm$\,6.1   &  0.29\,$\pm$\,0.08
    &  25.3\,$\pm$\,8.0 & 0.40\,$\pm$\,0.09 &  25.6\,$\pm$\,3.9 & 0.28\,$\pm$\,0.06 \\
& N-S-MLP  & 18.2\,$\pm$\,4.5  & 0.52\,$\pm$\,0.10   & 4.1\,$\pm$\,2.0   &   0.95\,$\pm$\,0.04
    &  \textit{8.0\,$\pm$\,3.9} & \textit{0.86\,$\pm$\,0.07} & 12.6\,$\pm$\,2.2 & 0.70\,$\pm$\,0.06 \\
& MCSL &  \textit{ 4.6\,$\pm$\,4.6}  & \textit{0.90\,$\pm$\,0.13}   & \textit{1.7\,$\pm$\,1.6}  &  \textit{0.97\,$\pm$\,0.04}
    &  18.1\,$\pm$\,6.6 & 0.72\,$\pm$\,0.14 & \textit{3.1\,$\pm$\,1.9} & \textit{0.92\,$\pm$\,0.05}\\ \midrule
\multirow{6}{*}{\rotatebox{90} {Sep  data}} & PC&    32.7\,$\pm$\,6.5  & 0.28\,$\pm$\,0.07 &   24.4\,$\pm$\,5.6   &   0.46\,$\pm$\,0.11
    &    30.6\,$\pm$\,8.0 & 0.41\,$\pm$\,0.09 &  29.5\,$\pm$\,5.6 & 0.42\,$\pm$\,0.10 \\
& GES   &   28.6\,$\pm$\,5.5  & 0.34\,$\pm$\,0.06 &   20.5\,$\pm$\,3.7   & 0.61\,$\pm$\,0.06
    &    34.4\,$\pm$\,11.3 & 0.52\,$\pm$\,0.09 &  29.3\,$\pm$\,5.5 & 0.51\,$\pm$\,0.07 \\
& DAG-GNN  &   31.7\,$\pm$\,6.1  & 0.12\,$\pm$\,0.04 &   26.8\,$\pm$\,5.8   &   0.26\,$\pm$\,0.06
    &    34.1\,$\pm$\,9.7 & 0.46\,$\pm$\,0.07 &  26.5\,$\pm$\,4.0 & 0.27\,$\pm$\,0.05 \\
& NOTEARS &  31.7\,$\pm$\,6.0  & 0.11\,$\pm$\,0.04 &   25.7\,$\pm$\,5.9   &   0.29\,$\pm$\,0.07
    &    25.4\,$\pm$\,7.4 & 0.42\,$\pm$\,0.07 &  25.6\,$\pm$\,3.8 & 0.29\,$\pm$\,0.06 \\
& N-S-MLP  & 19.5\,$\pm$\,4.7  & 0.52\,$\pm$\,0.07 &   \textbf{6.5\,$\pm$\,1.9}   &    \textbf{0.92\,$\pm$\,0.03}
    &   {\bf 16.1\,$\pm$\,8.6}  & {\bf 0.86\,$\pm$\,0.07} &  16.2\,$\pm$\,3.3 & 0.70\,$\pm$\,0.07 \\
& MCSL & 24.8\,$\pm$\,5.5  & \textbf{0.88\,$\pm$\,0.07} &   20.4\,$\pm$\,3.8   &   0.91\,$\pm$\,0.05
    &    30.2\,$\pm$\,5.1 & 0.67\,$\pm$\,0.12 &  16.2\,$\pm$\,5.3 & \textbf{0.87\,$\pm$\,0.05} \\ \midrule
& GS-FedDAG &   \textbf{6.2\,$\pm$\,4.0}  & 0.85\,$\pm$\,0.10 &   8.5\,$\pm$\,2.8   &   {\bf 0.93\,$\pm$\,0.05}
    &    21.4\,$\pm$\,7.9 & 0.71\,$\pm$\,0.14 &  {\bf 8.1\,$\pm$\,3.2} & 0.85\,$\pm$\,0.05 \\ 
& AS-FedDAG &  {\bf 5.0\,$\pm$\,4.2}  & {\bf 0.88\,$\pm$\,0.11 } &  {\bf 3.3\,$\pm$\,2.5}   &   0.92\,$\pm$\,0.07
    &    \textbf{20.1\,$\pm$\,8.3} & \textbf{0.72\,$\pm$\,0.14} &  \textbf{5.6\,$\pm$\,2.8} &  \textbf{0.86\,$\pm$\,0.06} \\
\bottomrule
\end{tabular}}
\end{table}

\begin{table}[ht]
\centering
\caption{Results on heterogeneous setting with the different number of observations, (20 nodes, ER2).}
\label{tab:results_noniid_datanum}
\resizebox{0.95\textwidth}{!}{
\begin{tabular}{clllllllllllll}
\toprule
~&  ~& \multicolumn{2}{c}{n =100} & \multicolumn{2}{c}{n =300} & \multicolumn{2}{c}{n =600} & \multicolumn{2}{c}{n =900} \\
    \cmidrule(lr){3-4} \cmidrule(lr){5-6} \cmidrule(lr){7-8} \cmidrule(lr){9-10}
~& ~& \multicolumn{1}{c}{SHD $\downarrow$} & \multicolumn{1}{c}{TPR $\uparrow$} & \multicolumn{1}{c}{SHD $\downarrow$} & \multicolumn{1}{c}{TPR $\uparrow$}
    & \multicolumn{1}{c}{SHD $\downarrow$} & \multicolumn{1}{c}{TPR $\uparrow$} & \multicolumn{1}{c}{SHD $\downarrow$} & \multicolumn{1}{c}{TPR $\uparrow$}
\\ \midrule
\multirow{6}{*}{\rotatebox{90} {All   data}} & PC &  55.5\,$\pm$\,8.5  & 0.21\,$\pm$\,0.06 &   57.3\,$\pm$\,5.7   &   0.29\,$\pm$\,0.07
    &    60.4\,$\pm$\,9.8 & 0.32\,$\pm$\,0.11 &  62.4\,$\pm$\,6.6 & 0.29\,$\pm$\,0.10 \\
& GES   &  82.8\,$\pm$\,13.7  & 0.38\,$\pm$\,0.12 &   96.4\,$\pm$\,14.9   &   0.48\,$\pm$\,0.08
    &    102.9\,$\pm$\,13.6 & 0.51\,$\pm$\,0.08 &  106.3\,$\pm$\,14.3 & 0.50\,$\pm$\,0.11 \\
& DAG-GNN    &  61.8\,$\pm$\,14.7  & 0.39\,$\pm$\,0.07 &   56.8\,$\pm$\,9.7   &   0.37\,$\pm$\,0.08
    &    57.7\,$\pm$\,12.0 & 0.38\,$\pm$\,0.08 &  57.9\,$\pm$\,12.1 & 0.32\,$\pm$\,0.08 \\
& NOTEARS    & 58.7\,$\pm$\,12.8  & 0.41\,$\pm$\,0.12 &   57.6\,$\pm$\,10.2   &   0.44\,$\pm$\,0.06
    &    57.3\,$\pm$\,12.9 & 0.43\,$\pm$\,0.08 &  59.4\,$\pm$\,10.3 & 0.39\,$\pm$\,0.10 \\
& N-S-MLP  &  111.2\,$\pm$\,14.4  & 0.92\,$\pm$\,0.10 &   101.0\,$\pm$\,16.8   &   0.92\,$\pm$\,0.05
    &    100.8\,$\pm$\,14.7  & 0.90\,$\pm$\,0.10 &  97.6\,$\pm$\,14.8 & 0.90\,$\pm$\,0.07 \\
& MCSL &  49.0\,$\pm$\,8.1  & 0.62\,$\pm$\,0.06 &   54.0\,$\pm$\,10.0   &   0.70\,$\pm$\,0.10
    &    53.8\,$\pm$\,9.6 & 0.73\,$\pm$\,0.10 &  57.6\,$\pm$\,11.6 & 0.73\,$\pm$\,0.08 \\ \midrule
\multirow{6}{*}{\rotatebox{90} {Sep  data}} & PC&   31.2\,$\pm$\,5.7  & 0.30\,$\pm$\,0.05 &   29.0\,$\pm$\,5.9   &   0.39\,$\pm$\,0.06
    &    28.5\,$\pm$\,6.3 & 0.44\,$\pm$\,0.07 &  27.9\,$\pm$\,6.6 & 0.47\,$\pm$\,0.08 \\
& GES   &    35.1\,$\pm$\,8.3  & 0.48\,$\pm$\,0.10  &  31.6\,$\pm$\,9.8   &   0.57\,$\pm$\,0.08
    &   30.0\,$\pm$\,8.0 &  0.62\,$\pm$\,0.06 &  30.5\,$\pm$\,10.7& 0.64\,$\pm$\,0.07 \\
& DAG-GNN  &  29.9\,$\pm$\,7.2  & 0.66\,$\pm$\,0.09 &   20.3\,$\pm$\,5.0   &   0.67\,$\pm$\,0.09
    &    18.5\,$\pm$\,4.9 & 0.67\,$\pm$\,0.09 &  18.0\,$\pm$\,5.2 & 0.66\,$\pm$\,0.11 \\
& NOTEARS  &    \textbf{16.3\,$\pm$\,3.4}  & 0.61\,$\pm$\,0.08   & \textbf{15.5\,$\pm$\,3.2}   &  0.60\,$\pm$\,0.08
    &  15.0\,$\pm$\,3.1 & 0.62\,$\pm$\,0.09 & 15.2\,$\pm$\,2.9 & 0.61\,$\pm$\,0.09 \\
& N-S-MLP   &   68.0\,$\pm$\,5.4  &\textbf{0.80\,$\pm$\,0.04}   & 22.6\,$\pm$\,3.3   &  \textbf{0.79\,$\pm$\,0.06}
    &  \textbf{12.7\,$\pm$\,2.6} & \textbf{0.80\,$\pm$\,0.05} & \textbf{11.8\,$\pm$\,2.8}  & \textbf{0.80\,$\pm$\,0.05}\\
& MCSL &    32.8,$\pm$\,5.4  & 0.49\,$\pm$\,0.08   & 26.4\,$\pm$\,5.5  &  0.53\,$\pm$\,0.09
    &  23.3\,$\pm$\,5.8 & 0.56\,$\pm$\,0.08 & 23.1\,$\pm$\,6.5  & 0.56\,$\pm$\,0.07 \\ \midrule
& GS-FedDAG &    {\bf 11.6\,$\pm$\,5.6}  & {\bf 0.83\,$\pm$\,0.11} &  {\bf 7.1\,$\pm$\,6.1}   &  {\bf 0.90\,$\pm$\,0.12} 
    &   {\bf 6.2\,$\pm$\,4.7 } & {\bf 0.89\,$\pm$\,0.09} &  {\bf 6.0\,$\pm$\,5.5} & {\bf 0.91\,$\pm$\,0.11} \\ 
\bottomrule
\end{tabular}}
\end{table}

\begin{table}[ht]
\centering
\caption{Results on randomly selecting models-info of partial clients  (heterogeneous data, 20 nodes, ER2).}
\label{tab:results_client_selection}
\resizebox{0.95\textwidth}{!}{
\begin{tabular}{clllllllllllll}
\toprule
~&  ~& \multicolumn{4}{c}{Homogeneous data} & \multicolumn{4}{c}{Heterogeneous data} \\
    \cmidrule(lr){3-6} \cmidrule(lr){7-10}
~&  ~& \multicolumn{2}{c}{ER2 with 10 nodes} & \multicolumn{2}{c}{ER2 with 20 nodes} & \multicolumn{2}{c}{ER2 with 10 nodes} & \multicolumn{2}{c}{ER2 with 20 nodes} \\
    \cmidrule(lr){3-4} \cmidrule(lr){5-6} \cmidrule(lr){7-8} \cmidrule(lr){9-10}
~& ~& \multicolumn{1}{c}{SHD $\downarrow$} & \multicolumn{1}{c}{TPR $\uparrow$} & \multicolumn{1}{c}{SHD $\downarrow$} & \multicolumn{1}{c}{TPR $\uparrow$}
    & \multicolumn{1}{c}{SHD $\downarrow$} & \multicolumn{1}{c}{TPR $\uparrow$} & \multicolumn{1}{c}{SHD $\downarrow$} & \multicolumn{1}{c}{TPR $\uparrow$}
\\ \midrule
\multirow{5}{*}{$\frac{r}{m}$} & $10\%$ &  3.8\,$\pm$\,2.4  & 0.78\,$\pm$\,0.14 &  8.6\,$\pm$\,4.8   &   0.77\,$\pm$\,0.13
    &   3.8\,$\pm$\,1.4 & 0.93\,$\pm$\,0.05 &  8.5\,$\pm$\,5.4 & 0.89\,$\pm$\,0.07 \\
& 20\%   &  3.2\,$\pm$\,2.0  & 0.81\,$\pm$\,0.12 &   6.7\,$\pm$\,4.8   &   0.82\,$\pm$\,0.13
    &    2.5\,$\pm$\,2.1 & 0.97\,$\pm$\,0.04 &  8.2\,$\pm$\,5.4 & 0.87\,$\pm$\,0.09 \\
&  50\%  & 2.9\,$\pm$\,1.8  & 0.83\,$\pm$\,0.11  &  5.8\,$\pm$\,4.4   &   0.85\,$\pm$\,0.12
    &  1.8\,$\pm$\,1.4 & 0.99\,$\pm$\,0.02 &  6.3\,$\pm$\,5.1 & 0.89\,$\pm$\,0.10 \\
& 80\%    &  2.7\,$\pm$\,1.9  & 0.84\,$\pm$\,0.12 &   6.0\,$\pm$\,3.9   &  0.86\,$\pm$\,0.10
    &    1.8\,$\pm$\,1.3 & 0.99\,$\pm$\,0.02 &  5.9\,$\pm$\,4.1 & 0.90\,$\pm$\,0.08 \\
& 100\%    & 2.4\,$\pm$\,2.0  & 0.86\,$\pm$\,0.12 &   6.2\,$\pm$\,4.0   &   0.85\,$\pm$\,0.10
    &  1.9\,$\pm$\,1.6 & 0.99\,$\pm$\,0.02 &  6.2\,$\pm$\,4.7 & 0.89\,$\pm$\,0.09 \\
\bottomrule
\end{tabular}}
\end{table}

\begin{table}[ht]
\centering
\caption{Empirical results on \textbf{fMRI Hippocampus} dataset (Part 2).}
\label{tab:real_data_supp}
\resizebox{0.95\textwidth}{!}{
\begin{tabular}{lllllllll}
\toprule
  & \multicolumn{3}{c}{All data} & \multicolumn{3}{c}{Separate data} & \multirow{2}{*}{GS-FedDAG} & \multirow{2}{*}{AS-FedDAG} \\ \cmidrule(lr){2-4} \cmidrule(lr){5-7} 
 & GES & N-S-MLP & DAG-GNN &  GES & N-S-MLP & DAG-GNN \\ \midrule
SHD $\downarrow$ & 8.0\,$\pm$\,0.0 & 9.0\,$\pm$\,0.0  & 5.4\,$\pm$\,0.5 & 8.3\,$\pm$\,1.2 & 11.3\,$\pm$\,1.0 & 8.2\,$\pm$\,1.9 & \textbf{6.4\,$\pm$\,0.9} &  \textbf{5.0\,$\pm$\,0.0} \\
NNZ & 11.0\,$\pm$\,0.0 & 12.0\,$\pm$\,0.0  & 3.3\,$\pm$\,0.8 & 8.5\,$\pm$\,1.1 & 14.4\,$\pm$\,0.8 & 5.7\,$\pm$\,1.4 & 6.8\,$\pm$\,0.6 & 5.0\,$\pm$\,0.0 \\
TPR $\uparrow$ & 0.43\,$\pm$\,0.00 & 0.43\,$\pm$\,0.00  & 0.23\,$\pm$\,0.07 & 0.31\,$\pm$\,0.17 & \textbf{0.44\,$\pm$\,0.10} &  0.17\,$\pm$\,0.18 & 0.27\,$\pm$\,0.12 & \textbf{0.29\,$\pm$\,0.00} \\
FDR $\downarrow$ & 0.73\,$\pm$\,0.00 & 0.75\,$\pm$\,0.00  & 0.52\,$\pm$\,0.09 & 0.75\,$\pm$\,0.12 & 0.78\,$\pm$\,0.05 & 0.80\,$\pm$\,0.18 & \textbf{0.72\,$\pm$\,0.11} & \textbf{0.60\,$\pm$\,0.00} \\
\bottomrule
\end{tabular}}
\end{table}

\begin{figure}[ht]
\centering
\begin{center}
    \includegraphics[width=50mm]{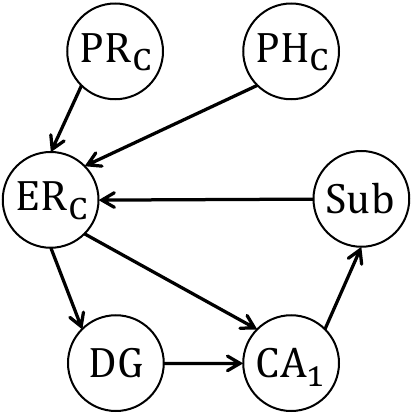}
\end{center}
\caption{Anatomical causal-effect relationships of \textbf{fMRI Hippocampus} dataset}
\label{fig:real_data_gt_graph}
\end{figure}

\section{More discussions on the experimental results}
\label{app:result_discussions}
\textbf{Why does our method outperform other methods even some baseline methods using all data for training?} \\
Let us first discuss the AS-FedDAG (All-Shared FedDAG), which shares all model parameters (both $\bm \Phi$ and $\bm U$) among all clients. If we set $it_{fl}$ as $1$ in AS-FedDAG, AS-FedDAG is totally the same as MCSL using all data for training. For simplicity, we mark all parameters (actually $\bm \Phi$ and $\bm U$) of client $c_k$ together as $\theta^{c_k}$. Let us consider the $t$-th iteration when all clients receive the average parameters $\theta_t$ from the server and update their parameters by $\theta_t$.  

For AS-FedDAG, firstly, we mark the gradients obtained by using the local data of client $c^k$ for $k \in [m]$ as $g^{c_k}_t$. Then each client $c_k$ updates its parameters for one step by $\theta^{c_k}_t = \theta_t - lr\times g^{c_k}_t$, where $lr$ is the learning rate. Afterward, the server collects all parameters and averages them to get $\theta_{t+1} = \frac{\sum_{k=1}^{m}\theta^{c_k}_t}{m} = \frac{\sum_{k=1}^{m}(\theta_t-lr\times g^{c_k}_t)}{m} =\theta_t-lr\times \frac{\sum_{k=1}^{m} g^{c_k}_t}{m}$. For MCSL, there is only one $\theta$. If MCSL uses full gradient information, then $\theta_{t+1} = \theta_t-lr\times \frac{\sum_{k=1}^{m} g^{c_k}_t}{m}$ (the full gradient is just the average of gradients from all samples). We can find that the updated parameters are totally the same. Then if $it_{fl} > 1$, we average all parameters every $it_{fl}$ iterations. Even though the exact updating procedures are not the same, the expectations of updated parameters are the same. This is why we say that \textit{MCSL trained on all data can serve as an approximate upper bound of our method but unobtainable} in our paper. 

In GS-FedDAG (Graph-Shared FedDAG) method, only all graphs are averaged. However, this partial information-sharing mechanism also helps on benefiting information from other clients to find a better solution~\citep{collins2021exploiting}.  
\section{Discussions on Assumptions}

\subsection{Data heterogeneity}
\label{App:diss_data_heterogeneity}
The general heterogeneous data setup should include the distribution shift caused by interventions since interventions on certain variables would also lead to heterogeneous distribution. Previous work~\citep{huang2020causal} has investigated this case and proposes the CD-NOD algorithm, which enhances the PC method, to learn from heterogeneous data. However, CD-NOD needs to identify some edge directions by capturing the changing information among distributions. That is to say, this method which needs to gather all data and cause the raw data leakage, of course. In our paper, we restrict our attention to the ANMs, which care more about the mechanisms and noise shift among different clients. Moreover, finding the identifiability conditions for learning graphs from the general heterogeneous data (both mechanisms shift and interventional data) in the federated setup is a challenging but important problem, which is left for future work.

\subsection{Invariant DAG assumption}
\label{app:invariant_dag}
Firstly, let us skip the homogeneous data setting of FedDAG, which only assumes all SEMs are totally the same but data are generated at different local clients. Then, we mainly talk about the heterogeneous setting that assumes SEMs vary, but DAG is shared among different clients. 

Essentially, an SEM models the physical processes of a system and the generation process behind observations. Intuitively, different SEMs usually describe different systems. Then, naturally, the DAGs may be different. In the case that the deployed systems on different clients are not the same, our method will break down because of the model misspecification. Unfortunately, it is not straightforward to extend our current framework to deal with this case, and we leave it for future work.

In this paper, we leave the variant causal graphs case aside and focus on the invariant graph case. This can be explained by the fact that a system can have various SEMs at different statuses~\citep{huang2020causal}. In the real world, some cases can be supported by our assumptions. The first example can be fMRI recordings. As pointed in~\citep{huang2020causal}, fMRI recordings are usually non-stationary because information flows in the brain may change with stimuli, tasks, and attention of the subject. Our federated setting only has one more assumption that fMRI recordings among different clients cannot be shared. The second example can be causal gene regulatory network inference~\citep{omranian2016gene}. The causal direction among genes, i.e., which gene regulates which gene, is believed to be the same. However, the SEM mechanism could vary in each individual due to individual properties, such as age, gender, etc. Also, the assumption that domain shifts can also come from the distribution shifts of the exogenous variables (noise terms in our paper) has been widely accepted in the machine learning field, such as invariant causal prediction~\citep{peters2016causal}, IRM~\citep{arjovsky2019invariant}.




\end{document}